\documentclass{ieeetj}

\usepackage{caption}
\usepackage{subcaption}
\usepackage{amsmath,amssymb,amsfonts}
\usepackage{algorithmic}
\usepackage{graphicx,color}
\usepackage{textcomp}
\usepackage{xcolor}
\usepackage{colortbl}
\usepackage{indentfirst}
\usepackage{url}
\usepackage[sort&compress,numbers]{natbib}
\usepackage{hyperref}
\usepackage[capitalize,nameinlink]{cleveref}
\crefname{section}{Sec.}{Secs.}
\Crefname{section}{Sec.}{Secs.}
\crefname{subsection}{Sec.}{Secs.}
\Crefname{subsection}{Sec.}{Secs.}
\crefname{subsubsection}{Sec.}{Secs.}
\Crefname{subsubsection}{Sec.}{Secs.}
\Crefname{figure}{Fig.}{Figs.}
\Crefname{table}{Table}{Tables}
\makeatletter
\renewcommand{\p@subsection}{\thesection.}
\renewcommand{\p@subsubsection}{\thesection.\thesubsection.}
\makeatother
\usepackage{multirow}
\usepackage{algorithm,algorithmic}
\usepackage{booktabs}
\usepackage{xspace}
\usepackage{makecell}

\usepackage{amssymb}
\usepackage{pifont}

\def\BibTeX{{\rm B\kern-.05em{\sc i\kern-.025em b}\kern-.08em
    T\kern-.1667em\lower.7ex\hbox{E}\kern-.125emX}}
\AtBeginDocument{\definecolor{tmlcncolor}{cmyk}{0.93,0.59,0.15,0.02}\definecolor{NavyBlue}{RGB}{0,86,125}}

\usepackage{Packages/sparo_acronyms}
\usepackage{Packages/sparo_math}
\usepackage{Packages/sparo_SIunits}
\usepackage{Packages/sparo_misc}

\definecolor{bl}{HTML}{4285F4}

\definecolor{rl}{HTML}{DB4437}
\newcommand{\rl}[1]{{\textcolor{rl}{#1}}}

\definecolor{gl}{HTML}{0F9D58}
\newcommand{\gl}[1]{{\textcolor{gl}{#1}}}

\DeclareMathOperator*{\argmin}{argmin}

\newcommand{\caveconcept}{\emph{cave exploration}\xspace}
\newcommand{\surfaceconcept}{\emph{scientific surface exploration}\xspace}

\def\OJlogo{\vspace{-4pt}$<$Society logo(s) and publication title will appear here.$>$}
\def\seclogo{\vspace{10pt}$<$Society logo(s) and publication title will appear here.$>$}

\def\authorrefmark#1{\ensuremath{^{\textbf{#1}}}}

\newlength{\biogapA}
\newlength{\biogapB}
\newlength{\biogapC}

\begin{document}
\receiveddate{9 February, 2026}
\reviseddate{22 June, 2026}
\accepteddate{31 July, 2026}
\doiinfo{XXXX.2022.1234567}
\markboth{}{Kim {et al.}}

\title{SKiD-SLAM: Resource-Aware Distributed Multi-Robot LiDAR SLAM for Planetary Missions with Field Deployments at Analog Sites}

\author{Hogyun Kim\authorrefmark{1}, Juwon Kim\authorrefmark{1}, Dongjin Cho\authorrefmark{1}, Jiwon Choi\authorrefmark{1}, Geonmo Yang\authorrefmark{1}, \\
Seokhwan Jeong\authorrefmark{1}, Jiyun Lee\authorrefmark{1}, Miryeong Park\authorrefmark{1}, Hyungtae Lim\authorrefmark{2}, and Younggun Cho\authorrefmark{1}}
\affil{Department of Electrical and Computer Engineering, Inha University, Incheon 22212, Republic of Korea}
\affil{Laboratory for Information \& Decision Systems (LIDS), Massachusetts Institute of Technology (MIT), Cambridge, MA 02139, USA}
\corresp{Corresponding authors: Hyungtae Lim and Younggun Cho (e-mail: shapelim@mit.edu; yg.cho@inha.ac.kr).}
\authornote{This work was supported by National Research Foundation of Korea~(NRF) grant~(Nos. RS-2026-2555148 and RS-2025-02217000) and Institute of Information \& Communications Technology Planning \& Evaluation~(IITP) grant~(RS-2022-II220448) \\ funded by the Korea government~(MSIT).}

\begin{abstract}
In future planetary exploration missions, teams of robots will need to collaboratively build maps in unknown environments, from subsurface caves and lava tubes to crater-strewn surface terrains.
Such collaboration enables faster and more robust mapping than a single robot can achieve by sharing observations and computation across the team.
However, this collaboration typically assumes seamless communication and abundant computing resources, whereas planetary environments impose severe communication limits, constrained onboard computation, and the absence of any central infrastructure.
In this field report, we present \textit{SKiD-SLAM}, a distributed multi-robot light detection and ranging~(LiDAR) simultaneous localization and mapping~(SLAM) system for such communication-constrained planetary environments, and report its deployment across two representative mission concepts, \ie (i)~\caveconcept and (ii)~\surfaceconcept.
To build the system, we survey the communication and computation budgets of actual planetary missions, and select modules that fit these budgets for each core function of distributed multi-robot LiDAR SLAM.
These core functions are recognizing revisited places across robots' maps, estimating inter-robot relative poses, and fusing trajectories without a central server.
Before field deployment, we verify the assembled system on public datasets that resemble planetary environments and in a simulator built on the National Aeronautics and Space Administration~(NASA) Jet Propulsion Laboratory~(JPL)'s Open Source Rover.
We then validate our SKiD-SLAM in both mission concepts through field deployments in South Korea, where (i)~a team of wheeled rovers explores the Majang lava-tube cave on Jeju Island, and (ii)~quadruped robots carry out a crater-and-rock science scenario on a planetary analog terrain at the Korea Institute of Civil Engineering and Building Technology~(KICT) in Yeoncheon.
Across both deployments, our SKiD-SLAM reliably performed distributed multi-robot LiDAR SLAM within the 250\,kbps inter-robot communication budget, demonstrating its feasibility under challenging field conditions.
The supplementary materials, including our code and field data, are publicly available at \href{https://sparolab.github.io/research/skid_slam/}{\texttt{https://sparolab.github.io/research/skid\_slam/}}.
\end{abstract}

\begin{IEEEkeywords}
Distributed LiDAR SLAM, Multi-Robot Systems, Planetary Analog Environment, CADRE Project, Space Robotics, Field Robotics
\end{IEEEkeywords}


\maketitle
\section{Introduction}
\IEEEPARstart{T}{he} Cooperative Autonomous Distributed Robotic Exploration~(CADRE) project of the National Aeronautics and Space Administration~(NASA)~\cite{NASA_CADRE, delaCroix2024CADRE, rabideau2025planning, saboia2024cadre} aims to develop and demonstrate distributed multi-robot autonomy for future planetary exploration missions.
In such missions, teams of autonomous robots will need to collaboratively build maps in unknown environments on the Moon and Mars, including caves, lava tubes, and crater-strewn surfaces, to support scientific exploration and resource utilization.
However, achieving robust and efficient multi-robot simultaneous localization and mapping~(SLAM) in these resource-constrained field environments presents substantial challenges, including limited communication bandwidth, constrained onboard computation, and the absence of any central infrastructure~\cite{stathoulopoulos2023frame, stathoulopoulos2024frame, lajoie2025multi, varadharajan2025multi, jeong2024diter, kim2025diter++, giubilato2026s3li}.

Recently, distributed multi-robot light detection and ranging~(LiDAR) SLAM systems have achieved remarkable progress, demonstrating accurate inter-robot loop closing and globally consistent mapping across robot teams~\cite{chang2022lamp, kim2022lt, gil2025ephemerality, damigos2024communication, kang2025uni, wei2024large, lim2025multi, huang2022disco, xie2021rdc, zhong2023dcl, lajoie2023swarm, lyu2025drcm, wang2026communication}.
However, most of these systems have been developed and validated on urban and indoor benchmarks, where distinctive geometric structure, high-bandwidth wireless links, and supporting infrastructure are often available.
These assumptions do not hold in planetary environments.
Consequently, their behavior under the kilobit-class, intermittent, and infrastructure-free conditions of planetary missions remains largely unexamined.

This gap motivates our field report.
Rather than proposing another algorithm, we select each module of distributed multi-robot LiDAR SLAM to fit the communication and computation budgets of actual planetary missions, and report on the operation of the resulting system in the field.
Concretely, we organize our study around two representative planetary mission concepts.
These two concepts, distinguished by terrain, platform, and team composition, recur throughout the report, from the mission concepts and field deployments to the simulation, evaluation, and lessons learned.
Drawing on these deployments, we further distill lessons for operating distributed multi-robot LiDAR SLAM beyond benchmark conditions.

\subsection{Mission Concept}
Here, we describe the two mission concepts that frame our study of distributed multi-robot exploration, along with the operating assumptions they share.
The first is \caveconcept~\cite{dominguez2025cooperative}.
A team cooperatively maps a confined, lightless lava tube, a type of subsurface environment that may shelter future habitats but that no planetary robot has yet entered.
The entire team shares a single objective, \ie to build one consistent map of the passages.

\begin{figure*}[t]
    \centering
    \def\figwidth{\textwidth}%
        {%
        \includegraphics[clip, trim= 0 0 0 0, width=\figwidth]{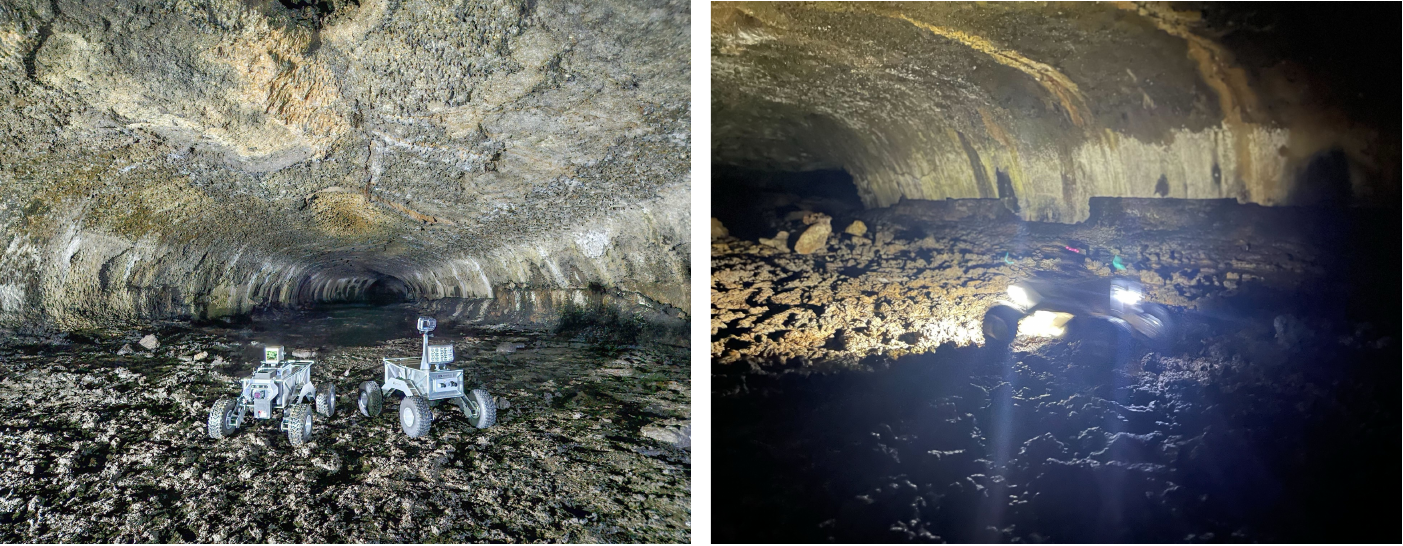}
        }
        \vspace{-0.4cm}
        \captionsetup{justification=justified}
        \caption{Field deployment at the Majang lava-tube cave on Jeju Island.
        The cave floor is covered with loose rocks and rubble, and its geometrically self-similar walls receive no natural light, conditions representative of planetary lava tubes.
        We deploy wheeled rovers to perform distributed multi-robot SLAM in this GPS-denied environment.}
        \label{fig:fig1_majang}
        \vspace{-0.4cm}
\end{figure*}
\begin{figure*}[t]
    \centering
    \def\figwidth{\textwidth}%
        {%
        \includegraphics[clip, trim= 0 0 0 0, width=\figwidth]{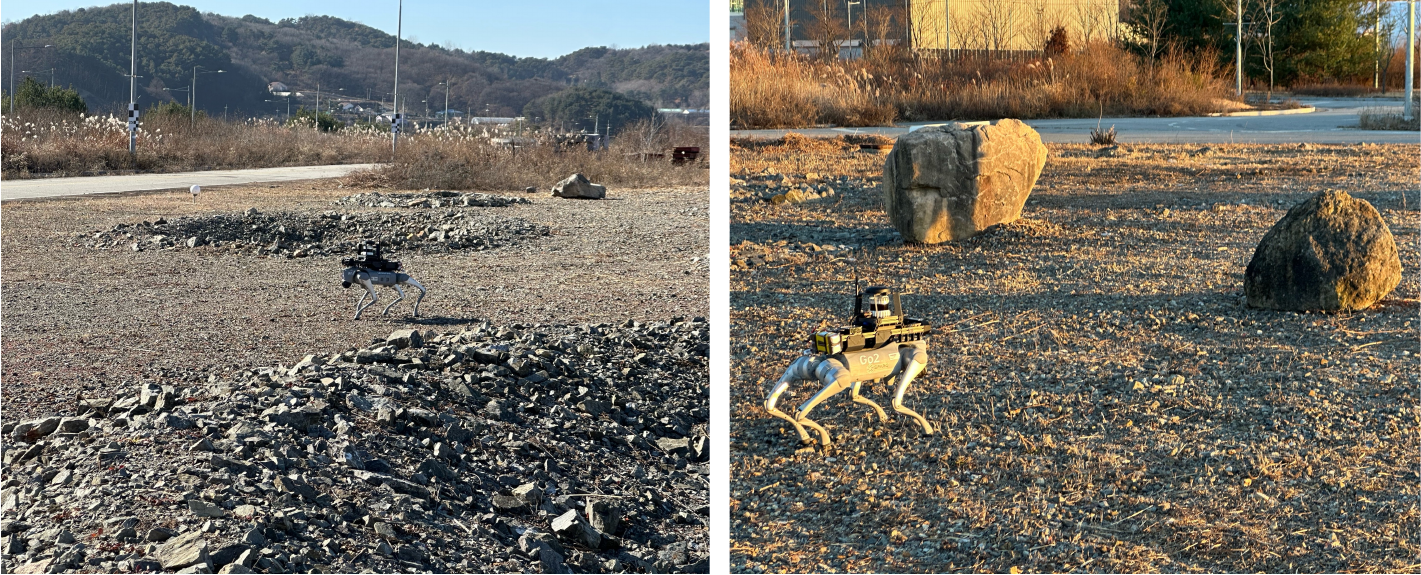}
        }
        \vspace{-0.4cm}
        \captionsetup{justification=justified}
        \caption{Field deployment at the KICT planetary analog terrain in Yeoncheon.
        The surface is shaped by numerous impact-crater-like depressions and scattered rocks that emulate the open surface conditions of the Moon.
        We deploy quadruped robots to perform distributed multi-robot SLAM across this cratered terrain.}
        \label{fig:fig2_yeoncheon}
        \vspace{-0.4cm}
\end{figure*}

The second is \surfaceconcept~\cite{arm2023scientific}.
A team explores open, cratered terrain, with the robots divided across complementary roles rather than sharing one objective.
In a representative scenario, one robot traverses the crater field to map its geometry, while another inspects scattered rocks for science-oriented observations.
Here, the team size follows from the division of labor in the scenario, not from platform availability.

Both concepts share the same starting condition.
A team of $N$ robots is deployed in an unknown environment, and each robot starts in its own local coordinate frame, with no shared global reference established a priori~\cite{huang2022disco, xie2021rdc, zhong2023dcl, lajoie2023swarm, lyu2025drcm, wang2026communication}.
Satellite imagery and orbital depth maps may provide a coarse initial estimate of each robot's pose in a global frame.
However, they lack the resolution required for precise localization or for inter-robot map alignment on the surface~\cite{krawciw2025sharing, nash2026software}.
Therefore, a consistent global map must be constructed on site, from the robots' own onboard observations.

Beyond this starting condition, both concepts share two operating assumptions.
First, the robots explore without any fixed infrastructure.
There is no fixed localization infrastructure (hereafter referred to as a base station), no positioning system such as the global positioning system~(GPS), and no central server with abundant compute for SLAM.
Instead, each robot carries an embedded-class onboard computer and a short-range radio, processes its observations onboard, and exchanges mapping data with the team over short-range wireless links~\cite{tranzatto2022cerberus, agha2022nebula}.
Robots cannot assume continuous connectivity, and they may operate beyond line of sight of one another for extended periods.
Therefore, each robot runs its SLAM pipeline onboard, and the team shares only the information needed to align and merge maps within the available budgets~\cite{choudhary2017distributed}.

Second, the robot platform may differ across mission concepts.
Each terrain demands a different form of mobility, \eg wheeled rovers where the ground is smooth and traversable, and legged robots where the terrain is rough and uneven~\cite{schuster2020arches}.
As the platform and LiDAR mounting change with the terrain, the system must remain robust to the resulting viewpoint differences~\cite{kang2025uni, lim2025multi}.

Within these concepts, the role of SKiD-SLAM is narrow but critical.
It must reconcile the robots' independent local maps into a single globally consistent map of the explored environment.
This must be achieved using only budget-compliant communication, without a central server, and without operator intervention in the mapping process.

\begin{figure}[htbp!]
	\centering
	\def\width{0.49\textwidth}%
        {%
		\includegraphics[clip, trim= 0 0 0 0, width=\width]{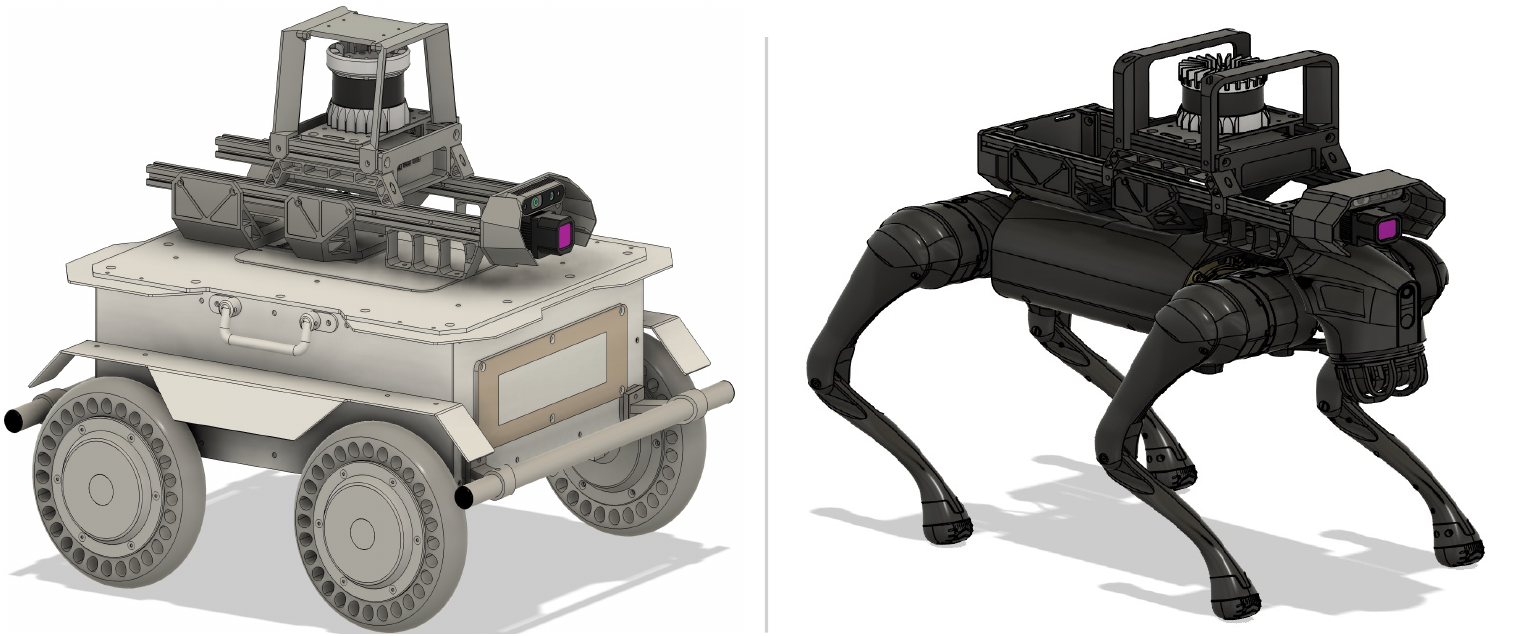}
	}
    \vspace{-0.25cm}
    \caption{Robot platforms and sensors, \ie LiDAR, IMU, camera, and thermal camera, used in our field experiments.
                 (Left)~A wheeled rover deployed for cave exploration at the Majang lava-tube cave on Jeju Island.
                 (Right)~A quadruped robot deployed for scientific surface exploration at the planetary analog testbed in Yeoncheon.
                 Note that the camera and thermal camera are used only for visual inspection and are not used by our SKiD-SLAM pipeline in this field report.
            }
    \label{fig:fig3_hardware}
    \vspace{-0.2cm}
\end{figure}

\begin{figure}[t]
    \centering
    \def\figwidth{0.49\textwidth}%
        {%
        \includegraphics[clip, trim= 0 0 0 0, width=\figwidth]{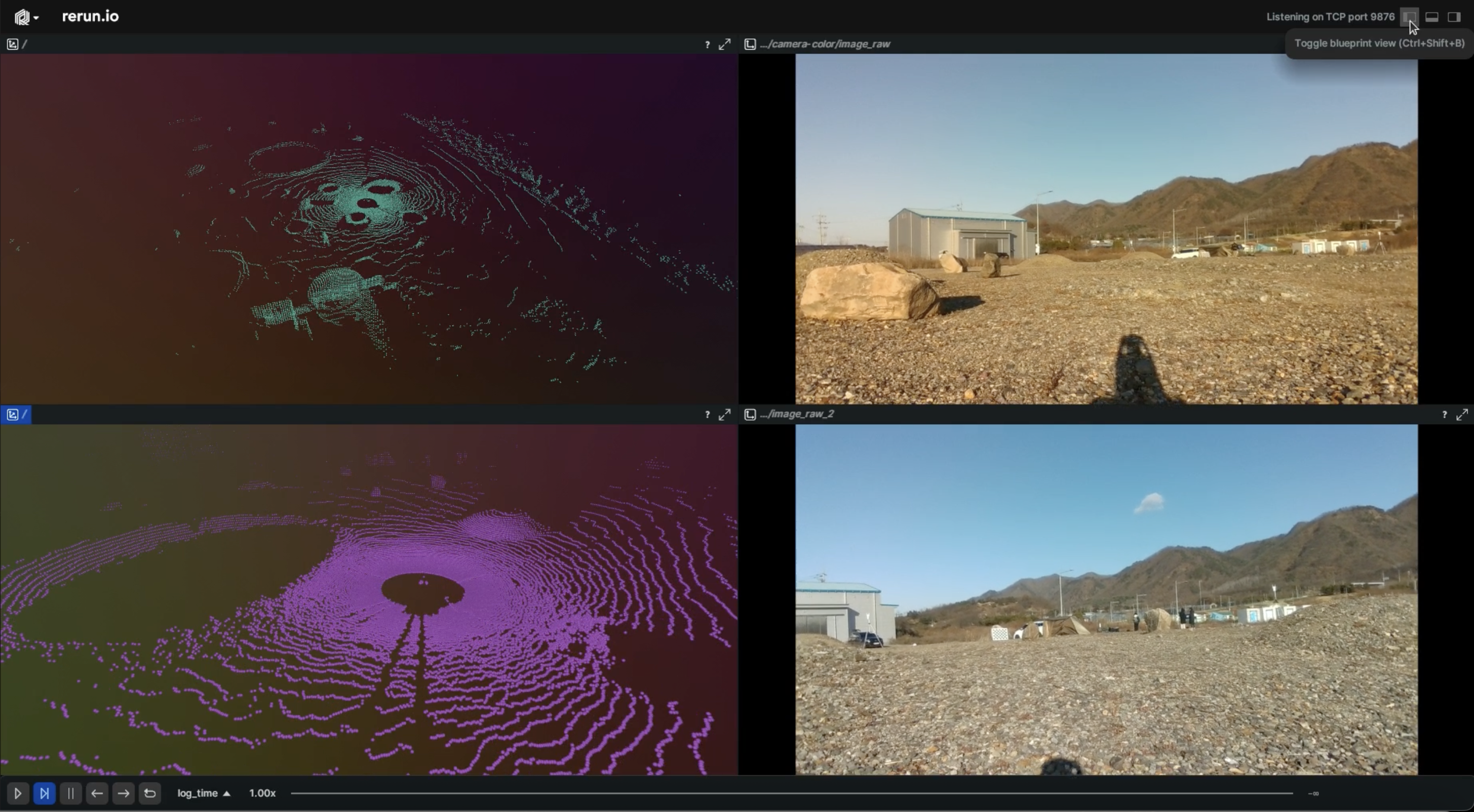}
        }
        \vspace{-0.25cm}
        \captionsetup{justification=justified}
    \caption{Our Rerun-based viewer~\cite{RerunSDK} for monitoring robot sensor data at an experimental monitoring workstation.
        The station communicates with the robots over a Wi-Fi access point to stream their data in real time.
        Here, two robots are deployed.
        (Top)~Scan data from robot~1, shown with the corresponding camera image.
        (Bottom)~Scan data from robot~2, shown with the corresponding camera image.
        The camera images are provided only for easier visual inspection and are not used by our algorithm, which relies solely on the LiDAR sensor.}
        \label{fig:fig4_viewer}
        \vspace{-0.2cm}
\end{figure}

\subsection{Field Test}
\label{sec:field_test}
We next describe how the two mission concepts were realized in the field, at two analog sites in South Korea.
For \caveconcept, the site is the Majang lava-tube cave on Jeju Island, a subterranean environment analogous to lunar and Martian lava tubes.
The traversed section spans roughly 20\,m $\times$ 40\,m, with ceiling heights ranging from 1\,m to 5\,m, and contains narrow and rugged passages that are difficult to traverse even for humans.
Critically, the cave receives no natural light, reproducing the permanently dark conditions of planetary caves and the lunar polar shadows that challenge vision-based perception.
We deployed three wheeled rovers from UEL\footnote{\url{https://uel.co.kr}} ($N=3$), whose drivetrains remain stable over the rough and uneven lava floor~(\figref{fig:fig1_majang})~\cite{lee2025soft}.

For \surfaceconcept, the site is a planetary analog terrain at the Korea Institute of Civil Engineering and Building Technology~(KICT) in Yeoncheon.
The site measures roughly 40\,m $\times$ 60\,m and reproduces a lunar surface through scattered rocks and impact-crater-like depressions, the deepest of which reach about 1\,m.
We deployed two Unitree\footnote{\url{https://www.unitree.com}} GO2 quadruped robots ($N=2$), whose legged mobility can negotiate the steep crater walls and scattered rocks of the site~(\figref{fig:fig2_yeoncheon}).

As shown in \figref{fig:fig3_hardware}, in both deployments, each robot was equipped with an Ouster OS1-64 LiDAR sensor\footnote{\url{https://ouster.com}}, a MicroStrain 3DM-GX5-25 inertial measurement unit~(IMU) sensor\footnote{\url{https://www.hbkworld.com}}, and an Intel NUC Pro Mini PC\footnote{\url{https://www.intel.com}} for running SKiD-SLAM.
Furthermore, the robots were teleoperated for safety, while our SKiD-SLAM performed distributed mapping onboard without any operator input.

For experimental transport and monitoring, the robots used a TP-Link\footnote{\url{https://www.tp-link.com}} Omada Wi-Fi access point, and each robot streamed its scans to a monitoring workstation for joint visualization using Rerun~\cite{RerunSDK}, as shown in \figref{fig:fig4_viewer}.
The access point emulated the physical inter-robot network for the field tests, while the workstation was used only for human supervision and was not part of the SLAM architecture.
All SLAM computation was performed onboard the robots in a fully distributed manner.
 
\section{Background}
\label{sec:background}
\subsection{Evolution of Planetary Exploration}
\label{sec:background:exploration}

\begin{figure*}[t]
    \centering
    \captionsetup{justification=justified}
    \captionsetup[subfigure]{justification=centering}



    \includegraphics[trim={0 0 0 0},clip,width=2\columnwidth]{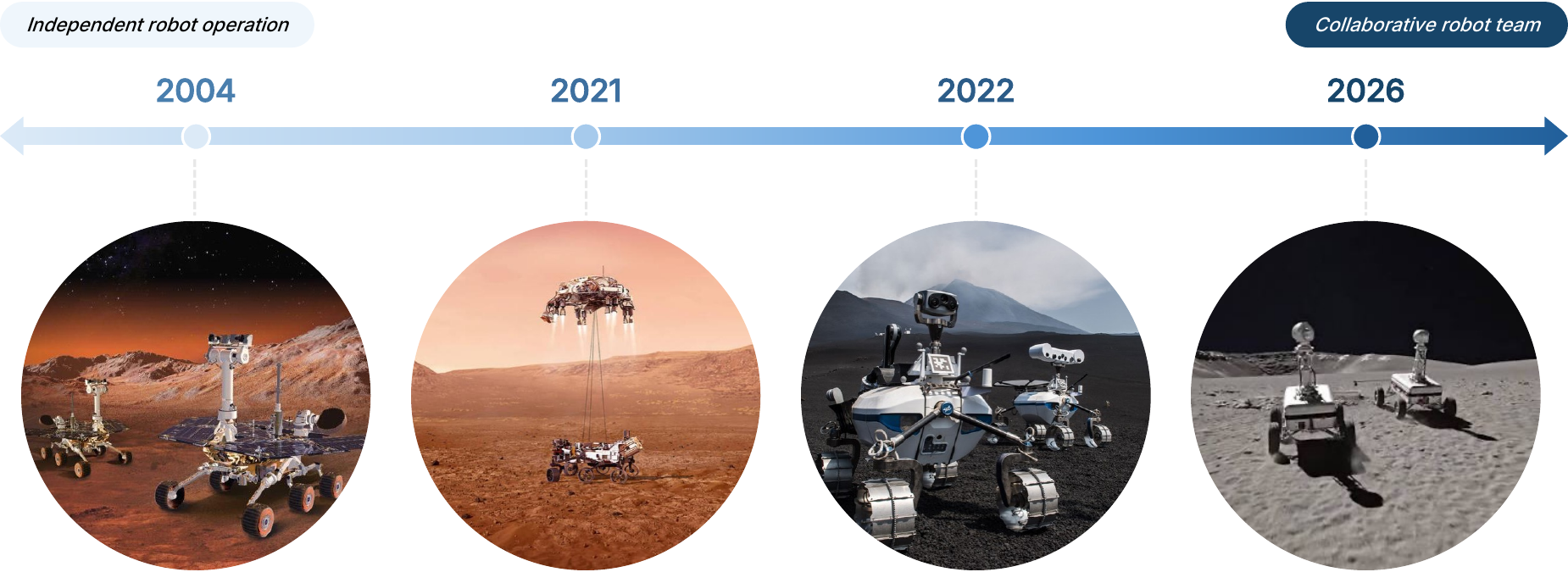}

    \vspace{0.2cm}   

    \begin{minipage}{1.92\columnwidth}
        \centering
        \begin{minipage}[t]{0.24\columnwidth}\centering
            \subcaption{Spirit and Opportunity~{\tiny(NASA)}\\Twin rovers operated independently at separate Martian sites}
        \end{minipage}\hfill
        \begin{minipage}[t]{0.23\columnwidth}\centering
            \subcaption{Perseverance~{\tiny(NASA)}\\A single, highly autonomous rover conducting solo surface science}
        \end{minipage}\hfill
        \begin{minipage}[t]{0.23\columnwidth}\centering
            \subcaption{ARCHES~{\tiny(DLR/ESA)}\\A heterogeneous robot team cooperating on a volcanic analog}
        \end{minipage}\hfill
        \begin{minipage}[t]{0.23\columnwidth}\centering
            \subcaption{CADRE~{\tiny(NASA)}\\A team of small rovers built to cooperate autonomously on the Moon}
        \end{minipage}
    \end{minipage}
    
    \caption{Representative milestones in planetary robotic exploration, showing a progression from isolated single-robot missions toward cooperative multi-robot teams.
    From left to right, the deployment paradigm shifts from independent operation, exemplified by NASA's Spirit and Opportunity~\cite{crisp2003mars} and Perseverance~\cite{farley2020mars}, toward cooperative autonomy, exemplified by the German Aerospace Center~(DLR) and European Space Agency~(ESA)'s ARCHES~\cite{schuster2020arches} and NASA's CADRE~\cite{NASA_CADRE,delaCroix2024CADRE,rabideau2025planning}.}
    \label{fig:fig5_background}
    \vspace{-0.3cm}
\end{figure*}

As shown in \figref{fig:fig5_background}, planetary robotic exploration has evolved from independent rovers operating alone on planetary surfaces toward cooperative robot teams now being prepared on Earth.
Planetary exploration seeks to answer questions that can only be resolved in situ, \eg the geological history of the Moon and Mars, the search for life, and the survey of resources for a long-term human presence.
Pursuing these questions requires robots that can perceive, map, and localize on their own.
For half a century, space agencies have explored planetary surfaces with one robot at a time.
NASA's lineage of Mars rovers illustrates this history.
First, Sojourner demonstrated that a wheeled robot could traverse another planet in 1997, covering roughly 100\,m around its lander over 83 sols~(Martian days)~\cite{golombek1997overview}.
Spirit and Opportunity~(\figref{fig:fig5_background}(a)) then conducted multi-year geological campaigns, with Opportunity ultimately driving more than 45\,km across Meridiani Planum~\cite{crisp2003mars}.
Next, Curiosity~\cite{grotzinger2012mars} and Perseverance~\cite{farley2020mars}~(\figref{fig:fig5_background}(b)) continue this lineage today, carrying laboratory-grade instruments to investigate past habitability, with Perseverance additionally caching samples for later return.
Other agencies have followed the same single-robot pattern.
For example, the China National Space Administration~(CNSA)'s Yutu-2 became the first rover on the lunar far side~\cite{li2021overview}, while the Indian Space Research Organisation~(ISRO)'s Pragyan operated near the lunar south pole~\cite{isro2023chandrayaan3}.

However, every one of these rovers explores alone, and their operation reflects this design.
Daily traverse plans are prepared by large ground teams, uplinked during limited communication windows, and executed with the rover largely on its own until the next contact through an orbiter or the Deep Space Network.
Even Ingenuity, the first mission to pair a second robot with a rover, was operated as a dependent asset of Perseverance rather than as a peer~\cite{balaram2018mars}.
Operating alone in this way leaves a mission fragile, dependent on a single platform surviving an unforgiving environment.
For example, a mobility failure can jeopardize an entire mission, as illustrated when Spirit became embedded in soft soil in 2009 and could not resume driving.

This fragility is what robot teams promise to overcome.
Multiple robots provide simultaneous measurements from different locations, cover terrain too risky for a single high-value asset, and degrade gracefully when an individual platform fails.
Crucially, however, multi-robot planetary exploration remains at an early stage; aside from the LEV-1/LEV-2 demonstration described below, it is still being prepared primarily on Earth through planetary analog campaigns.

The German Aerospace Center~(DLR)'s ARCHES~(\figref{fig:fig5_background}(c)) campaign demonstrated a heterogeneous robot team on a volcanic lunar analog, where a dedicated relay rover had to be deployed to maintain continuous connectivity~\cite{schuster2020arches}.
The European Union~(EU) CoRob-X project targets multi-robot exploration of lunar lava tubes, which are similar to the subsurface environments that motivate our cave mission concept~\cite{dettmann2022corob}.
The one case to reach space so far underscores how early this stage is: in 2024, the Japan Aerospace Exploration Agency~(JAXA)'s LEV-1~\cite{jaxa2024slim} and LEV-2~\cite{inazawa2024autonomous} were deployed together on the lunar surface, with LEV-2 sending its imagery to LEV-1 over a Bluetooth link and LEV-1 relaying it to Earth over a UHF direct-to-Earth link.
NASA's CADRE project~(\figref{fig:fig5_background}(d)) will deploy three cooperative rovers to the lunar Reiner Gamma region, and is being validated on the ground ahead of its lunar demonstration~\cite{NASA_CADRE, delaCroix2024CADRE, rabideau2025planning}.

This is precisely the stage at which our field report contributes, as we develop and validate distributed multi-robot SLAM in planetary analog environments on Earth, before such systems are deployed in space.
In particular, no robot team has yet operated inside an extraterrestrial cave or lava tube.
Our deployment in the Majang lava-tube cave directly addresses this gap, demonstrating distributed multi-robot SLAM in the kind of subsurface environment where future missions may need robots to operate instead of human surveyors.

\subsection{Communication and Computation Budgets in Planetary Missions}
\label{sec:background:budgets}

As discussed above, robot teams can make planetary missions more resilient than a single rover by distributing sensing and exploration across multiple platforms.
However, this advantage depends on the team's ability to share information, since cooperation fundamentally requires that robots communicate with one another.
Cooperation also requires each robot to process its own observations onboard, from local mapping to matching shared data.
On a planetary surface, both inter-robot communication and onboard computation are severely constrained.

Therefore, we examine (i)~the communication capacity available for sharing information between robots and (ii)~the computing capacity available onboard each robot.
We then adopt the operating budgets used throughout this work.

\subsubsection{Communication Capacity}
First, flown missions provide concrete numbers for how little bandwidth a planetary robot can expect.
Specifically, the direct-to-Earth link of the Curiosity rover operates between 500\,bps and 32\,kbps, and orbital relay raises the rate to 256\,kbps (Odyssey) or up to 2\,Mbps (MRO).
However, each orbiter is visible for only around eight minutes per sol~\cite{koktas2024communications}.

Moreover, robot-to-robot links on the surface are no more generous.
For example, the Ingenuity helicopter communicated with its base station on Perseverance through a 900\,MHz IEEE~802.15.4-based Zigbee radio~\cite{ieee802154} designed to relay data at 20\,kbps or 250\,kbps over distances of up to 1\,km~\cite{balaram2018mars}.
Similarly, the CADRE rovers communicate over mesh network radios providing a minimum of 1\,Mbps of aggregate bandwidth shared between all participating stations, and electromagnetic interference from the drive motors precludes continuous transmission across the mesh~\cite{saboia2024cadre}.

Consequently, inter-robot links in planetary missions are kilobit-class per robot, shared across the team, and intermittent by design.
For perspective, a modern home Wi-Fi connection based on Wi-Fi~6 is advertised at up to 9{,}600\,Mbps under ideal conditions, whereas the planetary inter-robot budget considered here is only 0.25\,Mbps.
This gap is especially severe for distributed LiDAR SLAM, where robots must repeatedly exchange descriptors and selected scan data rather than only short status messages; we revisit this system-level burden in \Cref{sec:field_eval:communication_budget}.

\subsubsection{Computing Capacity}
Onboard computation is constrained just as severely.
Radiation-hardened flight processors lag commercial hardware by decades, \eg the RAD750~\cite{berger2001rad750} that powers Curiosity and Perseverance runs at 133--200\,MHz with 256\,MB of RAM, far below even a modern embedded board.

To run more demanding autonomy, recent missions have turned to commercial off-the-shelf~(COTS) processors.
Ingenuity flew a Qualcomm Snapdragon 801, performing functions no radiation-hardened processor could have managed in that mission~\cite{balaram2018mars}, and the same class of processor is now reused on Perseverance for image-based global localization against orbital maps~\cite{nash2026software}.
Even so, solar power, battery capacity, and thermal limits cap the compute sustainable during a traverse.
While current flight processors remain highly constrained, future missions are expected to carry more capable COTS hardware as demonstrated by recent technology demonstrations.

\subsubsection{Operating Budgets}
Building on the communication and computation surveys in \Cref{sec:background:budgets}, we treat these two limits as the budgets that every module of SKiD-SLAM must respect.
For communication, following the robot-to-robot links flown to date~\cite{balaram2018mars} and IoT-class radio standards for infrastructure-free surface networks (\eg Wi-SUN, IEEE~802.15.4g~\cite{ieee802154g}), we assume at most 250\,kbps of inter-robot bandwidth within a 1\,km range.
For computation, we target compact onboard computers that future missions might realistically carry, acknowledging the evolving landscape of COTS processors in space applications.
Our validation focuses on ensuring that each module runs to completion without exhausting onboard memory, with real-time performance as a secondary goal.

Among the candidate methods for each module, we therefore select the one that runs within these communication and memory budgets while remaining as fast and accurate as possible, and assemble the selected modules into SKiD-SLAM.
 
\section{Related Works}
\label{sec:related_works}

As reported by \citet{burgard2005coordinated}, multi-robot systems are widely recognized to offer several advantages over single robot systems.
First, a multi-robot system has the potential to perform a single task faster than a single robot.
Second, using multiple robots introduces inherent redundancy, improving fault tolerance and making the team more resilient than a single robot.
Finally, overlapping information from multiple robots helps compensate for sensor uncertainty, improving overall reliability in field environments, where erroneous measurements are prevalent.

These same advantages motivate the move toward multi-robot planetary exploration.
For instance, NASA's CADRE project pursues exactly these goals, deploying a team of cooperative rovers to cover ground faster, to tolerate the loss of an individual robot, and to combine measurements from multiple vantage points~\cite{NASA_CADRE, delaCroix2024CADRE}.
However, in such planetary settings, these benefits must be realized under severe communication and computation limits, which makes the underlying multi-robot SLAM far more challenging than in terrestrial environments.

At the heart of this problem is inter- and intra-robot loop closure, which lets the robots recognize shared locations and merge their maps into a single consistent estimate.
Typically, loop closure is divided into three key components as follows: (i)~place recognition to identify candidate overlaps between observations from the same or different robots, (ii)~relative pose estimation, which we refer to as point cloud registration in this paper because our system uses LiDAR sensors, to estimate the transformation associated with each candidate overlap, and (iii)~a consistency check to reject false-positive loop closures before they distort the merged map.
In practice, place recognition first searches for candidate links using shared global descriptors.
Point cloud registration then estimates the relative pose of each candidate link from the associated scans.
Finally, the consistency check verifies the resulting constraints using scan data or each robot's local trajectory.
Based on these aspects, we analyze the existing multi-robot LiDAR SLAM frameworks.

Multi-robot LiDAR SLAM can be broadly classified into centralized and distributed approaches.
Centralized LiDAR SLAM~\cite{chang2022lamp, kim2022lt, gil2025ephemerality, damigos2024communication, kang2025uni, wei2024large, lim2025multi, kim2026commerge} depends on a remote server to aggregate maps and compute global SLAM for all robots.
However, in planetary environments, deploying such a central server is impractical, since there is no fixed infrastructure on site.
Therefore, we focus on distributed LiDAR SLAM~\cite{huang2022disco, xie2021rdc, zhong2023dcl, lajoie2023swarm, lyu2025drcm, wang2026communication}, which runs entirely on each robot's onboard computer and is thus inherently constrained by limited resources.
In the following, we review distributed LiDAR SLAM along the three components introduced above.

\subsection{Pose Estimation}
First, pose estimation provides each robot's local trajectory, serving as the front-end odometry.
LOAM~\cite{zhang2014loam} extracts edge and planar features for scan-to-map matching to achieve real-time and low-drift LiDAR odometry.
Building on this, subsequent methods tightly couple LiDAR and IMU to improve robustness and efficiency.
LIO-SAM~\cite{shan2020lio} fuses the two sensors within a factor graph smoothing framework.
FAST-LIO~\cite{xu2021fast} reformulates the Kalman gain of an iterated extended Kalman filter so that its cost scales with the state rather than the measurement dimension.
FAST-LIO2~\cite{xu2022fast} and Faster-LIO~\cite{bai2022faster} further accelerate map management through an incremental Kd-tree and an incremental sparse voxel structure, respectively.
Recently, various methods address specific failure modes.
For instance, DLIO~\cite{chen2023direct} applies continuous-time motion correction to compensate for point cloud distortion.
COIN-LIO~\cite{pfreundschuh2024coin} exploits LiDAR intensity images to reinforce registration in geometrically degenerate environments.

\subsection{Place Recognition}
Place recognition finds candidate links corresponding to identical or similar locations across inter- and intra-robot maps using shared global descriptors.
Existing approaches use lightweight global descriptors to address this issue.
For instance, \citet{huang2022disco} propose DiSCo-SLAM, which uses Scan Context~\cite{kim2018scan}, a descriptor designed for structured environments by dividing the surrounding space into bins and encoding each with the maximum elevation value.
\citet{xie2021rdc} propose RDC-SLAM, which uses DELIGHT~\cite{cop2018delight} for loop closure detection.
DELIGHT incorporates LiDAR intensity information into its descriptor, demonstrating its capability to operate in industrial parks with diverse characteristics, ranging from structured buildings to unstructured bushland.
\citet{zhong2023dcl} propose DCL-SLAM, which leverages LiDAR Iris~\cite{wang2020lidar}, representing LiDAR data as an image using Daugman’s rubber sheet model~\cite{john2009iris}.
\citet{lajoie2023swarm} propose Swarm-SLAM using both LiDAR-based Scan Context~\cite{kim2018scan} and vision-based place recognition with CosPlace~\cite{berton2022rethinking}.
\citet{lyu2025drcm} propose DRCM-CSLAM, which uses a learning-based approach for place recognition.
\citet{wang2026communication} propose a communication-efficient distributed SLAM system, which reduces the amount of data exchanged between robots.

Beyond the descriptors adopted in existing distributed SLAM frameworks~\cite{huang2022disco, xie2021rdc, zhong2023dcl, lajoie2023swarm, lyu2025drcm, wang2026communication}, several descriptors hold potential for inter-robot place recognition.
For instance, learning-based descriptors achieve strong place recognition performance.
LCDNet~\cite{cattaneo2022lcdnet} jointly identifies revisited places and estimates the 6-DoF relative transformation between point clouds, and is particularly robust to reverse loops.
LoGG3D-Net~\cite{vidanapathirana2022logg3d} introduces a local consistency loss so that local features remain consistent across revisits, yielding more repeatable global descriptors while operating in near real-time.
OverlapTransformer~\cite{ma2022overlaptransformer} augments a range-image network with a transformer to obtain a yaw-invariant descriptor that improves viewpoint robustness and reduces inference time.
However, they are difficult to apply in planetary settings because these networks must be trained on data representative of the deployment environment, which is scarce for planetary terrain that differs substantially from the urban and highway scenes used for training.
Moreover, many of these methods require computational resources beyond typical compact onboard computers.
In contrast, handcrafted descriptors such as SOLiD~\cite{kim2024narrowing} encode geometric features into compact representations, enabling lightweight and robust loop detection without any training, which makes them well-suited for resource-constrained planetary environments.

Although existing distributed SLAM systems already use relatively compact descriptors, their target settings are often networked campuses or urban areas rather than bandwidth-limited planetary missions.
Planetary operation therefore requires global descriptors with an even smaller communication footprint.

\subsection{Relative Pose Estimation via Point Cloud Registration}
Once place recognition identifies a revisited location shared by two robots, their relative pose at that location must be estimated through registration.
For this step, ICP-based local methods are commonly used in existing approaches~\cite{huang2022disco, zhong2023dcl}.
While these local methods can converge to accurate alignments when initialized near the correct solution, their basin of convergence is narrow~\cite{lim2024quatro++}.
Large inter-robot pose discrepancies can therefore push them into local minima or registration failure.

To address this limitation, global registration approaches search over a wider pose space to mitigate local minima, at the cost of increased computational complexity.
For instance, TEASER++~\cite{yang2020teaser} achieves robust registration through certifiable outlier rejection, but its heavy computation limits onboard applicability.
More recently, KISS-Matcher~\cite{lim2024kiss} integrates feature extraction, correspondence pruning, and efficient pose estimation into a unified framework, achieving a faster runtime than other global registration methods while maintaining accuracy.

While some of these methods are sufficiently efficient for onboard computation, planetary environments demand not only fast runtime but also reliable accuracy.
This is because registration errors directly propagate into inter-robot map distortion under resource-constrained operation.

\subsection{Consistency Check for False-Positive Rejection}
In multi-robot systems, erroneous measurements can cause perceptual aliasing, leading to map distortions that degrade the accuracy and consistency of mapping.
This motivates the need for robust techniques capable of detecting and eliminating false-positive links, or \textit{outliers}, to prevent severe distortions in multi-robot SLAM.
Therefore, these systems incorporate consistency checks to enhance the stability and reliability of mapping.

To this end, random sample consensus~(RANSAC)~\cite{fischler1981random} is a widely used approach that iteratively checks the geometric consistency of links to maintain robust map merging.
The graduated non-convexity~(GNC) method~\cite{yang2020graduated} approximates a solution without an initial guess and then incrementally increases non-convexity at each iteration.
This approach enhances robustness against local minima and improves efficiency by reducing the need for extensive iterations, making it well-suited for distributed mapping systems~\cite{tian2022kimera}.
The pairwise consistency maximization~(PCM) method~\cite{mangelson2018pairwise} evaluates pairwise consistency among inter-robot loop closure links to merge the multi-robot map robustly.
Each of these methods plays a critical role~\cite{huang2022disco} in ensuring robust multi-robot SLAM.
In particular, when applied in a two-stage process~(\eg RANSAC + PCM or GNC + PCM)~\cite{zhong2023dcl, tian2022kimera}, these methods contribute to reducing incorrect inter-robot loop closures.

To address these challenges, we design SKiD-SLAM, a robust, lightweight, and distributed multi-robot LiDAR SLAM system.
First, to enable minimal data exchange and rapid communication under the resource constraints of field robots, we employ SOLiD~\cite{kim2024narrowing}, one of the lightest global descriptors available at the time of this study.
Second, to handle cases in which large pose discrepancies between the viewpoints of two point clouds exceed the convergence range of ICP, we employ a two-stage registration approach consisting of global and local registration.
Finally, to alleviate map distortion, we identify and reject false-positive links among inter- and intra-robot measurements using a two-stage consistency check at the measurement and pose-graph levels.
 
\definecolor{myemerald}{rgb}{0.753, 0.898, 0.804}
\definecolor{mylightgreen}{rgb}{0.894, 0.933, 0.745}
\definecolor{myyellow}{rgb}{0.996, 0.972, 0.780}
\newcommand{\firstc}{\cellcolor{myemerald!100}}
\newcommand{\secondc}{\cellcolor{mylightgreen!100}}
\newcommand{\thirdc}{\cellcolor{myyellow!100}}
\newcommand{\oursdc}{\cellcolor{white!100}}

\section{Preliminaries}
\label{sec:preliminaries}

In this section, we formulate the objective function for our multi-robot LiDAR SLAM problem.
First, each robot estimates keyframe poses in its own local coordinate frame.
Note that our SKiD-SLAM is a front-end-agnostic approach; in this paper, we employ LIO-SAM~\cite{shan2020lio}, a factor graph-based LiDAR-inertial odometry consistent with our framework.

Next, let $\mathcal{X} = \{ \boldsymbol{x}_0, \cdots, \boldsymbol{x}_t \}$ be a set of these keyframe poses in $\text{SE(3)}$ from time 0 to time $t$.
The single robot SLAM problem can then be formulated as the following nonlinear least-squares optimization:
\begin{equation}
\mathcal{X}^* =  \argmin\limits_{\mathcal{X}} \mathcal{F}(\mathcal{X};\mathcal{C}).
\end{equation}
Here, the single robot pose-graph objective $\mathcal{F}$ is defined as follows:
\begin{equation}
\begin{aligned}
    \mathcal{F}(\mathcal{X}; & \mathcal{C}) =  \sum\limits_{\langle t, t+1 \rangle \in \mathcal{C}^{\text{odom}}}\left\|f\left(\boldsymbol{x}_t, \boldsymbol{x}_{t+1}\right)
                    - \hat{\boldsymbol{z}}_{t, t+1} \right\|_{\Omega_{t, t+1}}^2 \\
                  & + \sum_{\langle i, j \rangle \in \mathcal{C}^{\text{loop}}_{\text{intra}}}                
                      \left(\left\|f\left(\boldsymbol{x}_i, \boldsymbol{x}_j\right)
                    - \kappa(\mathcal{P}_i,\mathcal{P}_j) \right\|_{\Omega_{i, j}}^2 \right),
\end{aligned}
\label{equ:single_pgo}
\end{equation}
where $f(\cdot, \cdot)$ is a relative pose function that estimates the motion between two frames, \ie $f(\boldsymbol{x}_i, \boldsymbol{x}_j) \triangleq \boldsymbol{x}_i^{-1} \boldsymbol{x}_j$ is the pose $\boldsymbol{x}_j$ expressed in the frame of $\boldsymbol{x}_i$, $\hat{\boldsymbol{z}}$ is the odometry measurement, $\mathcal{C}^{\text{odom}}$ is an odometry constraint set between the consecutive $t$-th and $(t+1)$-th poses of a robot, $\mathcal{C}^{\text{loop}}_{\text{intra}}$ is a loop constraint set between two non-consecutive $i$-th and $j$-th poses of a robot~(\ie $|i-j|>1$), and $\kappa(\cdot, \cdot)$ is a relative pose estimator between two point clouds, $\mathcal{P}_i$ and $\mathcal{P}_j$, from non-consecutive frames.
By minimizing the Mahalanobis distance ($|| \cdot ||^2_{\Omega}$) weighted by the information matrices $\Omega_{t, t+1}$ and $\Omega_{i,j}$, we optimize the pose set so that the estimated and observed transformations are best aligned.
Here, we define the full constraint set as $\mathcal{C} = \mathcal{C}^{\text{odom}}\cup\mathcal{C}^{\text{loop}}_{\text{intra}}$.
The sets $\mathcal{C}^{\text{odom}}$ and $\mathcal{C}^{\text{loop}}_{\text{intra}}$ are disjoint, so $\mathcal{C}^{\text{odom}}$ $\cap$ $\mathcal{C}^{\text{loop}}_{\text{intra}}$ $= \varnothing$.

Next, we extend \eqref{equ:single_pgo}, which applies only to a single robot, to multi-robot systems.
Assume that we deploy $N$ robots, $\mathcal{N} = \{ 1, 2,\cdot\cdot\cdot,n, \cdot\cdot\cdot, N \}$.
Then, $\mathcal{X}_n$ represents the poses of the $n$-th robot, and $\mathcal{X}_{\text{total}} = \bigcup_{n \in \mathcal{N}} \mathcal{X}_n$ is the set of all keyframe poses from the $N$ robots.
The multi-robot SLAM problem can then be solved as follows:
\begin{equation}
    \mathcal{X}^*_{\text{total}} = \underset{ \mathcal{X}_{\text{total}}}{\arg\min} (\mathcal{F}_{\text{intra}} + \mathcal{F}_{\text{inter}}),
\label{equ:multi_pose}
\end{equation}
where $\mathcal{F}_{\text{intra}}$ represents the intra-robot objective across all robots and is defined as follows:
\begin{equation}
    \mathcal{F}_{\text{intra}} = \sum_{n \in \mathcal{N}} \mathcal{F}(\mathcal{X}_n; \mathcal{C}_n),
\label{equ:multi_intra}
\end{equation}
where $\mathcal{C}_n$ denotes the set of all constraints for the $n$-th robot, consisting of $\mathcal{C}_n^{\text{odom}}$ and $\mathcal{C}^{\text{loop}}_{n, \text{intra}}$ (\ie $\mathcal{C}_n = \mathcal{C}_n^{\text{odom}}\cup\mathcal{C}^{\text{loop}}_{n,\text{intra}}$).
Here, $\mathcal{F}_{\text{inter}}$ in \eqref{equ:multi_pose} represents the inter-robot loop-closure objective and is defined over cross-robot correspondences as follows:
\begin{equation}
    \mathcal{F}_{\text{inter}} =  \sum\limits_{\langle k, l \rangle \in \mathcal{C}_{\text{inter}}^{\text{loop}}}
                                  \left(\left\|f\left(\boldsymbol{x}_k^n, \boldsymbol{x}_l^m\right)
                                - \kappa(\mathcal{P}_k^n,\mathcal{P}_l^m) \right\|_{\Omega_{k, l}}^2 \right),
\label{equ:multi_inter}
\end{equation}
where $\mathcal{C}^{\text{loop}}_{\text{inter}}$ is the set of all inter-robot loop constraints, each connecting the $k$-th pose of the $n$-th robot and the $l$-th pose of the $m$-th robot~(\ie $\boldsymbol{x}_k^n \in \mathcal{X}_n$, $\boldsymbol{x}_l^m \in \mathcal{X}_m$, $m \in \mathcal{N} \backslash \{n\}$) with their corresponding point clouds, $\mathcal{P}_k^n, \mathcal{P}_l^m$, and $\Omega_{k, l}$ is the information matrix. 

\section{SKiD-SLAM: Resource-Aware Distributed Multi-Robot LiDAR SLAM}
\label{sec:method}

\begin{figure*}[t]
    \centering
    \def\width{\textwidth}%
        {%
        \includegraphics[clip, trim= 10 25 10 100, width=\width]{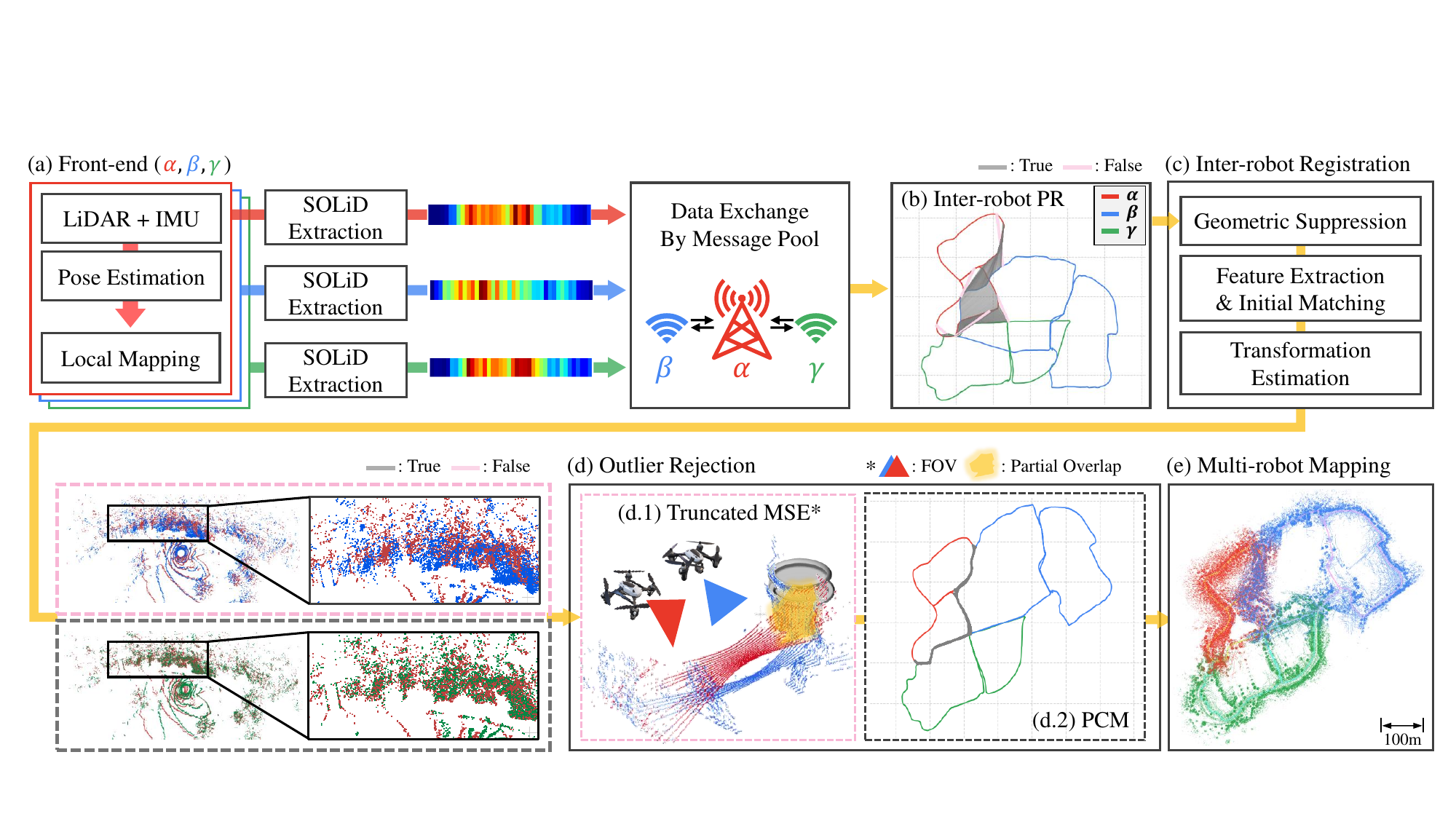}
        }
        \vspace{-0.3cm}
        \captionsetup{justification=justified}
        \caption{Overview of our distributed LiDAR SLAM framework, called \textit{SKiD-SLAM}.
        Building on \textit{SOLiD}~\cite{kim2024narrowing} and \textit{KISS-Matcher}~\cite{lim2024kiss}, our \textit{SKiD-SLAM} consists of five key components. The three robots are denoted by \textbf{\rl{$\alpha$}}, \textbf{$\beta$}, and \textbf{\gl{$\gamma$}}, respectively.
        In the distributed case, robot $\alpha$ receives shared data from robots $\beta$ and $\gamma$.
        (a) LIO-SAM-based pose estimation~\cite{shan2020lio} using LiDAR-IMU fusion.
        (b) Inter-Robot Place Recognition using \textit{SOLiD}~\cite{kim2024narrowing} descriptors from robot $\alpha$ and associated robots.
        (c) Coarse-to-fine inter-robot registration using \textit{KISS-Matcher} and Small-GICP~\cite{small_gicp} for coarse and fine alignment, respectively.
        (d) Two-stage outlier rejection, where truncated mean squared error~(MSE) is used to verify the validity of each loop individually, and PCM~\cite{mangelson2018pairwise} is used to verify the geometric consistency of loops across different trajectories.
        (e) Final distributed multi-robot mapping result centered on robot $\alpha$.}
        \label{fig:fig6_pipeline}
        \vspace{-0.3cm}
\end{figure*}

Our SKiD-SLAM is structured as shown in \figref{fig:fig6_pipeline}.
The framework consists of five main components: (i) a front-end comprising odometry estimation, SOLiD extraction, and local mapping, (ii) inter- and intra-robot place recognition using SOLiD descriptors shared through the message pool, (iii) inter- and intra-robot registration to estimate relative poses, (iv) outlier rejection to filter false links, and (v)~unified global mapping for the robot team.
Since both inter- and intra-robot place recognition and registration share the same module, this section explains the procedure from an inter-robot perspective.

\subsection{System Overview}
As the number of robots increases, centralized SLAM becomes computationally demanding on the central server~\cite{lajoie2022towards}. 
In particular, carrying a high-performance server into field environments is highly inefficient.
To address this, we adopt a distributed SLAM approach similar to~\cite{huang2022disco}.
For the distributed SLAM formulation, we reformulate \eqref{equ:multi_intra} and \eqref{equ:multi_inter} into an objective function for robot $\alpha$ as follows:
\begin{align}
     & \mathcal{X}_{\text{dis},\alpha}^{*}  = \underset{\mathcal{X}_{\text{dis},\alpha}}{\arg\min}
    \bigg(\mathcal{F}(\mathcal{X}_{\text{dis},\alpha}; \mathcal{C}_{\alpha}) \; + \notag \\
       &                                   \sum\limits_{\langle c, q \rangle \in \mathcal{C}^{\text{loop}}_{\text{dis,inter}}}
                                  \left(\left\|f\left(\boldsymbol{x}_c^\alpha, \boldsymbol{x}_q^\beta\right)
                                - \kappa(\mathcal{P}_c^\alpha,\mathcal{P}_q^\beta) \right\|_{\Omega_{c, q}}^2 \right)
    \bigg),
\label{equ:dis_pgo}
\end{align}
where $\mathcal{N}_\alpha = \{\alpha, r_1, r_2, \cdots \}$ is the set of indices of robots associated with robot $\alpha$ (\eg robots that can have links connected to robot $\alpha$ within 30\,m).
$\mathcal{C}^{\text{loop}}_{\text{dis,inter}} \subset \mathcal{C}^{\text{loop}}_{\text{inter}}$ is the set of inter-robot loop constraints, each connecting the $c$-th pose of robot $\alpha$ and the
$q$-th pose of robot $\beta$~(\ie $\boldsymbol{x}_c^\alpha \in \mathcal{X}_\alpha$, $\boldsymbol{x}_q^\beta \in \mathcal{X}_\beta$, $\beta \in \mathcal{N}_\alpha \backslash \{ \alpha \}$) with their 
corresponding point clouds, $\mathcal{P}_c^\alpha, \mathcal{P}_q^\beta$, and $\Omega_{c, q}$ is the information matrix. 
The designation of robot $\alpha$ is purely notational in the formulation, and any robot can assume this role.

Here, $\mathcal{X}_{\text{dis},\alpha}$, which collects the poses that robot $\alpha$ optimizes, is defined as follows:
\begin{equation}
    \mathcal{X}_{\text{dis},\alpha} = \mathcal{X}_{\alpha} \cup 
    \left\{ \boldsymbol{x}_q^\beta \; \middle|
    \begin{aligned}
        & \; \exists \beta \in \mathcal{N}_\alpha \backslash \{ \alpha \}, \\
        & \; \boldsymbol{x}_q^\beta \in \mathcal{X}_{\beta}, \langle c, q \rangle \in \mathcal{C}^{\text{loop}}_{\text{dis,inter}}
    \end{aligned}
    \right\}.
\end{equation}

Unlike \eqref{equ:multi_pose}, which optimizes over all $N$ robots, the distributed objective in \eqref{equ:dis_pgo} optimizes the poses of robot $\alpha$ and only those poses of its associated robots that form inter-robot loops with robot $\alpha$.

\subsection{Inter-Robot Place Recognition}
Inter-robot place recognition determines the candidates for $\mathcal{C}^{\text{loop}}_{\text{dis,inter}}$ in \eqref{equ:dis_pgo} by testing whether two robots, $\alpha$ and $\beta$, have observed the same place (\eg the $c$-th place observed by robot $\alpha$ and the $q$-th place observed by robot $\beta$).
To solve this problem, we use SOLiD~\cite{kim2024narrowing}, a lightweight, robust, and rotation-invariant descriptor.
SOLiD maps their respective point clouds, $\mathcal{P}_c^\alpha$ and $\mathcal{P}_q^\beta$, into global descriptors ($\mathbf{D}_{c}^\alpha = \omega(\mathcal{P}_{c}^\alpha) \in \mathbb{R}^e$, $\mathbf{D}_q^\beta = \omega(\mathcal{P}_q^\beta)\in \mathbb{R}^e$), where $\omega(\cdot)$ is an embedding function and $e$ is the embedding dimension.
We then measure the descriptor distance $d$ as follows:
\begin{equation}
   d = \phi (\mathbf{D}_c^{\alpha}, \mathbf{D}_q^\beta),
\label{equ:solid}
\end{equation}
where $\phi(\cdot, \cdot)$ is a Euclidean distance function, and we accept the pair $\langle c, q \rangle$ as a candidate element of $\mathcal{C}^{\text{loop}}_{\text{dis,inter}}$ only when $d < \tau_\text{dist}$.
Pairs that do not satisfy this threshold are rejected.

\begin{figure}[t]
	\centering
	\def\width{0.49\textwidth}%
    {%
    \includegraphics[clip, trim= 0 60 0 90, width=\width]{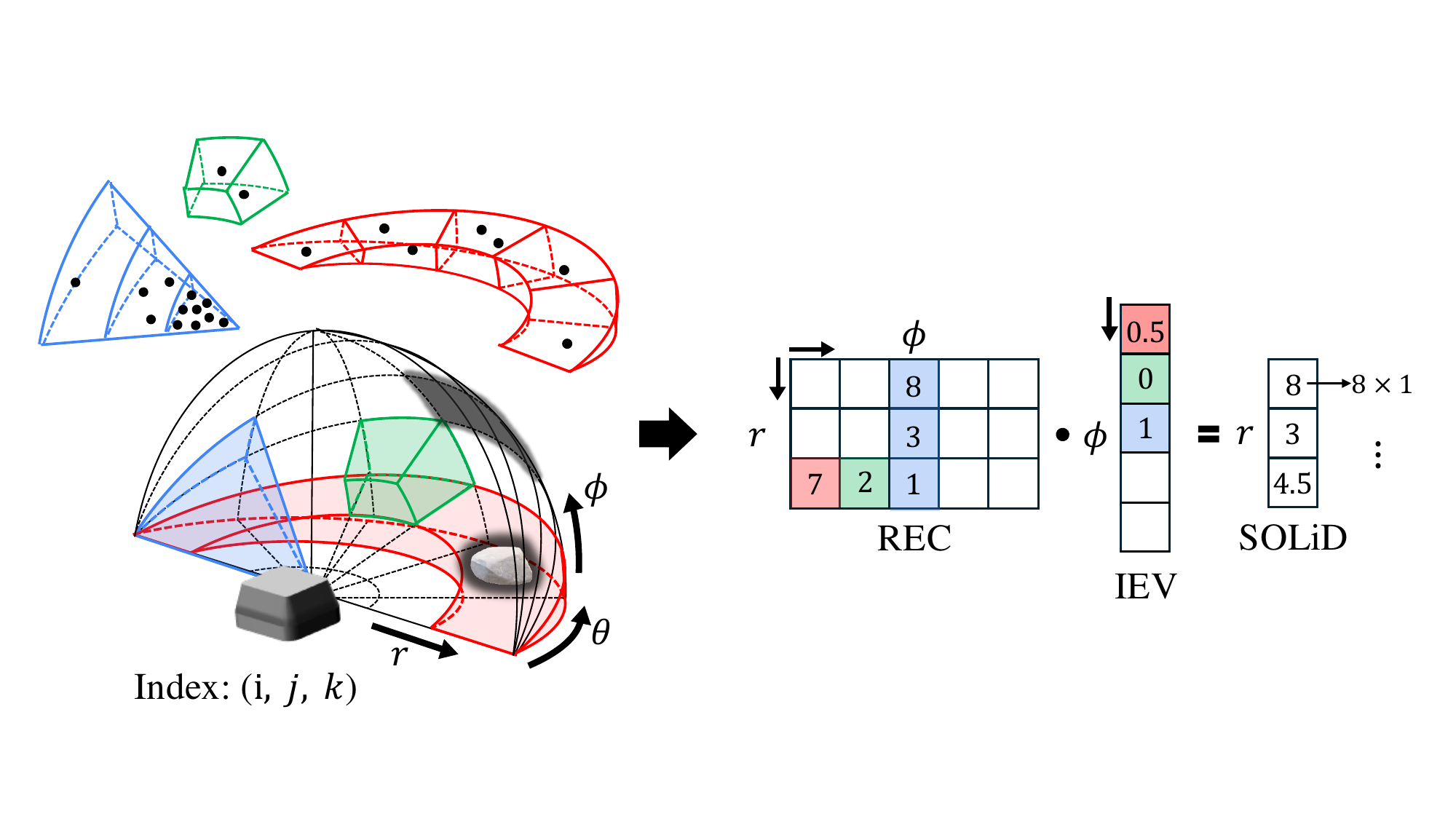}
	}
    \vspace{-0.4cm}
    \caption{Overview of SOLiD descriptor construction~\cite{kim2024narrowing}.
             Assuming the green-shaded regions represent occlusion or erroneous measurements caused by reflections of laser rays, directly using raw bin counts in such regions could lead to perceptual aliasing, whereas SOLiD mitigates this through the dot product between REC and IEV.
            }
    \label{fig:fig7_solid}
    \vspace{-0.6cm}
\end{figure}

Using SOLiD offers two practical advantages for inter-robot place recognition.
First, SOLiD is applicable across a range of fields of view~(FOVs) from 60$^\circ$ to 360$^\circ$ and remains effective even under occlusion conditions.
Second, SOLiD is less sensitive than descriptors that rely directly on raw geometric or intensity features in field environments.
For instance, SOLiD captures small but important structures such as the rock in \figref{fig:fig7_solid}, and maintains robust performance even in the presence of erroneous measurements.
In contrast, Scan Context~\cite{kim2018scan}, which uses a geometric feature~(\ie the elevation value of each bin) as the representative value in polar coordinate space, can overlook such structures.
Additionally, in environments such as caves where the intensity values of individual points exhibit little variation~\cite{novakova2022correcting}, extracting distinctive intensity features becomes challenging.

Following Kim and Kim~\cite{kim2018scan}, we adopt the distance function $\phi(\cdot, \cdot)$ in \eqref{equ:solid} using a Euclidean metric, and employ a KD-tree as an indexing structure for efficient nearest-neighbor search, enabling faster correspondence retrieval.

\subsection{Inter-Robot Registration}
Inter-robot registration uses $\kappa(\cdot, \cdot)$ in \eqref{equ:dis_pgo} to estimate the relative pose between $\mathcal{P}_c^\alpha$ and $\mathcal{P}_q^\beta$ for the same place observed by robots $\alpha$ and $\beta$ in $\mathcal{C}^{\text{loop}}_{\text{dis,inter}}$.
To estimate the relative pose, we use KISS-Matcher~\cite{lim2024kiss}, a robust global registration method.
For the matched scan pair $\langle c,q\rangle$, let $\mathbf{p}_a^\alpha \in \mathcal{P}_c^\alpha$ and $\mathbf{p}_b^\beta \in \mathcal{P}_q^\beta$ denote a pair of corresponding 3D points indexed by $\langle a,b\rangle \in \mathcal{A}$.
Then, $\kappa(\mathcal{P}_c^\alpha, \mathcal{P}_q^\beta)$ can be expressed as follows:
\begin{equation}
    \hat{\mathbf{R}}, \hat{\mathbf{t}} = \argmin\limits_{\mathbf{R}, \mathbf{t}}\sum\limits_{\langle a, b \rangle \in \mathcal{A} \backslash \hat{\mathcal{O}}}  \rho \bigg(\left\|(\mathbf{p}_a^\alpha - \mathbf{R}\mathbf{p}_b^\beta) - \mathbf{t} \right\|_2 \bigg).
\label{equ:kiss}
\end{equation}
Here, $\mathbf{R} \in \text{SO(3)}$ and $\mathbf{t} \in \mathbb{R}^3$ are the relative rotation matrix and translation vector, respectively.
The estimates $\hat{\mathbf{R}}$ and $\hat{\mathbf{t}}$ bring the point cloud of robot $\beta$ into the frame of robot $\alpha$, consistent with $f(\boldsymbol{x}_c^\alpha, \boldsymbol{x}_q^\beta)$ in \eqref{equ:dis_pgo}.
Furthermore, $\rho(\cdot)$ is a robust kernel for suppressing large errors caused by false correspondences, $||\cdot||_2$ is the Euclidean norm, $\mathcal{A}$ is an initial correspondence set, and $\hat{\mathcal{O}}$ denotes a set of estimated false correspondences.

As reported by \citet{lim2024kiss}, in the presence of a large pose discrepancy, ICP-based local registration may fail to converge to the optimal solution as viewpoint differences grow.
However, once a reliable initial guess for the relative pose is available as a coarse alignment, local registration can reliably converge to an accurate solution as a fine alignment.
For example, \citet{kim2018scan} demonstrate that a descriptor-derived initial guess can make ICP effective in reverse loops, such as those encountered at intersections.
Similarly, our global registration module estimates an accurate initial guess before local refinement.

This coarse registration stage provides two key benefits.
First, we reduce false correspondences that arise from repeating patterns in the environment~\cite{yang2020teaser} by applying geometric suppression.
This allows us to perform a sanity check on the reliability of the relative pose by verifying the cardinality of $\mathcal{A} \backslash \hat{\mathcal{O}}$.
Second, the method keeps the registration pipeline scalable by combining Faster-Point Feature Histograms~(Faster-PFH), a more efficient alternative to Fast Point Feature Histograms~(FPFH)~\cite{rusu2009fast}, with k-core-based graph-theoretic pruning that replaces the maximum-clique bottleneck with a linear-time pruning step.
This design allows us to solve the objective in \eqref{equ:kiss} on submaps without a substantial runtime increase, and then refine $\mathbf{R}$ and $\mathbf{t}$ toward the optimum through fine alignment with Small-GICP~\cite{small_gicp}.

\subsection{Inter-Robot Outlier Rejection}
In field environments, erroneous measurements caused by laser ray reflections or structural repetitiveness can produce false-positive links in $\mathcal{C}^{\text{loop}}_{\text{dis,inter}}$, leading to severe map distortions. To address this, we employ a two-stage outlier rejection process: (i)~truncated mean squared error~(MSE) and (ii)~pairwise consistency maximization~(PCM)~\cite{mangelson2018pairwise}.
In the following, we describe each stage in detail.

\subsubsection{Truncated MSE}
First, we reject outlier inter-robot associations based on a fitness score $s$, which evaluates the geometric alignment quality between the matched point clouds $\mathcal{P}_{c}^\alpha$ and $\hat{\mathcal{P}}_{q}^\beta$ for geometric verification.
Here, $\hat{\mathcal{P}}_{q}^\beta \triangleq \{ \hat{\mathbf{R}} \mathbf{p}_b^\beta + \hat{\mathbf{t}} \mid \mathbf{p}_b^\beta \in \mathcal{P}_{q}^\beta \}$ denotes the point cloud of robot $\beta$ registered into the frame of $\mathcal{P}_{c}^\alpha$ using the relative pose $\hat{\mathbf{R}}$ and $\hat{\mathbf{t}}$ estimated in \eqref{equ:kiss}.
In field deployments, robots may traverse the same area in opposite or partially reversed directions, resulting in limited overlap between point clouds. 
In such cases, a small number of large residual correspondences can dominate the score and cause the MSE to be skewed by a few large residuals.

To mitigate this effect, we apply a truncated MSE that suppresses the influence of large residuals, focusing the evaluation on the overlapping region. 
Let $\mathcal{M}$ be the set of correspondences obtained by nearest-neighbor matching between $\mathcal{P}_{c}^\alpha$ and $\hat{\mathcal{P}}_{q}^\beta$.
Let $d_{\lambda} \triangleq \|\mathbf{p}_{{\lambda}}^\alpha - \hat{\mathbf{p}}_{{\lambda}}^\beta\|_2^2$ denote the squared Euclidean distance between the $\lambda$-th matched point pair $(\mathbf{p}_{{\lambda}}^\alpha \in \mathcal{P}_{c}^\alpha, \hat{\mathbf{p}}_{{\lambda}}^\beta \in \hat{\mathcal{P}}_{q}^\beta)$ for $\lambda = 1, \cdots, |\mathcal{M}|$.
The fitness score is then computed as follows:
\begin{equation}
    s =
    \frac{1}{N_{\text{inliers}}}
    \sum_{\lambda = 1}^{|\mathcal{M}|}
    [\![ d_{\lambda} < \tau_{\text{MSE}} ]\!] \, d_{\lambda} .
\end{equation}
Here, $N_{\text{inliers}}$ is the number of inlier correspondences that satisfy $d_{\lambda} < \tau_{\text{MSE}}$.
$[\![\cdot]\!]$ denotes the Iverson bracket that equals $1$ if the condition is true and $0$ otherwise, and $\tau_{\text{MSE}}$ is a user-defined threshold on the squared distance.
We then accept the candidate association $\langle c, q \rangle$ only if it retains enough truncated inliers and attains a sufficiently small score, \ie $N_{\text{inliers}} \geq N_{\text{min}}$ and $s < \tau_{s}$.
In this criterion, $N_{\text{min}}$ is the minimum number of inlier correspondences required for the score to be meaningful, which also excludes the degenerate case of $N_{\text{inliers}} = 0$, and $\tau_{s}$ is a user-defined threshold on the score.

By truncating large correspondence distances at the measurement level, the proposed score emphasizes the geometrically consistent overlapping region and reduces sensitivity to large residuals caused by limited overlap during opposite or partially reversed traversals.
This robustness complements the subsequent pose-level outlier rejection.

\subsubsection{Pairwise Consistency Maximization}
Next, let $\boldsymbol{x}_{c_1},\boldsymbol{x}_{c_2} \in \mathcal{X}_\alpha$ and $\boldsymbol{x}_{q_1},\boldsymbol{x}_{q_2} \in \mathcal{X}_\beta$ be the $c_1$-th and $c_2$-th poses of robot $\alpha$ and the $q_1$-th and $q_2$-th poses of robot $\beta$.
Then, we derive intra-robot relative measurements, $\boldsymbol{z}_{c_1 c_2}$ and $\boldsymbol{z}_{q_2 q_1}$, from the respective robots' odometry constraints in $\mathcal{C}_{\alpha}$ and $\mathcal{C}_{\beta}$ in \eqref{equ:multi_intra}, and inter-robot measurements, $\boldsymbol{z}_{c_1 q_1}$ and $\boldsymbol{z}_{c_2 q_2}$, from $\hat{\mathbf{R}}$ and $\hat{\mathbf{t}}$ in~\eqref{equ:kiss}.
Based on these measurements, we apply PCM to reject loop closure outliers using the consistency residual.
Let $\Delta \triangleq(\ominus \boldsymbol{z}_{c_1 q_1}) \oplus \boldsymbol{z}_{c_1 c_2} \oplus \boldsymbol{z}_{c_2 q_2} \oplus \boldsymbol{z}_{q_2 q_1}$ denote the consistency transform, and let $\hat{\mathbf{t}}_{\Delta}$ denote its translation component.
We then define the pairwise consistency residual $p$ and its indicator $\delta$ as follows:
\begin{equation}
p \triangleq \|\hat{\mathbf{t}}_{\Delta}\|_2,
\qquad
\delta \triangleq [\![ p < \tau_{\mathrm{PCM}} ]\!].
\end{equation}
Here, $\oplus$ and $\ominus$ denote pose composition and inversion following~\cite{smith1990estimating}, $[\![\cdot]\!]$ denotes the Iverson bracket, and $\tau_{\mathrm{PCM}}$ is a user-defined threshold.
We regard two candidate inter-robot loop closures as pairwise consistent only if $\delta=1$; otherwise, the pair is deemed inconsistent.
After performing the PCM step by solving a maximum clique problem, we accept the selected loop closures as final loop constraints for $\mathcal{C}^{\text{loop}}_{\text{dis,inter}}$.

In sum, by combining measurement-level verification with a pose-level consistency check, our two-stage outlier rejection enables reliable inter-robot loop closure in field environments, where erroneous measurements and large viewpoint discrepancies are prevalent.

\subsection{Distributed Pose Graph Optimization}
Finally, after the inter-robot loop constraints in $\mathcal{C}_{\text{dis,inter}}^{\text{loop}}$ are verified through two-stage outlier rejection, robot $\alpha$ optimizes the distributed objective in \eqref{equ:dis_pgo}.
As reported by \citet{huang2022disco}, the rotation component lies in the non-convex space SO(3), so directly optimizing \eqref{equ:dis_pgo} with the poor initial guesses that typify inter-robot loop closures may converge to a local minimum.
To address this, we adopt the two-stage global-local optimization~\cite{huang2022disco}.

In the global step, robot $\alpha$ estimates only the coordinate transformations between its own frame and those of its associated robots $r \in \mathcal{N}_\alpha \backslash \{ \alpha \}$.
For each verified inter-robot loop in $\mathcal{C}_{\text{dis,inter}}^{\text{loop}}$, the relative pose from \eqref{equ:kiss} yields an estimate of the frame-to-frame transformation $\hat{\mathbf{T}}_{\alpha r} = \begin{bmatrix} \hat{\mathbf{R}}_{\alpha r} & \hat{\mathbf{t}}_{\alpha r} \\ \mathbf{0}^\top & 1 \end{bmatrix} \in \text{SE(3)}$, and a graph containing only these transformations is optimized to recover a consistent global alignment across $\mathcal{N}_\alpha$.
Because this graph involves only inter-robot transformations rather than full trajectories, it provides a stable and high-quality initial guess that is resilient to the unknown initial conditions of inter-robot loop closures.

In the local step, the separator poses of the associated robots, \ie the poses that form inter-robot loops with robot $\alpha$, are transformed into robot $\alpha$'s frame using the estimated $\hat{\mathbf{T}}_{\alpha r}$, and each inter-robot constraint is converted into a virtual intra-robot constraint.
Robot $\alpha$ then optimizes its local pose graph in $\mathcal{X}_{\text{dis},\alpha}$, jointly incorporating odometry constraints, intra-robot loop-closure constraints, and converted inter-robot constraints in a single unified step using iSAM2~\cite{kaess2012isam2} in GTSAM~\cite{dellaert2012factor}.

This two-stage scheme decouples the global alignment from the local refinement, ensuring that robot $\alpha$ resolves only minor numerical changes from odometry drift during local optimization, which improves the numerical stability and convergence of the distributed estimate under the limited onboard resources of field robots.

\section{Preliminary Evaluation}
\label{sec:pre_evaluation}

Before leaving the lab for the field, we conducted a series of controlled experiments to determine which place recognition and registration methods to deploy onboard.
Rather than assuming a particular choice, we evaluated several candidate methods and selected SOLiD~\cite{kim2024narrowing} and KISS-Matcher~\cite{lim2024kiss} based on their performance under conditions resembling our target field environments.
This section describes that validation process.

\subsection{Datasets}
To make this comparison meaningful, we matched the evaluation conditions to the two environments targeted by our field campaign: a subterranean cave and cratered planetary analog terrain.
For the cave setting, which reflects the Majang lava-tube cave, we used the publicly available GEODE dataset~\cite{chen2026heterogeneous}, whose underground sequences capture characteristics similar to our target site.
For the surface setting, however, no public dataset provides cratered planetary terrain with the multi-robot ground truth needed for our evaluation.
Therefore, we built a custom simulation environment for this scenario.

\begin{figure}[t]
    \centering
	\def\width{0.49\textwidth}%
    {%
    \includegraphics[clip, trim= 0 0 0 0, width=\width]{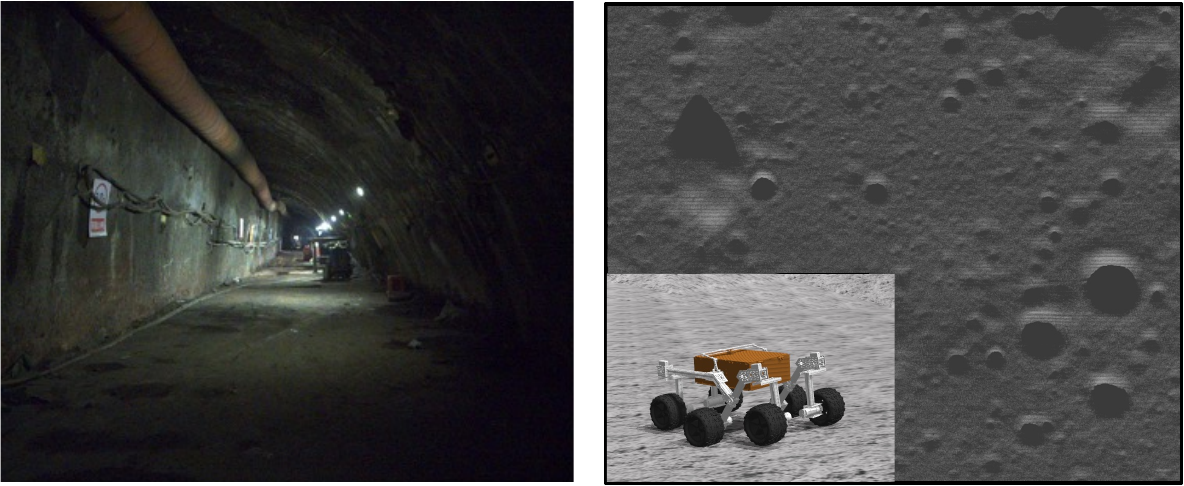}
	}
    \vspace{-0.4cm}
    \caption{Preliminary-evaluation environments matched to our target mission settings.
             (Left)~The \texttt{Tunnel} sequence from the GEODE dataset, representing the subterranean cave setting.
             (Right)~Our custom Gazebo-based \texttt{Moon} simulator built on NASA JPL's Open Source Rover, representing cratered planetary analog terrain; the inset shows the simulated rover.
            }
    \label{fig:fig8_pre_eval}
\end{figure}

To keep the simulated platform faithful to real planetary rovers, we built our rovers on NASA JPL's Open Source Rover~(OSR) project\footnote{\url{https://github.com/nasa-jpl/osr-rover-code}}, an openly released design that mirrors the rocker-bogie suspension and corner-steering kinematics of JPL's Mars rovers.
We integrated the OSR platform into a custom Gazebo-based multi-robot simulator\footnote{\url{https://github.com/dongjineee/rover_gazebo}}, where a team of such rovers traverses a crater-strewn lunar surface reconstructed from Gazebo's Moon world\footnote{\url{https://github.com/er1c-cartman/benchmark_lunar}} while running SKiD-SLAM onboard.
As shown in \figref{fig:fig8_pre_eval}, this setup allowed us to evaluate the full distributed pipeline, \ie inter-robot place recognition, relative pose estimation, and trajectory fusion, against known ground truth before committing to a field campaign.

\subsection{Evaluation Metrics}
We evaluate SKiD-SLAM using the metrics defined below, which jointly assess place recognition, registration, multi-robot alignment, and resource efficiency under the conditions described above.

\subsubsection{PR Curve}
To evaluate inter-robot place recognition, we use the precision-recall~(PR) curve, where precision and recall are defined as follows:
\begin{equation}
\text{Precision} = \frac{\mathrm{TP}}{\mathrm{TP}+\mathrm{FP}}, \quad \text{Recall} = \frac{\mathrm{TP}}{\mathrm{TP}+\mathrm{FN}},
\end{equation}
where TP, FP, and FN are true positives, false positives, and false negatives, respectively.

\subsubsection{Resource Efficiency}
To evaluate descriptor-level communication scalability under long-term and large-scale operation, we measure the total memory footprint of the shared global descriptors during multi-robot map merging as follows:
\begin{equation}
M_{\text{total}} = \sum_{r \in \mathcal{N}_\alpha \backslash \{ \alpha \}} \sum_{k=1}^{N_{r}} c(\mathbf{D}^r_k),
\label{equ:descriptor_memory}
\end{equation}
where $r \in \mathcal{N}_\alpha \backslash \{ \alpha \}$ indexes the robots participating in collaborative mapping, $N_r$ is the number of global descriptors shared by robot $r$, $\mathbf{D}^r_k$ is the $k$-th descriptor shared by robot $r$, and $c(\cdot)$ denotes the memory footprint, in bytes, of a single descriptor.
Note that this metric intentionally isolates descriptor exchange, which occurs repeatedly for place recognition.
The full communication burden also includes scan exchange for registration, so we evaluate the integrated system-level effect of both descriptor and scan exchange under bandwidth limits in \Cref{sec:field_eval:communication_budget}.

\subsubsection{RTE, RRE, and Success Rate}
To evaluate inter-robot registration, we use the success rate as the key performance metric.
A registration attempt is considered successful if its translation and rotation errors are below 2\,m and 5$^\circ$, respectively.
The success criteria were set based on the established convention in LiDAR registration literature~\cite{lim2024kiss, zhang2026deep}, where RTE $<$ 2\,m and RRE $<$ 5$^\circ$ are widely adopted thresholds for successful registration.

The relative translation error (RTE) and relative rotation error (RRE) are defined as follows:
\begin{itemize}
    \item RTE $= \frac{1}{N_{\text{success}}} \sum_{\varpi=1}^{N_{\text{success}}} \left\| \mathbf{t}_{\varpi,\text{GT}} - \hat{\mathbf{t}}_\varpi \right\|_2$,
    \item RRE $= \frac{180}{\pi N_{\text{success}}} \sum_{\varpi=1}^{N_{\text{success}}} \left| \cos^{-1}\!\left(\frac{\text{Tr}(\hat{\mathbf{R}}_\varpi^\top \mathbf{R}_{\varpi,\text{GT}}) - 1}{2}\right) \right|$,
    \item Success Rate $= \frac{N_{\text{success}}}{N_{\text{total}}} \times 100\,\%$,
\end{itemize}
where $N_{\text{success}}$ is the number of attempts whose errors fall below the acceptable thresholds (\ie RTE $< 2\,\text{m}$ and RRE $< 5^\circ$), $N_{\text{total}}$ is the total number of registration attempts, $\mathbf{t}_{\varpi,\text{GT}}$ and $\hat{\mathbf{t}}_\varpi$ are the ground truth and estimated translations, and $\mathbf{R}_{\varpi,\text{GT}}$ and $\hat{\mathbf{R}}_\varpi$ are the ground truth and estimated rotations.

To construct the evaluation pairs, we follow the loop-closing benchmark of~\cite{lim2024quatro++}.
The $A \sim B$ test samples inter-robot source-target pairs whose ground truth pose distance lies between $A$ and $B$, with a sufficiently large frame difference to exclude near-identical observations; further details are provided by \citet{lim2024quatro++}.
In our case, we set $A = 2$\,m and $B = 5$\,m.

\subsubsection{Multi-Robot Alignment Error}
To evaluate the accuracy of robot $\alpha$ against ground truth, we estimate the transformation $\mathbf{T}^{\text{GT}}_{\alpha}$ between the ground truth and the estimated pose of robot $\alpha$ using the Kabsch-Umeyama algorithm~\cite{umeyama1991least}, and assess the absolute pose error~(APE), $\hat{\mathbf{T}}_{\alpha,\text{APE}}$.
Evaluating robot $\beta$ with the same procedure would assess only its trajectory accuracy, which is inadequate for a comprehensive alignment evaluation.
Therefore, to evaluate the alignment of robot $\beta$ with respect to robot $\alpha$, we use the following alignment metric:
\begin{equation}
    \hat{\mathbf{T}}_{\beta,\text{APE}} = (\mathbf{T}^{\text{GT}}_{\beta})^{-1} \, \mathbf{T}^{\text{GT}}_{\alpha} \, \hat{\mathbf{T}}^{\alpha}_{\beta},
\end{equation}
where $\mathbf{T}^{\text{GT}}_{\beta}$ is the transformation between the ground truth and the estimated pose of robot $\beta$ using~\cite{umeyama1991least}, and $\hat{\mathbf{T}}^{\alpha}_{\beta}$ is the estimated transformation of robot $\beta$ relative to robot $\alpha$ from the distributed pose graph optimization back-end.
The APE is computed using the EVO trajectory evaluation tool~\cite{grupp2017evo}.

For clarity, we report the results in three ranks: {\setlength{\fboxsep}{1pt}\colorbox{myemerald}{\textbf{1st}}}, {\setlength{\fboxsep}{1pt}\colorbox{mylightgreen}{2nd}}, and {\setlength{\fboxsep}{1pt}\colorbox{myyellow}{3rd}}.

\subsection{Why SOLiD for Place Recognition?}
\begin{figure}[t]
	\centering
	\def\width{0.49\textwidth}%
        {%
		\includegraphics[clip, trim= 0 0 0 0, width=\width]{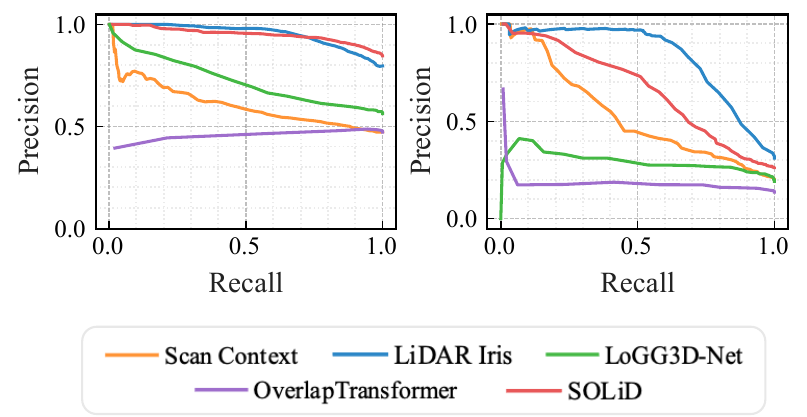}
	}
    \vspace{-0.3cm}
    \caption{Precision-recall~(PR) curves for inter-robot place recognition on the (left)~\texttt{Tunnel} and (right)~\texttt{Moon} sequences, corresponding to the cave and cratered surface settings.
            }
    \label{fig:fig9_pr_curve}
    \vspace{-0.3cm}
\end{figure}

For inter-robot place recognition, the key tradeoff is not accuracy alone but accuracy under a strict descriptor-exchange budget.
\figref{fig:fig9_pr_curve} reports the PR curves on the \texttt{Tunnel} and \texttt{Moon} sequences, which correspond to our cave and cratered surface settings, respectively.
On the \texttt{Tunnel} sequence, SOLiD is on par with LiDAR Iris, with both sustaining high precision across the full recall range, while the other descriptors degrade earlier.
On the \texttt{Moon} sequence, LiDAR Iris attains the highest precision and SOLiD follows closely, remaining well above Scan Context, OverlapTransformer, and LoGG3D-Net.
Overall, SOLiD is comparable to the best-performing descriptor in recognition accuracy across both target environments.

\begin{table*}[t]
\captionsetup{width=0.99\textwidth, justification=justified} 
\caption{Comparison of total descriptor memory usage, $M_{\text{total}}$ in \eqref{equ:descriptor_memory}, shared during multi-robot map merging on the \texttt{Tunnel} and \texttt{Moon} sequences.}
\setlength{\tabcolsep}{12pt} 
\centering\resizebox{0.98\textwidth}{!}{\tiny
\begin{tabular}{l|ccccc}
\toprule \midrule
\textbf{Sequence}    & \textbf{Scan Context}~\cite{kim2018scan}    & \textbf{LiDAR Iris}~\cite{wang2020lidar} & \textbf{OverlapTransformer}~\cite{ma2022overlaptransformer}    & \textbf{LoGG3D-Net}~\cite{vidanapathirana2022logg3d}    & \textbf{SOLiD}~\cite{kim2024narrowing} \\ \midrule
Descriptor size [B] $\downarrow$ & 9{,}600 & 230{,}400 & \secondc 2{,}048 & \secondc 2{,}048                        & \firstc \textbf{320} \\ \midrule
  \texttt{Tunnel}          & 48.506$\,$MB & 1.164$\,$GB & \secondc 10.347$\,$MB & \secondc 10.347$\,$MB  & \firstc \textbf{1.617}$\,$\textbf{MB} \\
  \texttt{Moon}            & 29.474$\,$MB & 707.379$\,$MB & \secondc 6.288$\,$MB & \secondc 6.288$\,$MB & \firstc  \textbf{0.982}$\,$\textbf{MB} \\
\midrule \bottomrule
\end{tabular}}
\label{tab:desc}
\end{table*}

However, when descriptor memory usage is also considered, \tabref{tab:desc} shows that SOLiD is particularly effective in communication-constrained environments.
Measured by $M_{\text{total}}$ in \eqref{equ:descriptor_memory}, SOLiD reduces the exchanged descriptor data by factors of approximately 30 relative to Scan Context, 720 relative to LiDAR Iris, and 6.4 relative to the learning-based descriptors.
Critically, the only descriptor that rivals SOLiD in accuracy, LiDAR Iris, imposes approximately 720 times more descriptor data per exchange, which would consume substantially more of a kilobit-class link's capacity.
Scan Context, OverlapTransformer, and LoGG3D-Net are lighter than LiDAR Iris, yet they trail substantially in accuracy on both sequences.

Therefore, among the compared descriptors, SOLiD provides the practical operating point for our target setting: sufficiently strong recognition performance for our field deployment, but with descriptor exchange reduced by factors ranging from 6.4 to 720 relative to the compared alternatives, as required by the communication budget defined in \Cref{sec:background:budgets}.
For this reason, we deploy SOLiD as the place recognition module of SKiD-SLAM.

\subsection{Why KISS-Matcher for Registration?}
Registration presents a different challenge: the method must tolerate large inter-robot viewpoint discrepancies before a reliable initial guess is available.
\tabref{tab:regi_eval} reports scan-level registration on the \texttt{Moon} sequence.
Local methods are accurate once they converge, yet they converge only within a narrow basin.
ICP and PLANE-ICP succeed on barely $2$ to $7\,\%$ of attempts, as their narrow convergence region cannot accommodate the viewpoint differences of inter-robot loops.
The GICP variants are accurate and improve markedly, reaching about $82$ to $83\,\%$ success, yet they still fail when the initial inter-robot discrepancy is large.
KISS-Matcher, a global method, raises the success rate to $98.7\,\%$ by searching over a wide pose space, although its coarse alignment alone leaves a larger residual error of $25.12\,$cm.
In contrast, pairing KISS-Matcher with Small-GICP in a coarse-to-fine manner attains the highest success rate of $99.7\,\%$ while matching the most accurate methods, with an RTE of $8.61\,$cm and an RRE of $0.223^\circ$, within $0.1\,$cm and $0.03^\circ$ of the best values in the table.
For inter-robot loops, our combination achieves a higher success rate than TEASER++ with Small-GICP while maintaining a comparable RTE, with a difference of only $0.07\,$cm between the two methods ($8.61\,$cm and $8.54\,$cm, respectively).
KISS-Matcher provides a global coarse alignment over a wide pose space, while Small-GICP refines it locally to recover the fine accuracy that global registration alone does not provide.

\begin{table}[b]
\vspace{-0.5cm}
\captionsetup{width=0.48\textwidth, justification=justified}
\caption{Scan-level registration performance evaluated in the \texttt{Moon} environment.}
\setlength{\tabcolsep}{10pt} 
\centering\resizebox{0.495\textwidth}{!}{
\begin{tabular}{l|ccc}
\toprule \midrule
\textbf{Method} & \textbf{RTE [cm] $\downarrow$}  & \textbf{RRE [$^\circ$] $\downarrow$} & \textbf{Success rate [\%] $\uparrow$} \\ \midrule
ICP \cite{rusinkiewicz2001efficient}    & 173.96                  & 2.183                          & 2.01                    \\
PLANE-ICP \cite{low2004linear}          & 136.58                  & 2.334                          & 6.61                    \\
GICP \cite{segal2009generalized}        & \thirdc 8.98            & \secondc 0.203                 & 83.33                   \\
Small-GICP \cite{small_gicp}               & \thirdc 8.98            & \secondc 0.203                 & 83.33                   \\
V-GICP \cite{koide2021voxelized}        & 9.24                    & \firstc $\boldsymbol{0.196}$   & 82.47                   \\
NDT \cite{takeuchi20063}                                  & N/A          & N/A                  & 0.00                   \\
NDT w/ Small-GICP \cite{takeuchi20063,small_gicp}        & 9.12            & 0.209                  & 83.48                   \\
TEASER++ \cite{yang2020teaser}       & 14.81                   & 0.753                          &  91.24                   \\
TEASER++ w/ Small-GICP                   & \firstc $\boldsymbol{8.54}$ & 0.217                      & \thirdc 92.24                   \\
KISS-Matcher \cite{lim2024kiss}        & 25.12                   & 1.002                          & \secondc 98.71                  \\
KISS-Matcher w/ Small-GICP              & \secondc 8.61           & 0.223                          & \firstc $\boldsymbol{99.71}$     \\  \midrule
\bottomrule
\end{tabular}}
\label{tab:regi_eval}
\end{table}

\begin{figure}[!t]
	\centering
	\def\width{0.49\textwidth}%
        {%
		\includegraphics[clip, trim= 0 0 0 0, width=\width]{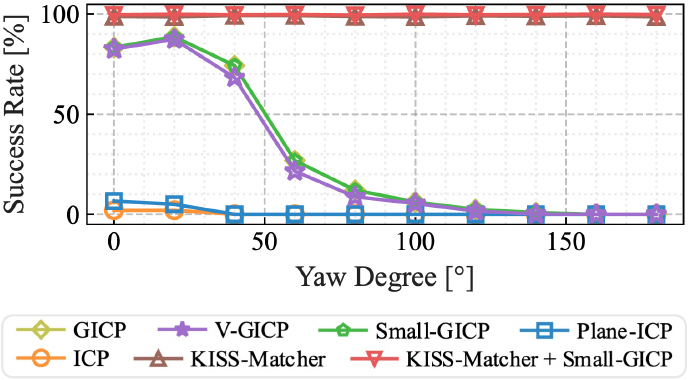}
	}
    \vspace{-0.3cm}
    \caption{Registration success rate on the \texttt{Moon} sequence under synthetic relative rotations.}
    \label{fig:fig10_success_rate}
    \vspace{-0.3cm}
\end{figure}

The advantage of this design becomes clear once the inter-robot viewpoint discrepancy grows.
Unlike intra-robot loop closures, inter-robot loop closures arise between robots that observe the same place from markedly different viewpoints, so the registration module must remain robust to large rotational discrepancies.
\figref{fig:fig10_success_rate} reports the success rate under synthetic relative rotations.
Local methods degrade sharply as the rotation grows, whereas KISS-Matcher searches over the full pose space and is therefore agnostic to the initial rotation, sustaining a high success rate where local methods collapse.

Robustness alone, however, is not sufficient under the planetary compute budget, since global registration is generally more demanding than local methods.
\tabref{tab:runtime_nano} therefore compares KISS-Matcher against TEASER++, a representative global registration method, on an Intel NUC 11 Pro.
Among the compared global methods that provide the required viewpoint robustness, KISS-Matcher runs fastest on this compact onboard computer and offers the best runtime--robustness tradeoff.
For this reason, we deploy KISS-Matcher, refined by Small-GICP, as the registration module of SKiD-SLAM.
\begin{table}[!t]
\vspace{-0.1cm}
\captionsetup{width=0.49\textwidth, justification=justified}
\caption{Runtime of point cloud registration methods on an Intel NUC 11 Pro evaluated on the \texttt{Moon} sequence.}
\setlength{\tabcolsep}{8pt} 
\centering\resizebox{0.495\textwidth}{!}{\tiny
\begin{tabular}{l|cccc}
\toprule \midrule
\multirow{2}{*}{\textbf{Method}} & \multicolumn{4}{c}{\textbf{Runtime [s]} $\downarrow$} \\ 
\cmidrule(lr){2-5}
 & Min & Mean & Median & Max \\ \midrule
ICP \cite{rusinkiewicz2001efficient}    & \firstc \textbf{0.002} & \secondc 0.005 & \firstc \textbf{0.004} & 0.093 \\
Plane-ICP \cite{low2004linear}          & \secondc 0.003 & \secondc 0.005 & \thirdc 0.005 & \thirdc 0.020 \\
GICP \cite{segal2009generalized}        & \secondc 0.003 & \secondc 0.005 & \thirdc 0.005 & \secondc 0.012 \\
Small-GICP \cite{small_gicp}            & \secondc 0.003 & \secondc 0.005 & \thirdc 0.005 & \thirdc 0.020 \\
VGICP \cite{koide2021voxelized}         & \secondc 0.003 & \firstc \textbf{0.004} & \firstc \textbf{0.004} & \firstc \textbf{0.009} \\
TEASER++ \cite{yang2020teaser}          & 0.446 & 0.673 & 0.640 & 1.335 \\
TEASER++ w/ Small-GICP                  & 0.460 & 0.683 & 0.649 & 1.345 \\
KISS-Matcher \cite{lim2024kiss}         & 0.016 & 0.047 & 0.054 & 0.098 \\
KISS-Matcher w/ Small-GICP              & 0.019 & 0.051 & 0.058 & 0.101 \\
\midrule \bottomrule
\end{tabular}}
\label{tab:runtime_nano}
\vspace{-0.2cm}
\end{table}

\begin{table*}[t!]
    \captionsetup{width=0.98\textwidth, justification=justified} 
    \caption{Comparison of absolute translation error~(ATE) and absolute rotation error~(ARE) using distributed LiDAR SLAM approaches.
             For the \texttt{Tunnel} sequence from the GEODE dataset, rotation ground truth is not provided; therefore, the ARE values are marked as N/A, indicating that we cannot evaluate them.}
    \centering\resizebox{0.99\textwidth}{!}{
    \begin{tabular}{l|cccccccc}
     \toprule \midrule
   \textbf{Datasets}    & \multicolumn{4}{c}{\texttt{Tunnel}}   & \multicolumn{4}{c}{\texttt{Moon}} \\ \cmidrule(lr){2-5}\cmidrule(lr){6-9}
    \textbf{Evaluation} & \multicolumn{2}{c}{\textbf{ATE [m]}}     & \multicolumn{2}{c}{\textbf{ARE [$^\circ$]}}      & \multicolumn{2}{c}{\textbf{ATE [m]}}    & \multicolumn{2}{c}{\textbf{ARE [$^\circ$]}} \\
    \cmidrule(lr){2-3}\cmidrule(lr){4-5}\cmidrule(lr){6-7}\cmidrule(lr){8-9}
    \textbf{Robot Name} & robot $\alpha$        & \multicolumn{1}{c}{robot $\beta$}           & robot $\alpha$                          & robot $\beta$     
                        & robot $\alpha$                   & \multicolumn{1}{c}{robot $\beta$}           & robot $\alpha$                          & robot $\beta$      \\ \midrule
                        
        DiSCo-SLAM \cite{huang2022disco} & \thirdc $0.752 \pm 0.029$ & \multicolumn{1}{c}{\thirdc $16.938 \pm 0.033$} & N/A & N/A & \secondc $0.137 \pm 0.008$ & \multicolumn{1}{c}{\thirdc $40.924 \pm 0.013$} & \firstc $\boldsymbol{0.316 \pm 0.018}$ & \secondc $2.380 \pm 0.051$ \\ \midrule
        DCL-SLAM \cite{zhong2023dcl} & \firstc $\boldsymbol{0.318 \pm 0.021}$ & \multicolumn{1}{c}{\secondc $5.551 \pm 0.041$} & N/A & N/A & \firstc $\boldsymbol{0.120 \pm 0.007}$ & \multicolumn{1}{c}{\secondc $1.851 \pm 0.013$} & \secondc $0.364 \pm 0.022$ & \thirdc $5.674 \pm 0.068$ \\ \midrule 
        SKiD-SLAM & \secondc $0.339 \pm 0.024$ & \multicolumn{1}{c}{\firstc $\boldsymbol{0.403 \pm 0.038}$} & N/A & N/A & \thirdc $0.198 \pm 0.010$ & \multicolumn{1}{c}{\firstc $\boldsymbol{1.046 \pm 0.014}$} & \thirdc $0.546 \pm 0.026$ & \firstc $\boldsymbol{0.080 \pm 0.006}$ \\ \midrule    \bottomrule
    \end{tabular}}
    \label{tab:slam_comparison}
    \vspace{-0.3cm}
\end{table*}

\subsection{Overall Performance of SKiD-SLAM}
\begin{figure}[t]
	\centering
	\def\width{0.49\textwidth}%
        {%
		\includegraphics[clip, trim= 0 0 0 0, width=\width]{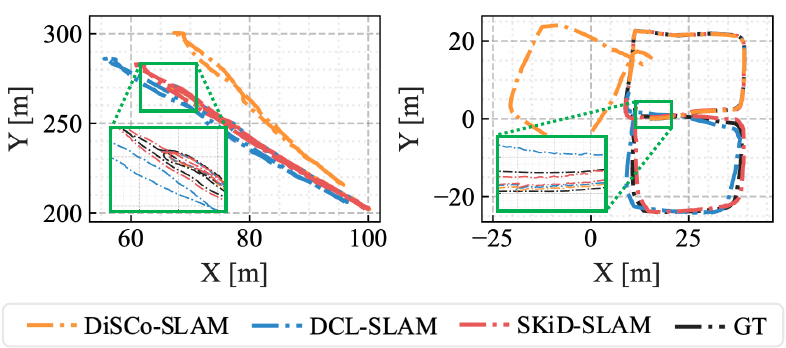}
	}
    \vspace{-0.3cm}
    \caption{Estimated trajectories from DiSCo-SLAM~\cite{huang2022disco}, DCL-SLAM~\cite{zhong2023dcl}, and SKiD-SLAM compared with ground truth on the \texttt{Tunnel} and \texttt{Moon} sequences.}
    \label{fig:fig11_traj}
    \vspace{-0.6cm}
\end{figure}

With SOLiD and KISS-Matcher selected for the place recognition and registration modules, we next assess the SKiD-SLAM system that integrates them.
\figref{fig:fig11_traj} and \tabref{tab:slam_comparison} show the quantitative results on the \texttt{Tunnel} and \texttt{Moon} sequences.

\begin{figure*}[t]
    \centering
    \captionsetup{justification=justified}
    \captionsetup[subfigure]{justification=centering}
    \begin{subfigure}{1.98\columnwidth}
        \includegraphics[trim={0 0 0 0},clip, width=\columnwidth]{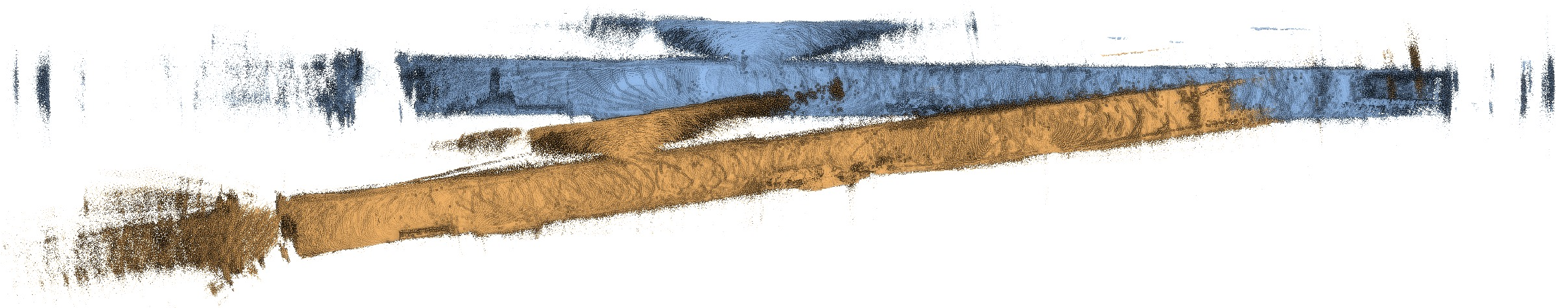}
        \caption{DiSCo-SLAM~\cite{huang2022disco}}
    \end{subfigure}
    \begin{subfigure}{1.98\columnwidth}
        \includegraphics[trim={0 0 0 0},clip, width=\columnwidth]{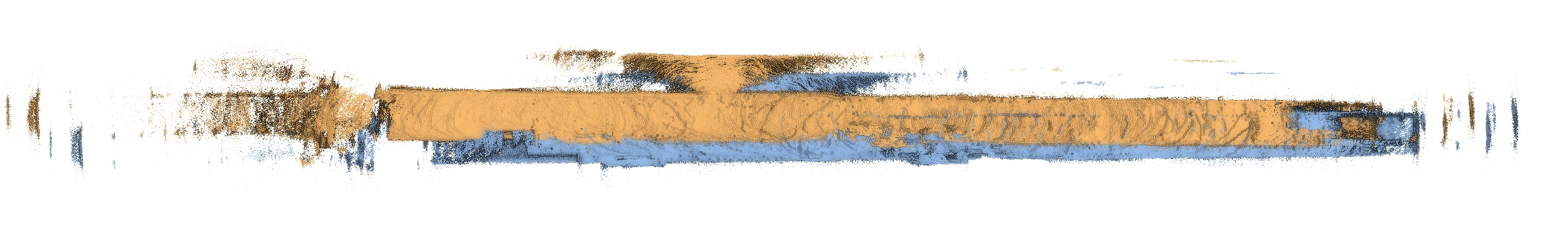}
        \caption{DCL-SLAM~\cite{zhong2023dcl}}
    \end{subfigure}
    \begin{subfigure}{1.98\columnwidth}
        \includegraphics[trim={0 0 0 0},clip, width=\columnwidth]{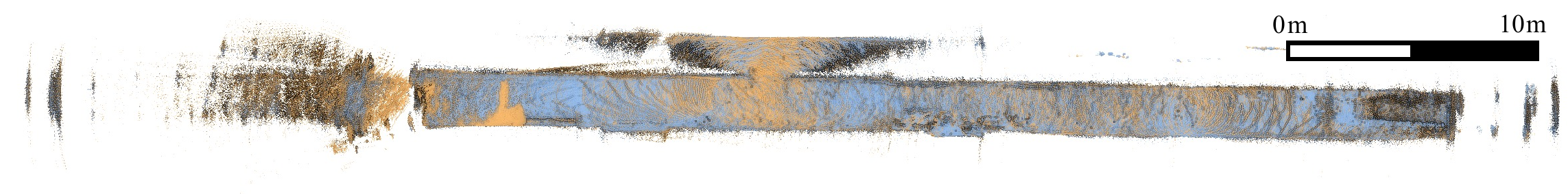}
        \caption{SKiD-SLAM}
    \end{subfigure}
    \caption{Qualitative multi-robot map merging results on the \texttt{Tunnel} sequence.
    (a)~DiSCo-SLAM and (b)~DCL-SLAM leave visible map misalignments, whereas (c)~SKiD-SLAM produces a consistent merged map.}
    \label{fig:fig12_cave1}
    \vspace{-0.4cm}
\end{figure*}

\begin{figure*}[t!]
    \centering
    \captionsetup{justification=justified}
    \captionsetup[subfigure]{justification=centering}
    \begin{subfigure}{0.66\columnwidth}
        \includegraphics[trim={0 0 0 0},clip, width=\columnwidth]{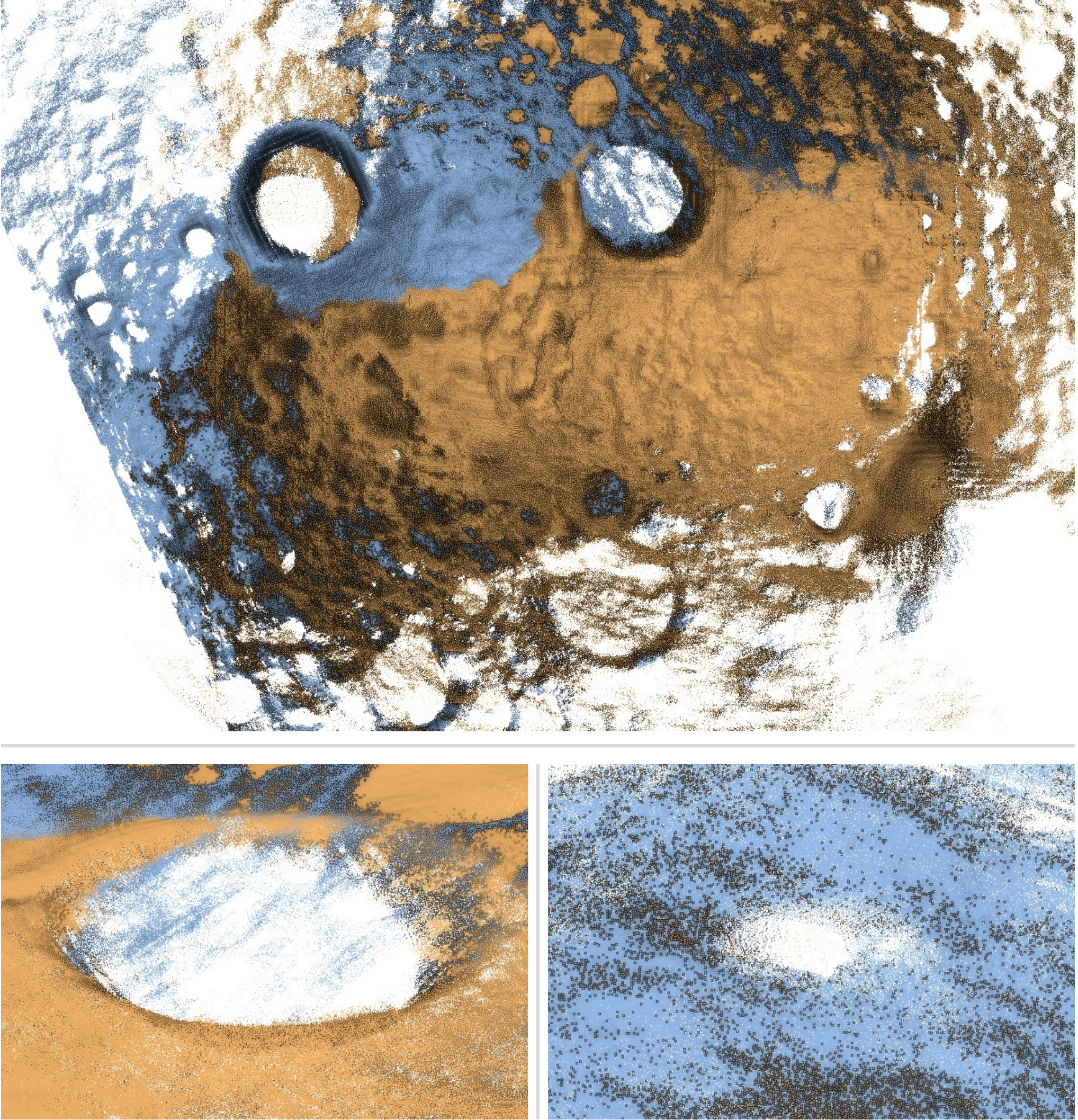}
        \caption{DiSCo-SLAM~\cite{huang2022disco}}
    \end{subfigure}
    \begin{subfigure}{0.66\columnwidth}
        \includegraphics[trim={0 0 0 0},clip, width=\columnwidth]{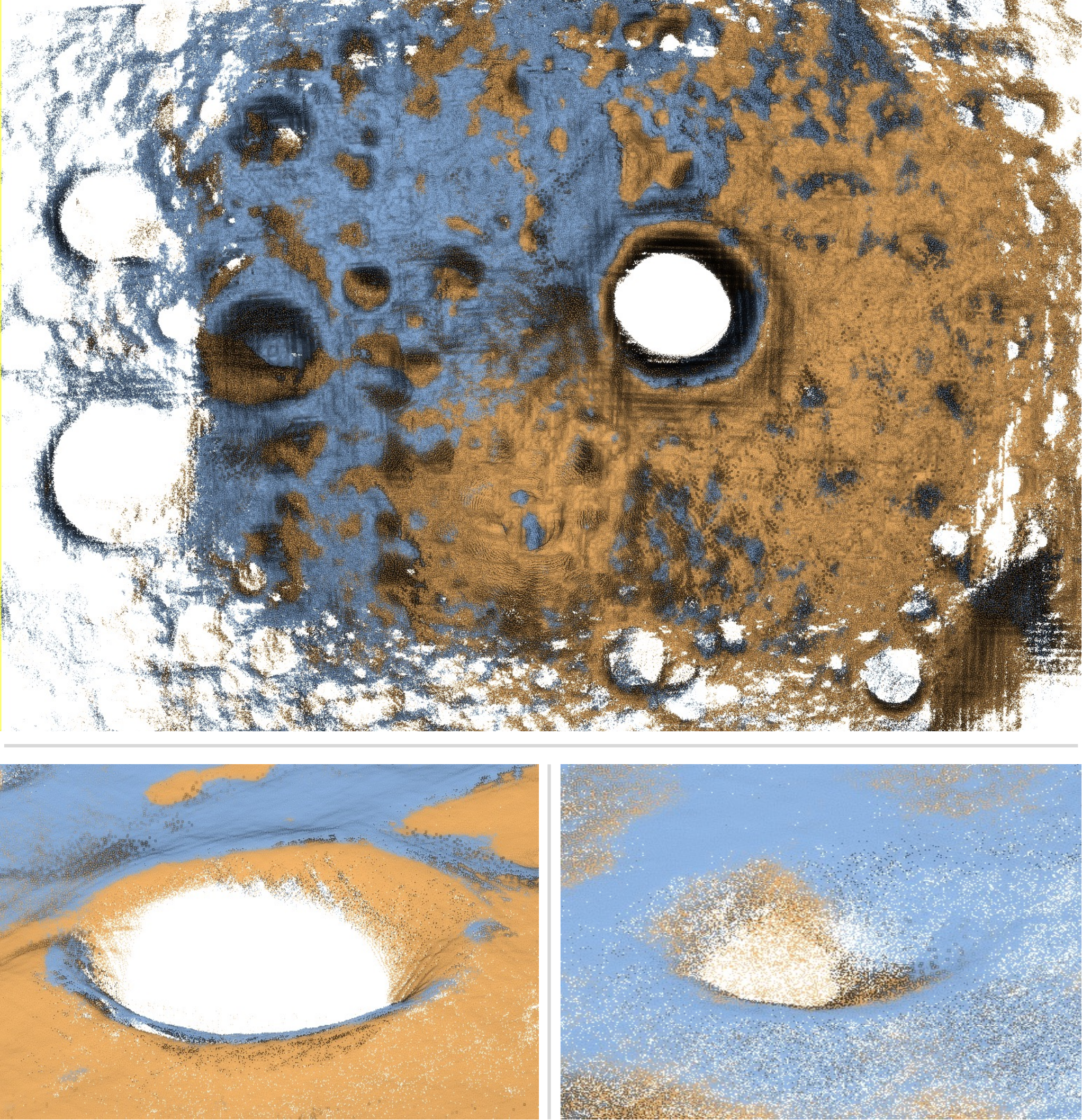}
        \caption{DCL-SLAM~\cite{zhong2023dcl}}
    \end{subfigure}
    \begin{subfigure}{0.66\columnwidth}
        \includegraphics[trim={0 0 0 0},clip, width=\columnwidth]{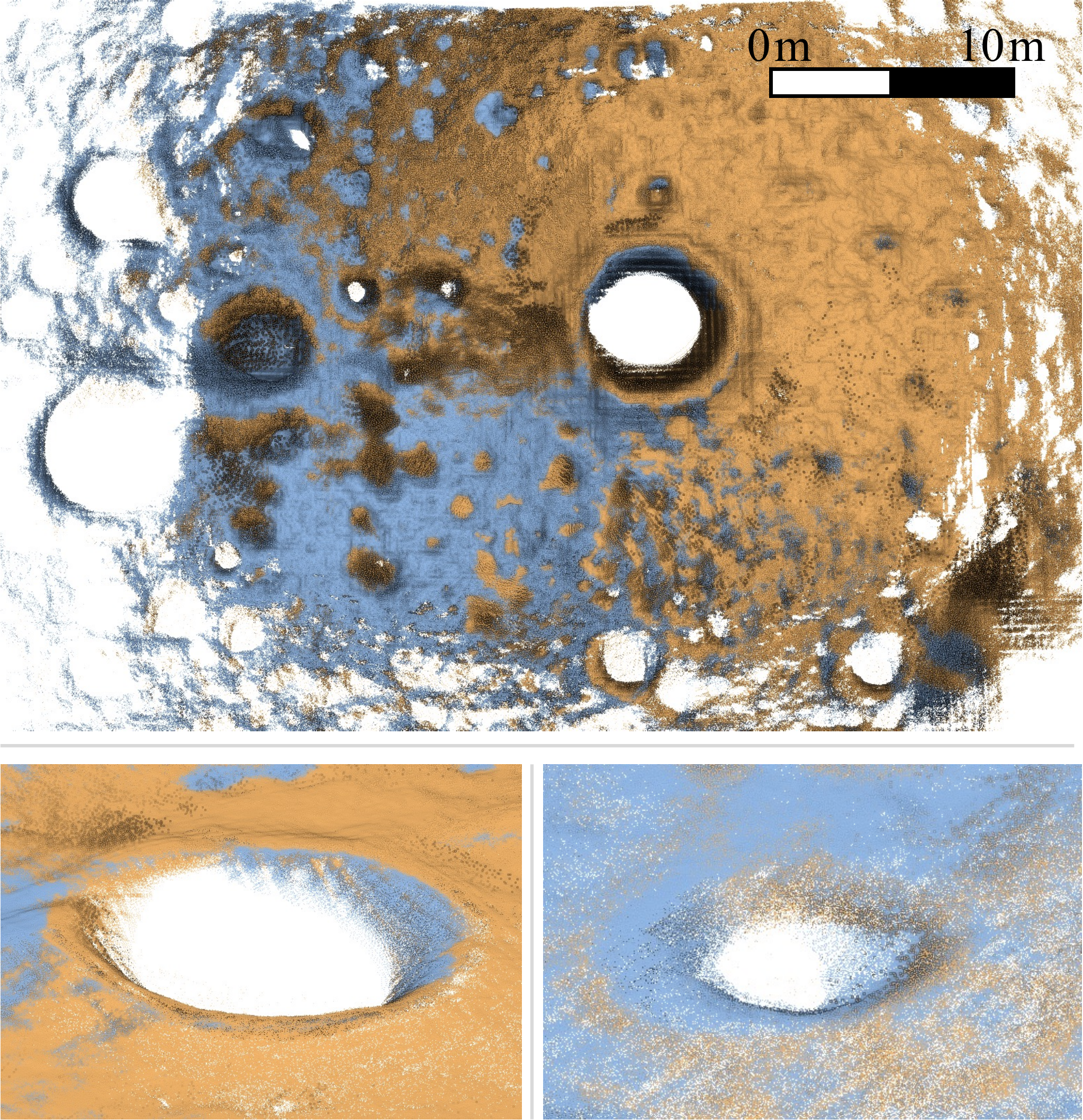}
        \caption{SKiD-SLAM}
    \end{subfigure}
    \caption{Qualitative multi-robot map merging results on the \texttt{Moon} sequence.
    (a)~DiSCo-SLAM and (b)~DCL-SLAM leave visible map misalignments, whereas (c)~SKiD-SLAM produces a consistent merged map.}
    \label{fig:fig13_cave2}
    \vspace{-0.4cm}
\end{figure*}

The qualitative multi-robot map merging results in Figs.~\ref{fig:fig12_cave1} and \ref{fig:fig13_cave2} further illustrate this performance.
While DiSCo-SLAM and DCL-SLAM produce visibly misaligned or distorted maps, SKiD-SLAM merges the robots' independent maps into a single consistent global map on both sequences.
Although the single robot trajectory accuracy of SKiD-SLAM is comparable to that of the baselines, \tabref{tab:slam_comparison} quantitatively confirms that its advantage is most pronounced in inter-robot alignment.
For the alignment of robot $\beta$ to robot $\alpha$, SKiD-SLAM attains the lowest error on both the \texttt{Tunnel} and \texttt{Moon} sequences, reducing the translational alignment error to $0.403\,$m on \texttt{Tunnel} and the rotational alignment error to $0.080^\circ$ on \texttt{Moon}, far below the baselines.
This confirms that the two-stage outlier rejection and distributed back-end of SKiD-SLAM align the robots' independent maps into a single consistent estimate, which is the central requirement of distributed multi-robot SLAM.

  \begin{table*}[t]
  \centering
  \captionsetup{width=0.99\textwidth, justification=justified}
  \caption{Scale of the field deployments, reporting the number of robots and the per-robot operating
  duration, trajectory length, and number of keyframes for each sequence.}
  \label{tab:field_deployments}
  \setlength{\tabcolsep}{12pt}
  \centering\resizebox{0.99\textwidth}{!}{\tiny
  \begin{tabular}{l|c|c|c|c}
  \toprule \midrule
  Sequence & \# of Robots & Duration [s] & Length [m] & \# of Keyframes \\
  \midrule
  Majang lava-tube cave            & 3 & 227.4 / 153.3 / 343.1 & 24.4 / 18.8 / 50.3 & 102 / 52 / 156 \\
  KICT planetary analog terrain  & 2 & 443.9 / 519.8              & 225.2 / 200.0        & 431 / 468 \\
  \midrule \bottomrule
  \end{tabular}}
  \end{table*}

Before leaving the lab for the field, we therefore conclude that SOLiD and KISS-Matcher are sound module choices for SKiD-SLAM: they yield a budget-compliant system that performs reliable distributed multi-robot SLAM and, in particular, accurate inter-robot alignment.

\section{Field Evaluation}
\label{sec:field_eval}
After validating the module choices in the lab, we now report how our SKiD-SLAM performs in the field, at the two analog sites introduced in \Cref{sec:field_test} (more details in \Cref{tab:field_deployments}).

\begin{figure*}[t]
    \centering
    \captionsetup{justification=justified}
    \captionsetup[subfigure]{justification=centering}
    \begin{subfigure}{0.99\columnwidth}
        \includegraphics[trim={0 0 0 0},clip, width=\columnwidth]{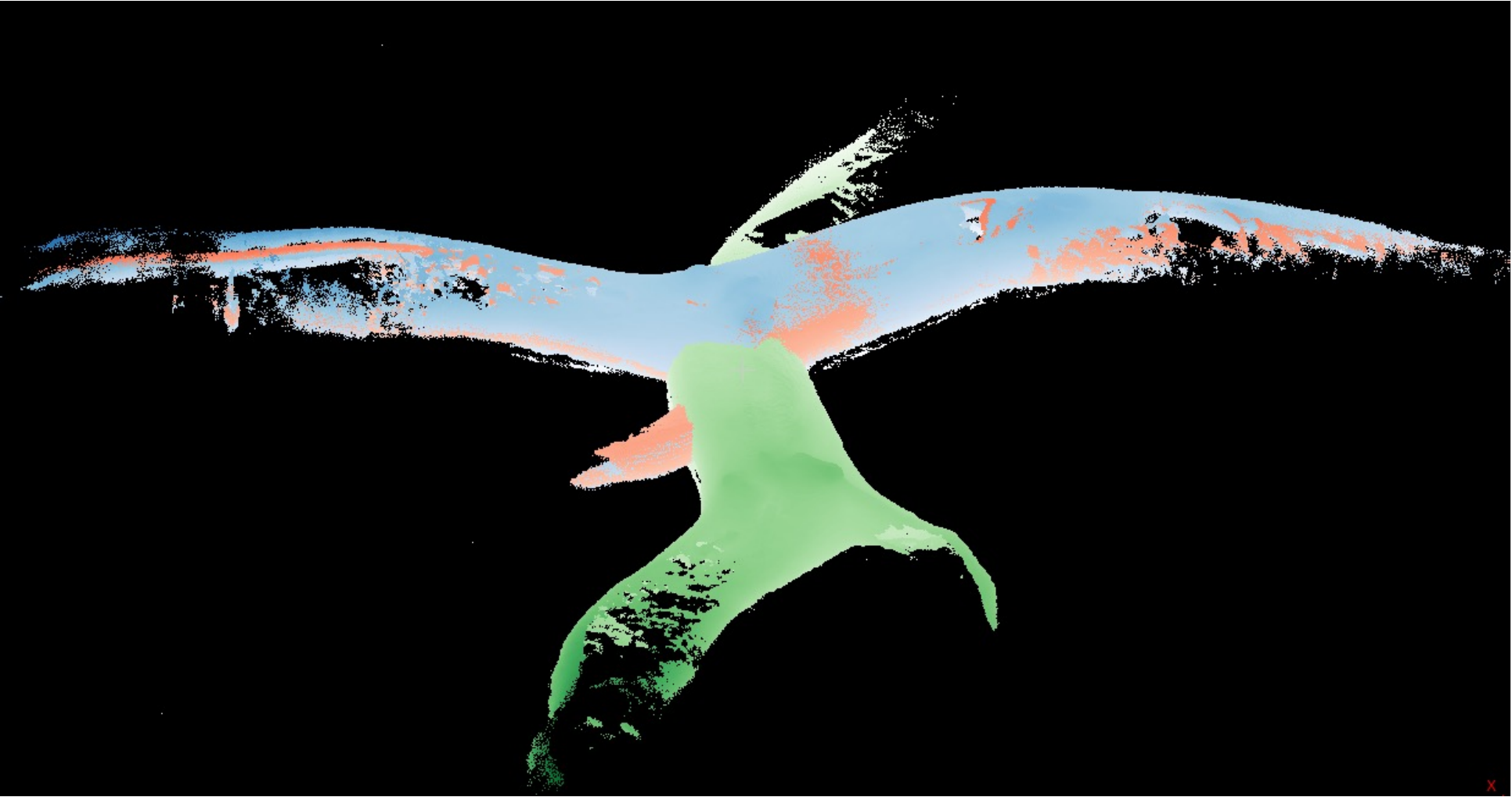}
        \caption{DiSCo-SLAM~\cite{huang2022disco}}
    \end{subfigure}
    \begin{subfigure}{0.99\columnwidth}
        \includegraphics[trim={0 0 0 0},clip, width=\columnwidth]{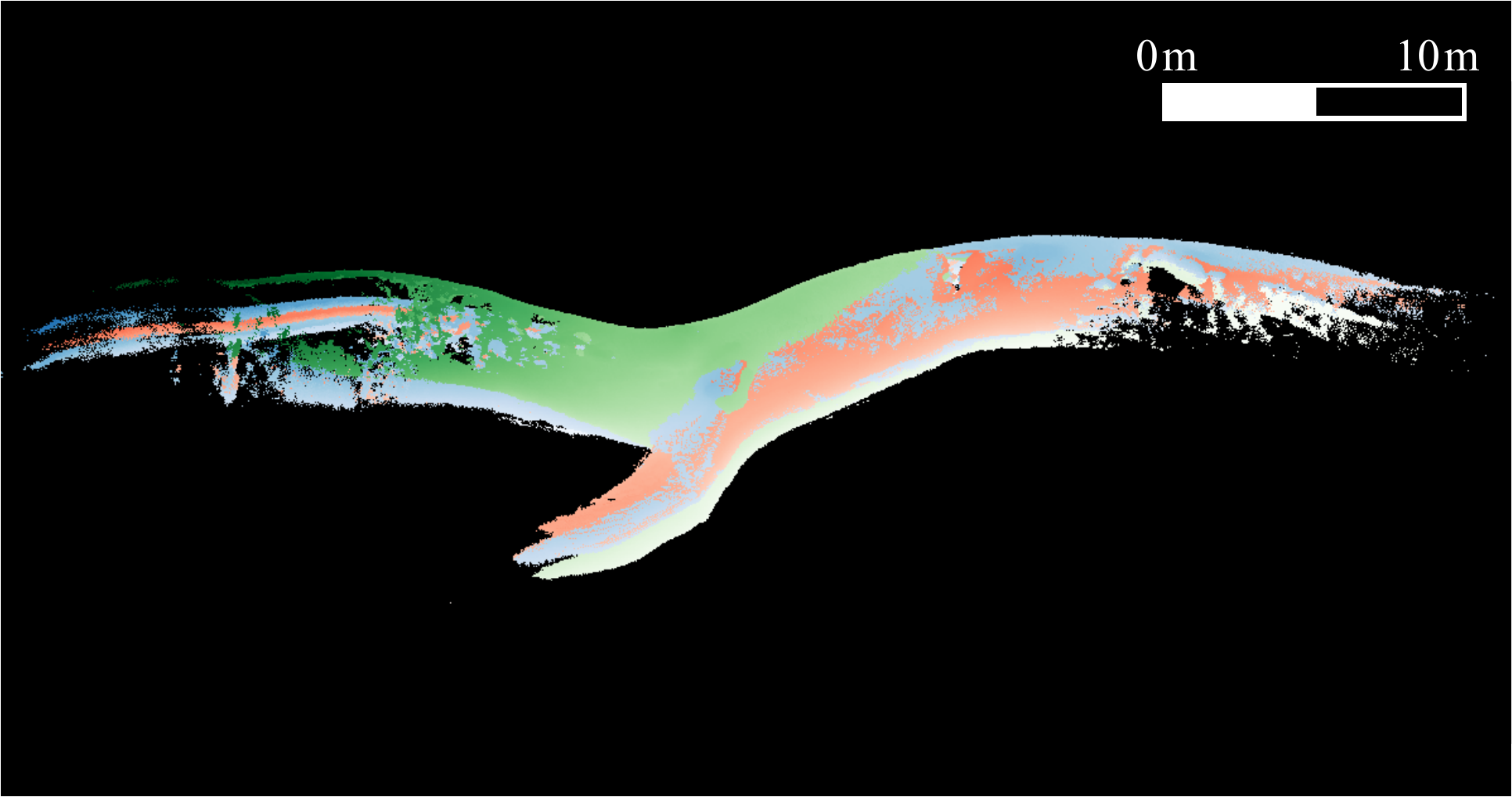}
        \caption{DCL-SLAM~\cite{zhong2023dcl}}
    \end{subfigure}
    \begin{subfigure}{0.99\columnwidth}
        \includegraphics[trim={0 0 0 0},clip, width=\columnwidth]{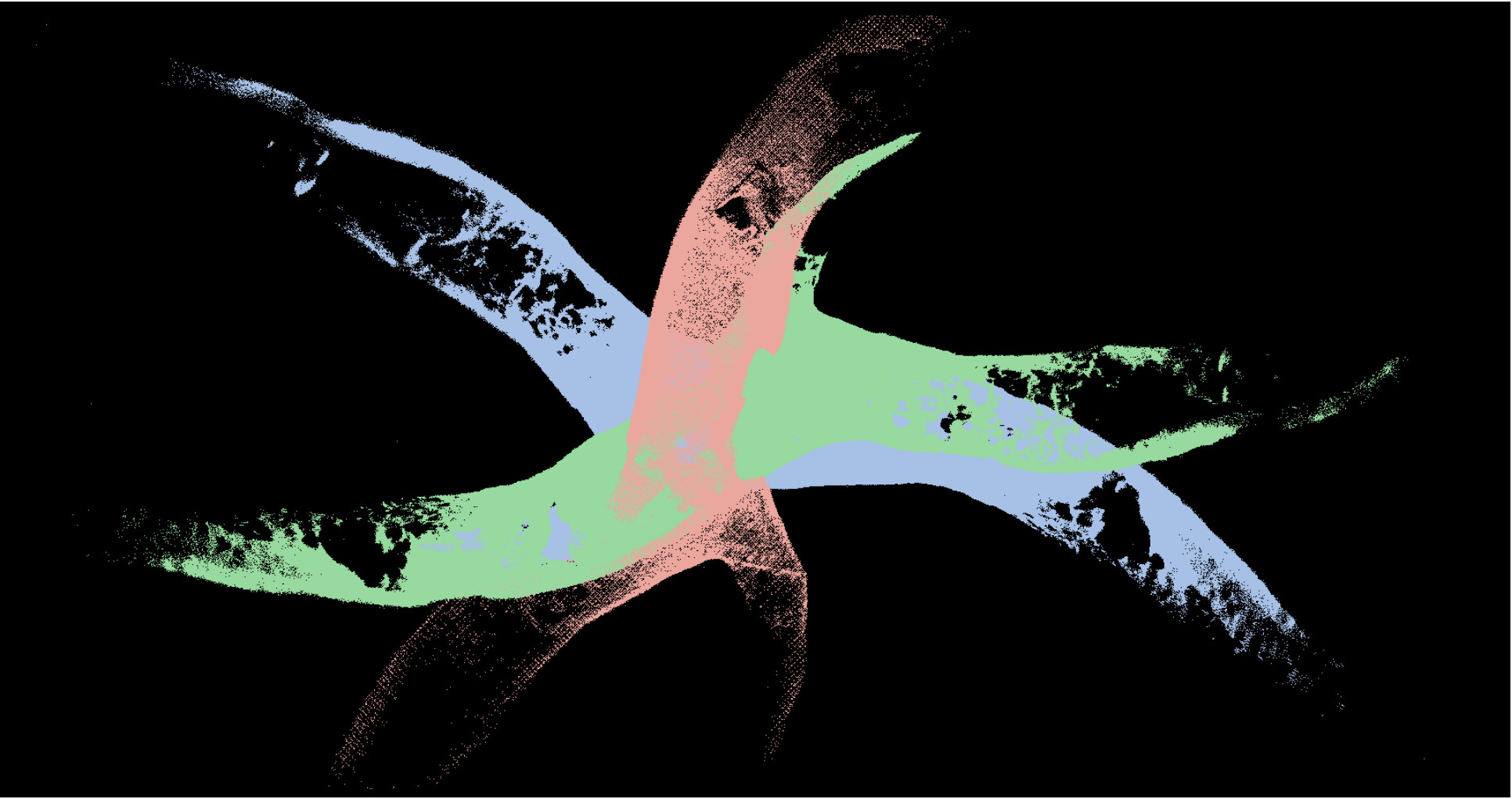}
        \caption{Multi-Proxy~\cite{wang2026communication}}
    \end{subfigure}
    \begin{subfigure}{0.99\columnwidth}
        \includegraphics[trim={0 0 0 0},clip, width=\columnwidth]{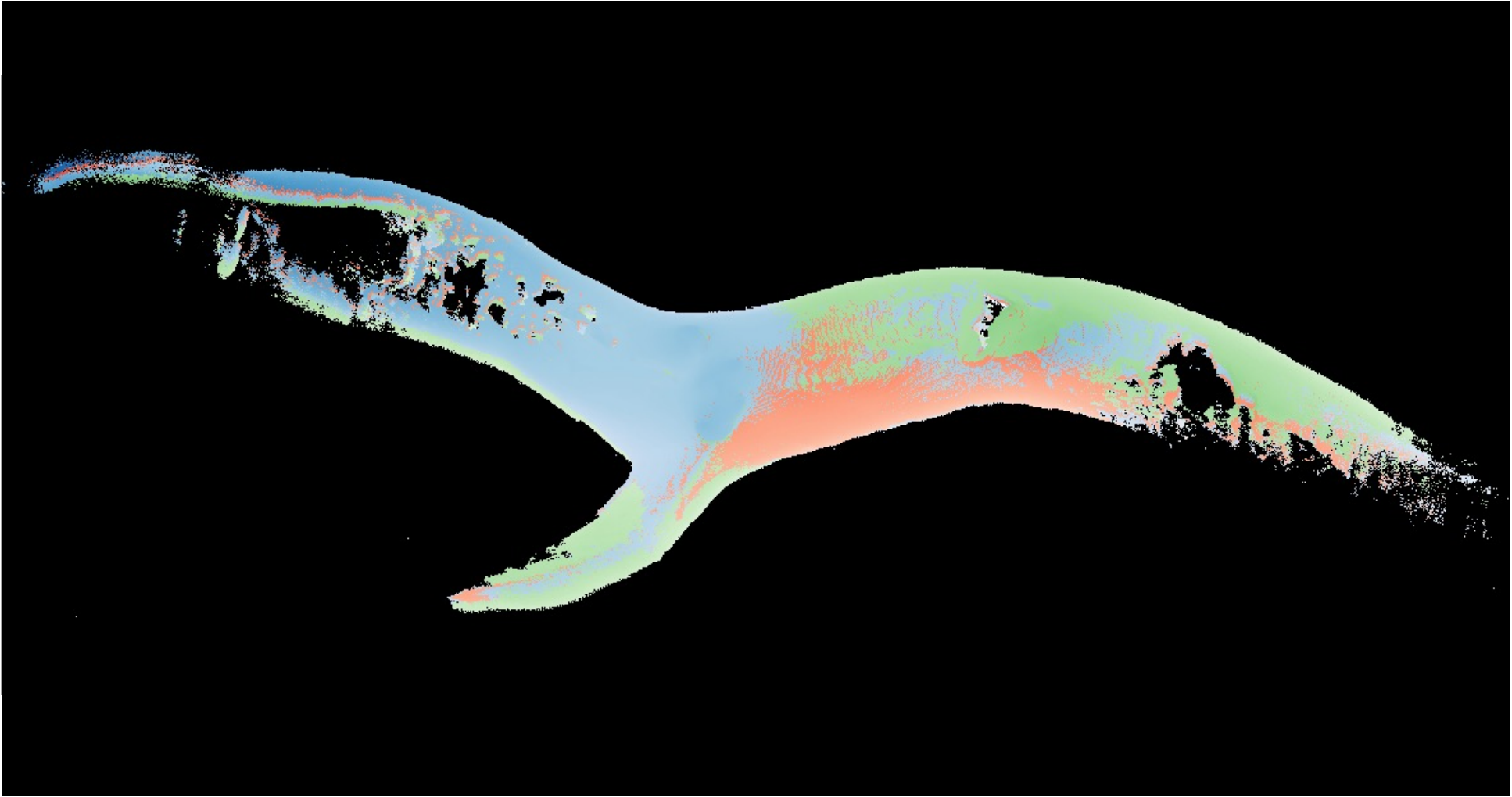}
        \caption{SKiD-SLAM}
    \end{subfigure}
    \caption{Qualitative multi-robot map merging results at the Majang lava-tube cave.
    (a)~DiSCo-SLAM, (b)~DCL-SLAM, and (c)~Multi-Proxy fail to merge the robots' maps consistently, whereas (d)~SKiD-SLAM produces a coherent global map of the cave.}
    \label{fig:fig14_majang}
    \vspace{-0.3cm}
\end{figure*}
\begin{figure*}[!h]
    \centering
    \captionsetup{justification=justified}
    \captionsetup[subfigure]{justification=centering}
    \begin{subfigure}{0.99\columnwidth}
        \includegraphics[trim={0 0 0 0},clip, width=\columnwidth]{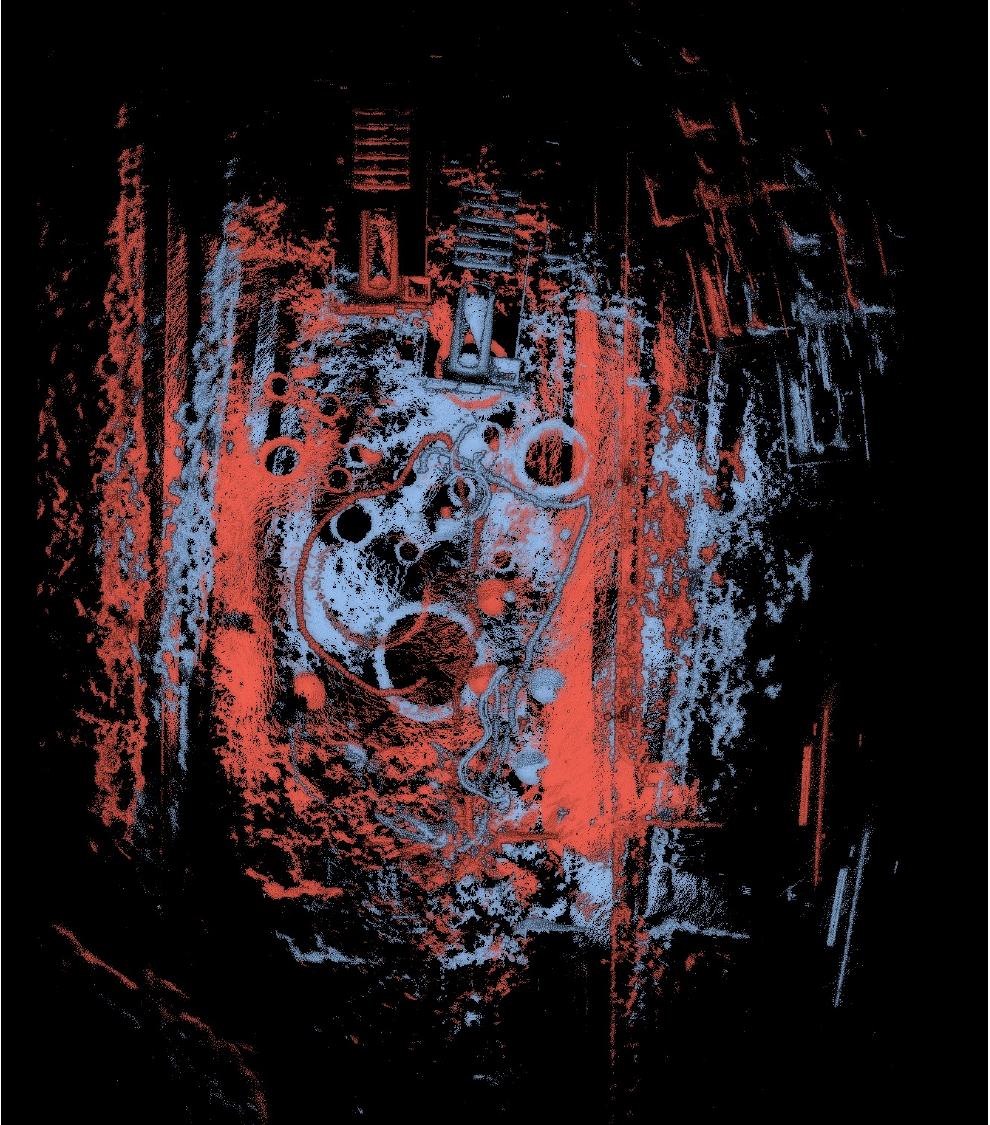}
        \caption{DiSCo-SLAM~\cite{huang2022disco}}
    \end{subfigure}
    \begin{subfigure}{0.99\columnwidth}
        \includegraphics[trim={0 0 0 0},clip, width=\columnwidth]{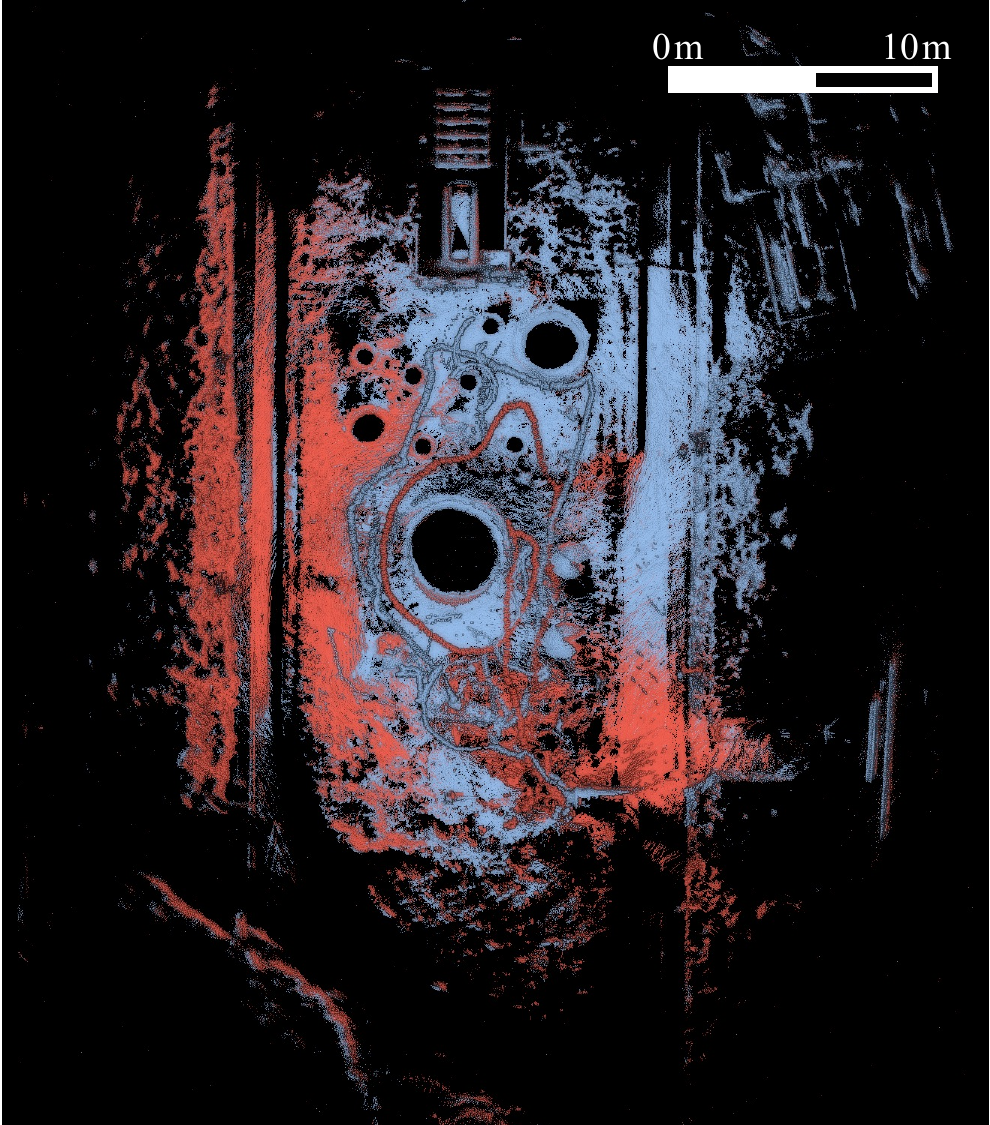}
        \caption{DCL-SLAM~\cite{zhong2023dcl}}
    \end{subfigure}
    \begin{subfigure}{0.99\columnwidth}
        \includegraphics[trim={0 0 0 0},clip, width=\columnwidth]{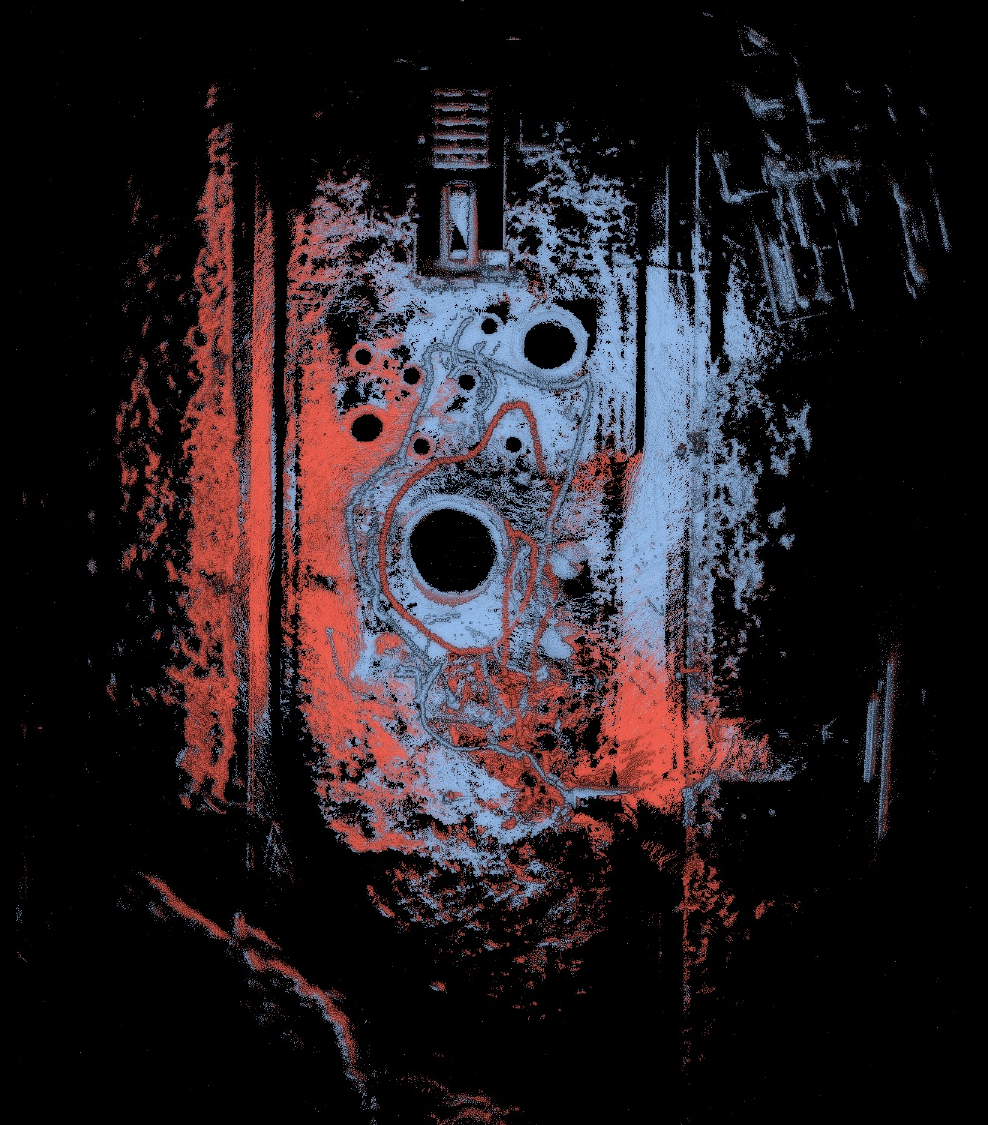}
        \caption{Multi-Proxy~\cite{wang2026communication}}
    \end{subfigure}
    \begin{subfigure}{0.99\columnwidth}
        \includegraphics[trim={0 0 0 0},clip, width=\columnwidth]{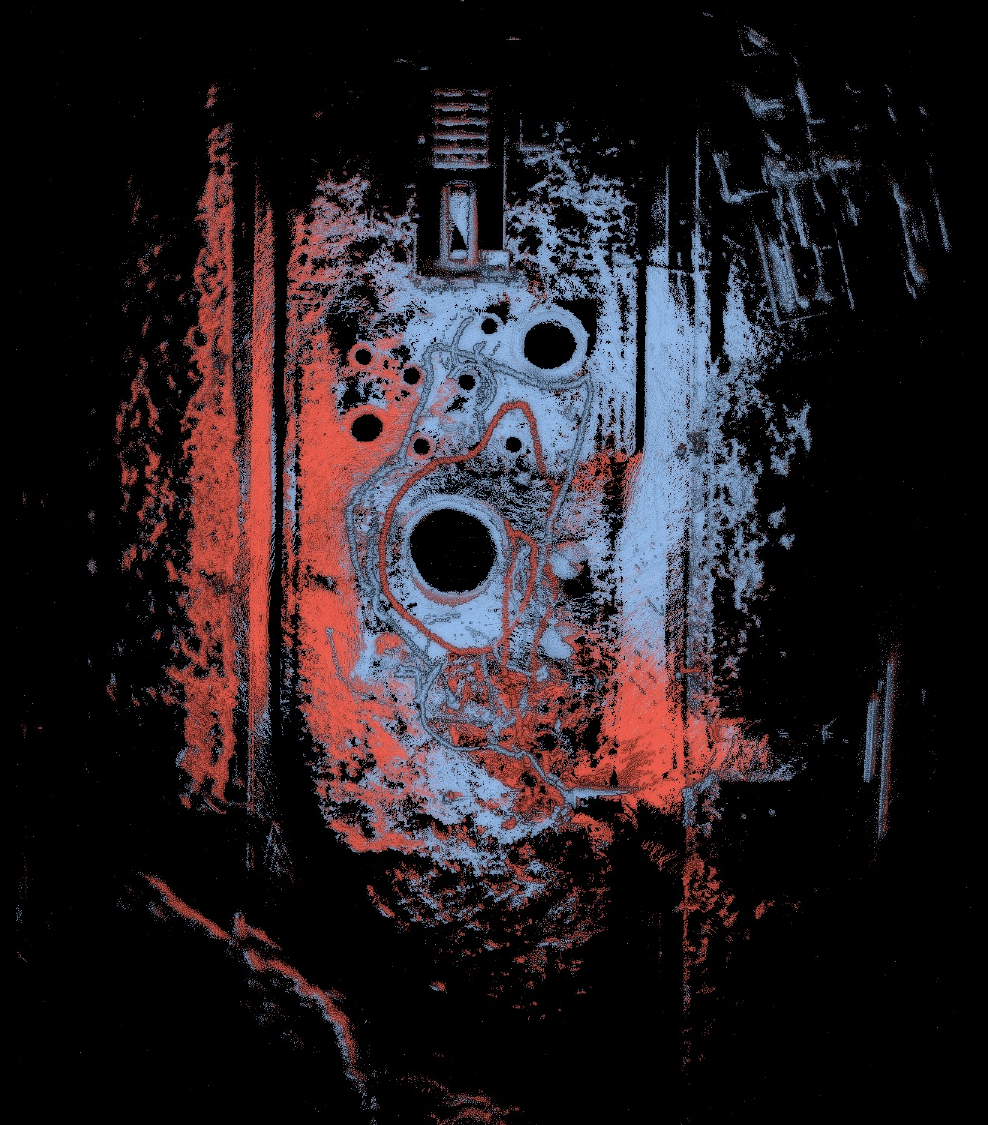}
        \caption{SKiD-SLAM}
    \end{subfigure}
    \caption{Qualitative multi-robot map merging results at the KICT planetary analog terrain in Yeoncheon.
    (a)~DiSCo-SLAM produces misaligned maps, whereas (b)~DCL-SLAM, (c)~Multi-Proxy, and (d)~SKiD-SLAM all successfully merge the robots' maps into coherent global maps.}
    \label{fig:fig15_yeoncheon}
\end{figure*}
\begin{figure*}[t]
    \centering
    \captionsetup{justification=justified}
    \captionsetup[subfigure]{justification=centering}
    \begin{subfigure}{1.00\columnwidth}
        \includegraphics[trim={0 0 0 0},clip, width=\columnwidth]{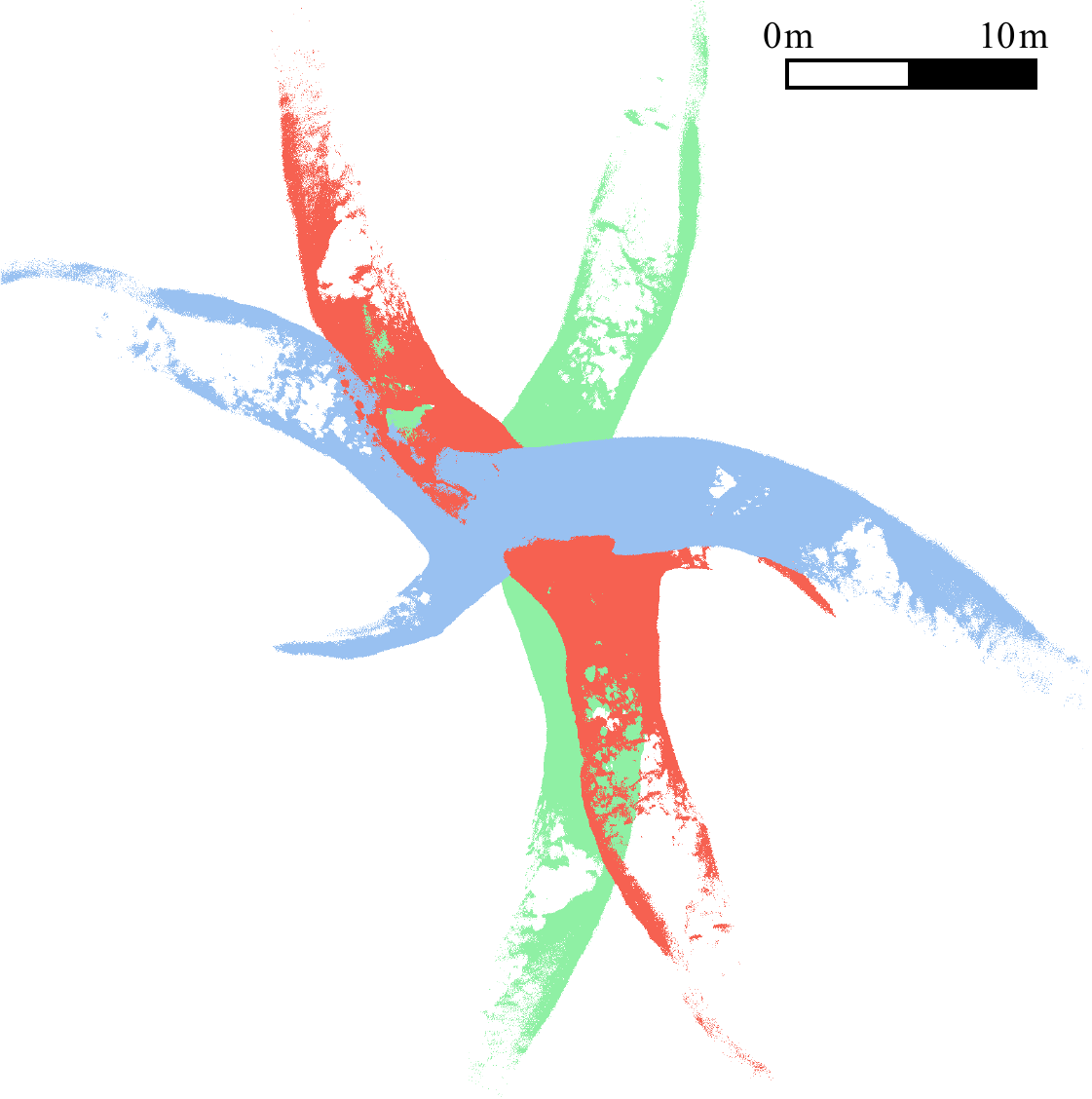}
        \vspace{-1.5em}
        \caption{Raw}
    \end{subfigure}
    \begin{subfigure}{0.33\columnwidth}
        \includegraphics[trim={0 0 0 0},clip, width=\columnwidth]{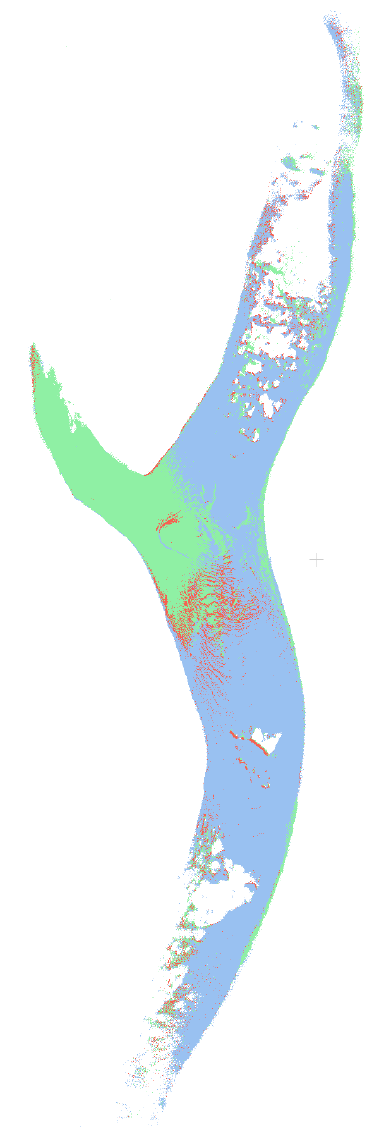}
        \vspace{-1.5em}
        \caption{$1{,}000\,\text{kbps}$}
    \end{subfigure}
    \begin{subfigure}{0.33\columnwidth}
        \includegraphics[trim={0 0 0 0},clip, width=\columnwidth]{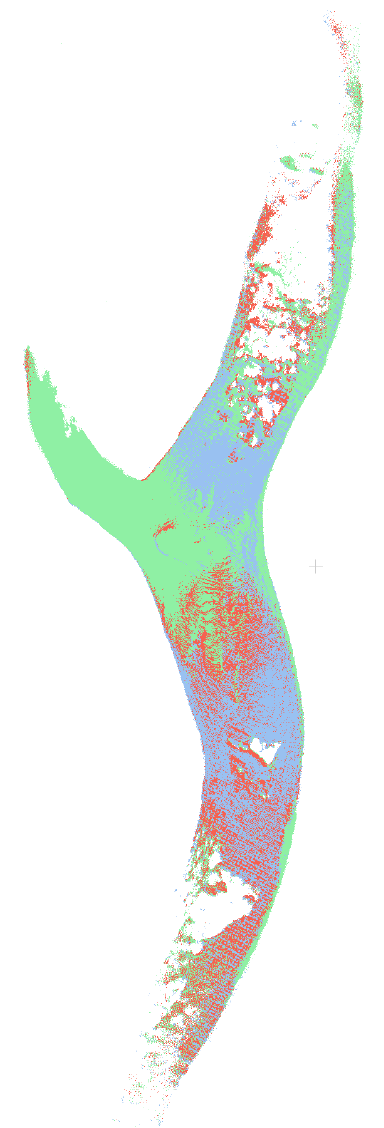}
        \vspace{-1.5em}
        \caption{$500\,\text{kbps}$}
    \end{subfigure}
    \begin{subfigure}{0.33\columnwidth}
        \includegraphics[trim={0 0 0 0},clip, width=\columnwidth]{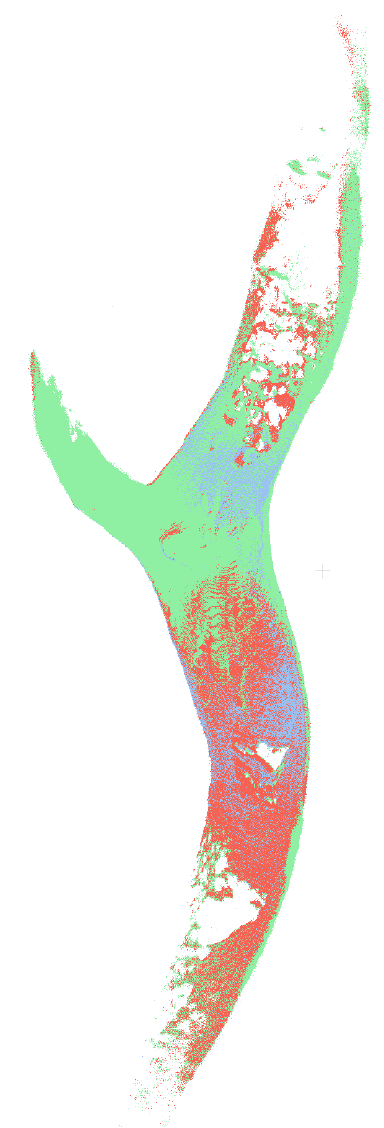}
        \vspace{-1.5em}
        \caption{$250\,\text{kbps}$}
    \end{subfigure}
    \caption{Map merging of SKiD-SLAM at the Majang lava-tube cave under bandwidth limits imposed via NetEm~\citep{hemminger2005network}.
    (a)~Raw communication without a bandwidth limit; limits of (b)~$1{,}000\,\text{kbps}$, (c)~$500\,\text{kbps}$, and (d)~$250\,\text{kbps}$.
    Our SKiD-SLAM merges the maps consistently down to $250\,\text{kbps}$, the inter-robot bandwidth budget assumed in \Cref{sec:background:budgets}.}
    \label{fig:fig16_bandwidth}
    \vspace{-5mm}
\end{figure*}
\begin{figure*}[!t]
    \centering
    \captionsetup{justification=justified}
    \captionsetup[subfigure]{justification=centering}
    \begin{subfigure}{0.99\columnwidth}
        \includegraphics[trim={0 0 0 0},clip, width=\columnwidth]{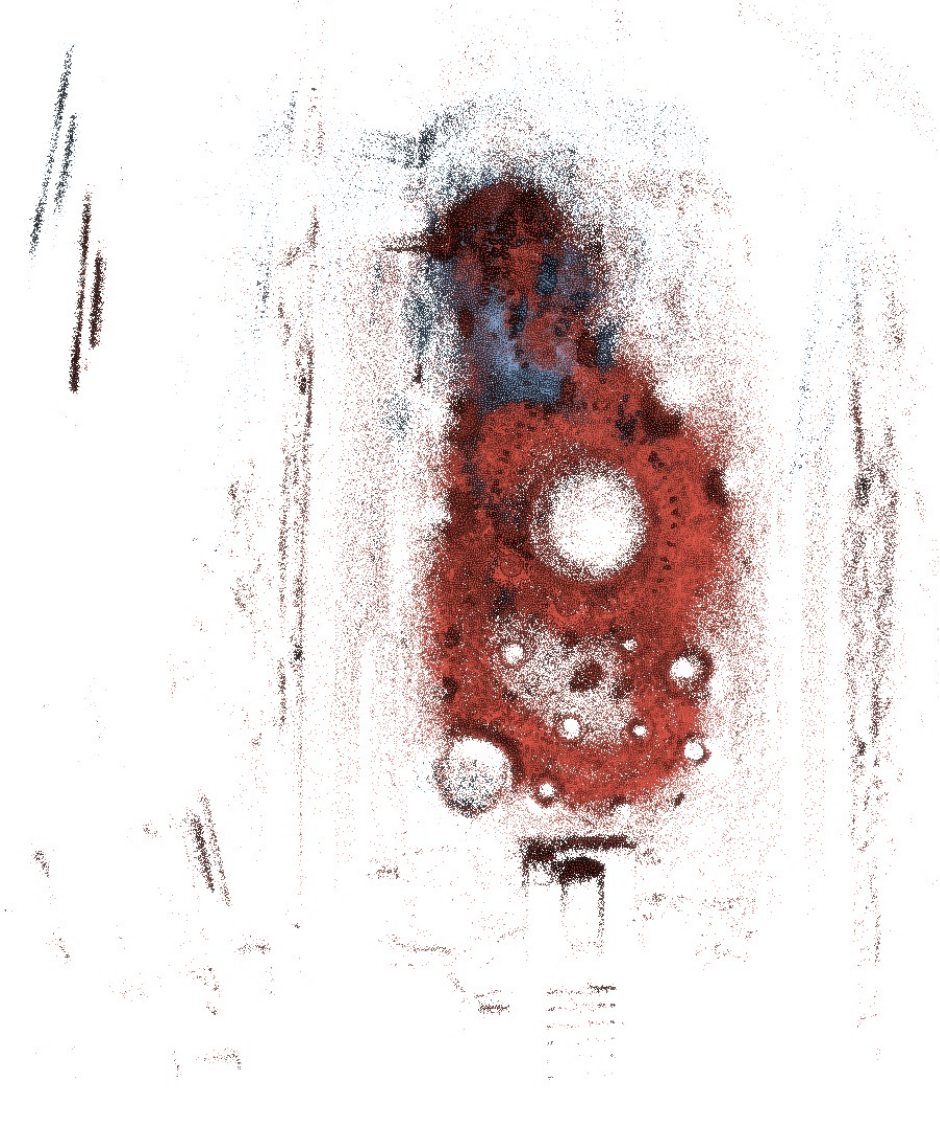}
        \vspace{-8mm}
        \caption{DiSCo-SLAM~\cite{huang2022disco}}
    \end{subfigure}
    \begin{subfigure}{0.99\columnwidth}
        \includegraphics[trim={0 0 0 0},clip, width=\columnwidth]{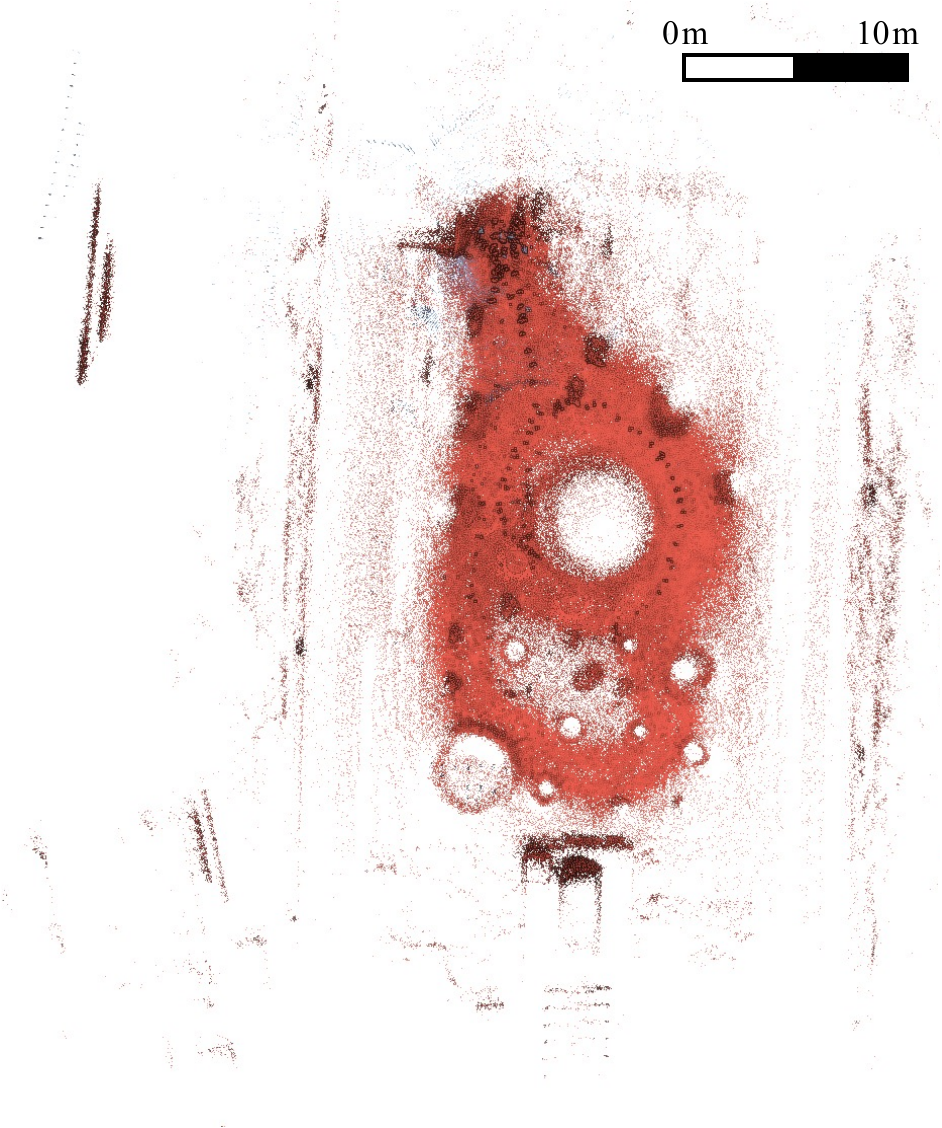}
        \vspace{-8mm}
        \caption{DCL-SLAM~\cite{zhong2023dcl}}
    \end{subfigure}
    \begin{subfigure}{0.99\columnwidth}
        \includegraphics[trim={0 0 0 0},clip, width=\columnwidth]{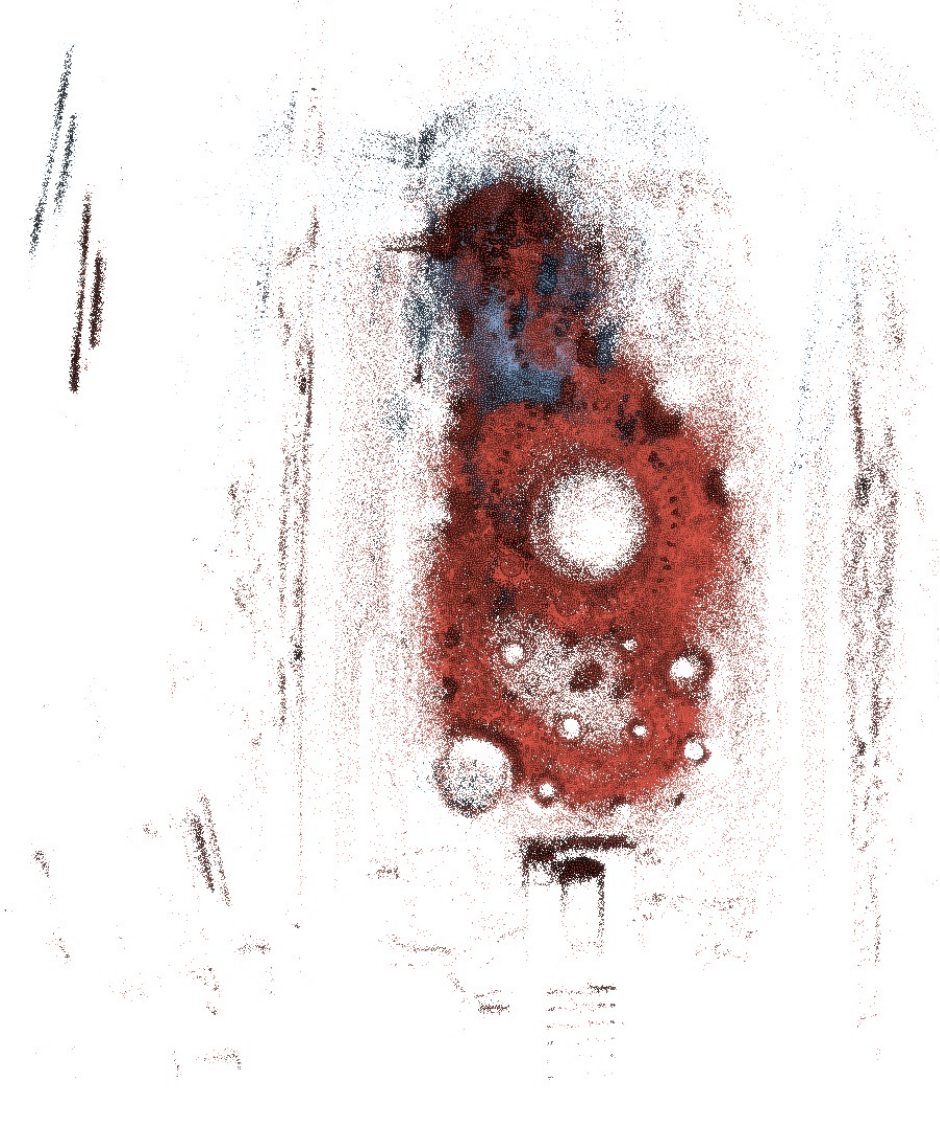}
        \vspace{-8mm}
        \caption{Multi-Proxy~\cite{wang2026communication}}
    \end{subfigure}
    \begin{subfigure}{0.99\columnwidth}
        \includegraphics[trim={0 0 0 0},clip, width=\columnwidth]{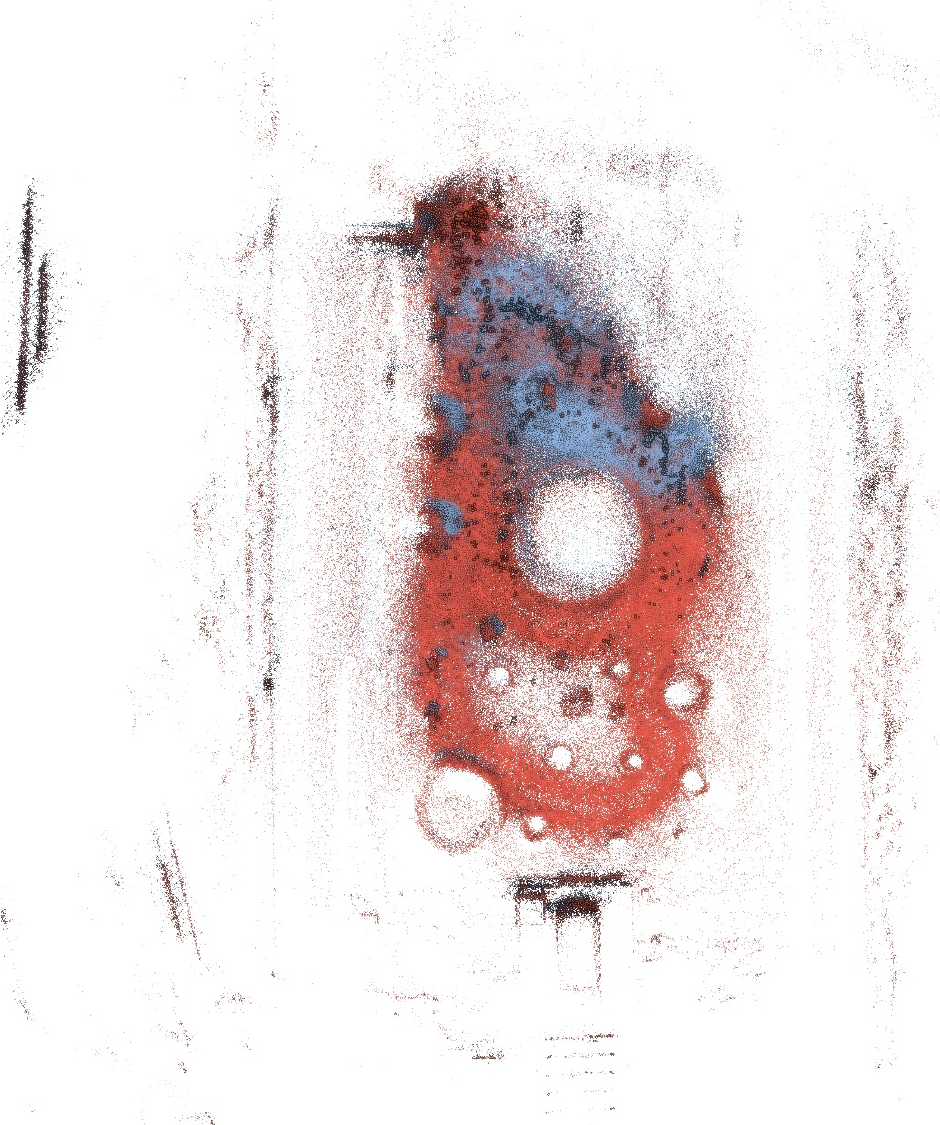}
        \vspace{-8mm}
        \caption{SKiD-SLAM}
    \end{subfigure}
    \caption{Multi-robot map merging at the KICT planetary analog terrain under a $250\,\text{kbps}$ bandwidth limit.
    (a)~DiSCo-SLAM, (b)~DCL-SLAM, and (c)~Multi-Proxy fail under this budget, whereas (d)~our SKiD-SLAM completes merging under the same limit.}
    \label{fig:fig17_yeoncheon_budget}
\end{figure*}

\subsection{Map Merging at the Field Sites}
First, we compare SKiD-SLAM against DiSCo-SLAM~\cite{huang2022disco}, DCL-SLAM~\cite{zhong2023dcl}, and Multi-Proxy~\cite{wang2026communication} on the data collected at both sites.
\figref{fig:fig14_majang} shows the results at the Majang lava-tube cave for \caveconcept, and \figref{fig:fig15_yeoncheon} those at the KICT planetary analog terrain in Yeoncheon for \surfaceconcept.
At the Majang cave, DiSCo-SLAM, DCL-SLAM, and Multi-Proxy fail to merge the robots' maps into a consistent estimate.
At the KICT terrain without bandwidth constraints, DiSCo-SLAM produces misaligned maps, while DCL-SLAM, Multi-Proxy, and SKiD-SLAM successfully merge the robots' maps~(\figref{fig:fig15_yeoncheon}).
This indicates that the module choices validated in the lab carry over to real field conditions, where the terrain, sensing, and platform differ from the datasets used for validation.

\subsection{Operation under the Communication Budget}
\label{sec:field_eval:communication_budget}
We next examine whether SKiD-SLAM operates within the inter-robot bandwidth budget assumed in \Cref{sec:background:budgets}, namely $250\,\text{kbps}$.
Because our SKiD-SLAM was the only evaluated method to merge maps reliably at the cave, we first stress-test it alone at Majang by progressively tightening the available bandwidth using NetEm~\citep{hemminger2005network}, which emulates network conditions by controlling bandwidth, latency, and packet loss.
As shown in \figref{fig:fig16_bandwidth}, our SKiD-SLAM continues to merge the maps consistently as the limit is reduced from $1{,}000\,\text{kbps}$ to $500\,\text{kbps}$ and down to $250\,\text{kbps}$.
Thus, under the tested NetEm configuration, its descriptor and scan exchange supported successful map merging within the imposed budget.

Finally, we compare all four methods at the KICT planetary analog terrain under the same $250\,\text{kbps}$ bandwidth limit imposed by NetEm.
As shown in \figref{fig:fig17_yeoncheon_budget} and \tabref{tab:bandwidth_degradation}, DiSCo-SLAM, DCL-SLAM, and Multi-Proxy all fail to merge their maps under this budget.
At this limit, their average transmitted message sizes and scan drop ratios are higher than those of SKiD-SLAM, as reported in \tabref{tab:bandwidth_degradation}.
In contrast, our SKiD-SLAM completes distributed map merging under the same limit.
This result links the descriptor-level efficiency measured in \tabref{tab:desc} and the communication measurements in \tabref{tab:bandwidth_degradation} to an operational outcome in the field: under the imposed communication budget, only SKiD-SLAM completed distributed map merging among the evaluated methods.

\section{Ablation Study}
\label{sec:ablation}
The field deployments in \Cref{sec:field_eval} demonstrate that our SKiD-SLAM merges multi-robot maps reliably across both planetary analog sites.
To identify which design choices are responsible for this reliability, we disable one component at a time and measure its effect on the cave and crater sequences.
As shown in Figs.~\ref{fig:fig19_ablation_majang} and \ref{fig:fig20_ablation_yeoncheon}, we first examine the SOLiD descriptor, the KISS-Matcher global registration, the truncated MSE for geometric verification, and the PCM for outlier rejection.
We then examine robustness to limited inter-robot overlap and communication impairments, and finally assess the full-pipeline computational burden on the embedded platform.

\subsection{Effect of the SOLiD Descriptor}
\label{sec:ablation:solid}
First, we replace SOLiD with Scan Context~\cite{kim2018scan} and keep the remaining pipeline unchanged.
On the KICT crater terrain, Scan Context still recognizes inter-robot places and the merge succeeds, as shown in Fig.~\ref{fig:fig20_ablation_yeoncheon}(a).
The open terrain provides distinct height variation along each azimuth, which the polar bird's-eye-view encoding of Scan Context captures reliably.
The Majang lava-tube cave removes this cue.
The narrow and enclosed geometry yields repetitive walls with little vertical structure, so Scan Context fails to distinguish places and the merge collapses, as shown in Fig.~\ref{fig:fig19_ablation_majang}(a).

We conclude that SOLiD recovers these associations in the same environment, which confirms that its spatially organized descriptor is necessary for place recognition in confined planetary subsurface settings rather than a redundant substitution.

\subsection{Effect of KISS-Matcher Registration}
\label{sec:ablation:kiss}
Next, we disable KISS-Matcher, leaving the Small-GICP refinement without a global-registration initialization.
On the KICT crater terrain, the SOLiD descriptor still recognizes the revisited place, but the resulting scan pair lacks a reliable coarse alignment to seed the refinement, so the two maps fail to converge into a consistent global map, as shown in Fig.~\ref{fig:fig20_ablation_yeoncheon}(b).
KISS-Matcher restores a reliable global registration in the same environment, which shows that robust global registration is the decisive component on this terrain rather than place recognition.
In the Majang lava-tube cave, removing KISS-Matcher has a smaller effect, as shown in Fig.~\ref{fig:fig19_ablation_majang}(b).

We conclude that KISS-Matcher is necessary to seed an accurate inter-robot transform on crater terrain where the descriptor alone underconstrains the alignment, rather than a redundant refinement stage.

\begin{table*}[!hbtp]
\captionsetup{width=0.99\textwidth, justification=justified}
\caption{Average transmitted message size and scan drop ratio (\%) under the network degradation condition, \ie bandwidth limitation, on the KICT planetary analog terrain.
                   Each size is measured as the serialized ROS message that the method's implementation transmits for inter-robot association, which is a float64 matrix or vector descriptor for DiSCo-SLAM, DCL-SLAM, and our SKiD-SLAM, and a descriptor carried in a point cloud message for Multi-Proxy, as its standard implementation transmits its proxy descriptor in point cloud form.
                   A lower value indicates better communication robustness.}
\setlength{\tabcolsep}{12pt}
\centering\resizebox{0.99\textwidth}{!}{\tiny
\begin{tabular}{l|c|c|c|c}
\toprule \midrule
Method & \makecell{DiSCo-SLAM \\ \cite{huang2022disco}} & \makecell{DCL-SLAM \\ \cite{zhong2023dcl}} & \makecell{Multi-Proxy \\ \cite{wang2026communication}} & Proposed \\ \midrule
Avg. transmitted message size [B] $\downarrow$ & \thirdc 9{,}600 & 230{,}400 & \secondc  1{,}585  & \firstc\textbf{320} \\ \midrule
Bandwidth ($1{,}000\,\text{kbps}$)                     & \thirdc 61.27   & 88.64       & \secondc 58.93     & \firstc \textbf{53.18} \\
Bandwidth ($500\,\text{kbps}$)                          & \thirdc 77.53   & 96.38       &  \secondc 66.26    & \firstc \textbf{62.65} \\
Bandwidth ($250\,\text{kbps}$)                          & \thirdc 93.34   & 99.71       &  \secondc 92.52    & \firstc \textbf{73.18} \\
\midrule \bottomrule
\end{tabular}}
\label{tab:bandwidth_degradation}
\end{table*}

\subsection{Effect of the Truncated MSE}
\label{sec:ablation:tmse}
\begin{figure*}[!t]
    \centering
    \captionsetup{justification=justified}
    \captionsetup[subfigure]{justification=centering}
    \begin{subfigure}{0.99\columnwidth}
        \includegraphics[trim={0 0 0 0},clip, width=\columnwidth]{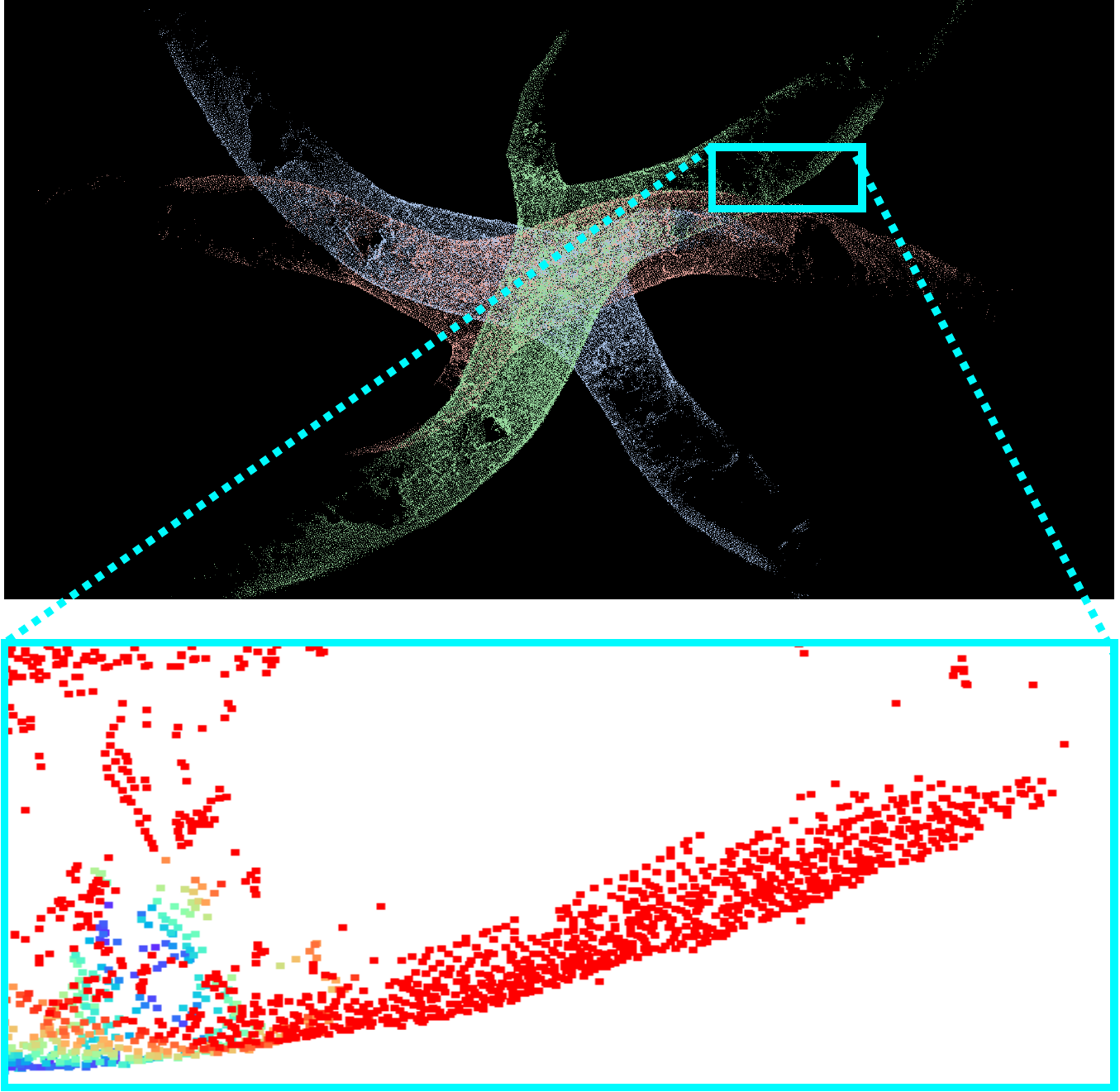}
        \caption{SKiD-SLAM w/o SOLiD~\cite{kim2024narrowing}}
    \end{subfigure}
    \begin{subfigure}{0.99\columnwidth}
        \includegraphics[trim={0 0 0 0},clip, width=\columnwidth]{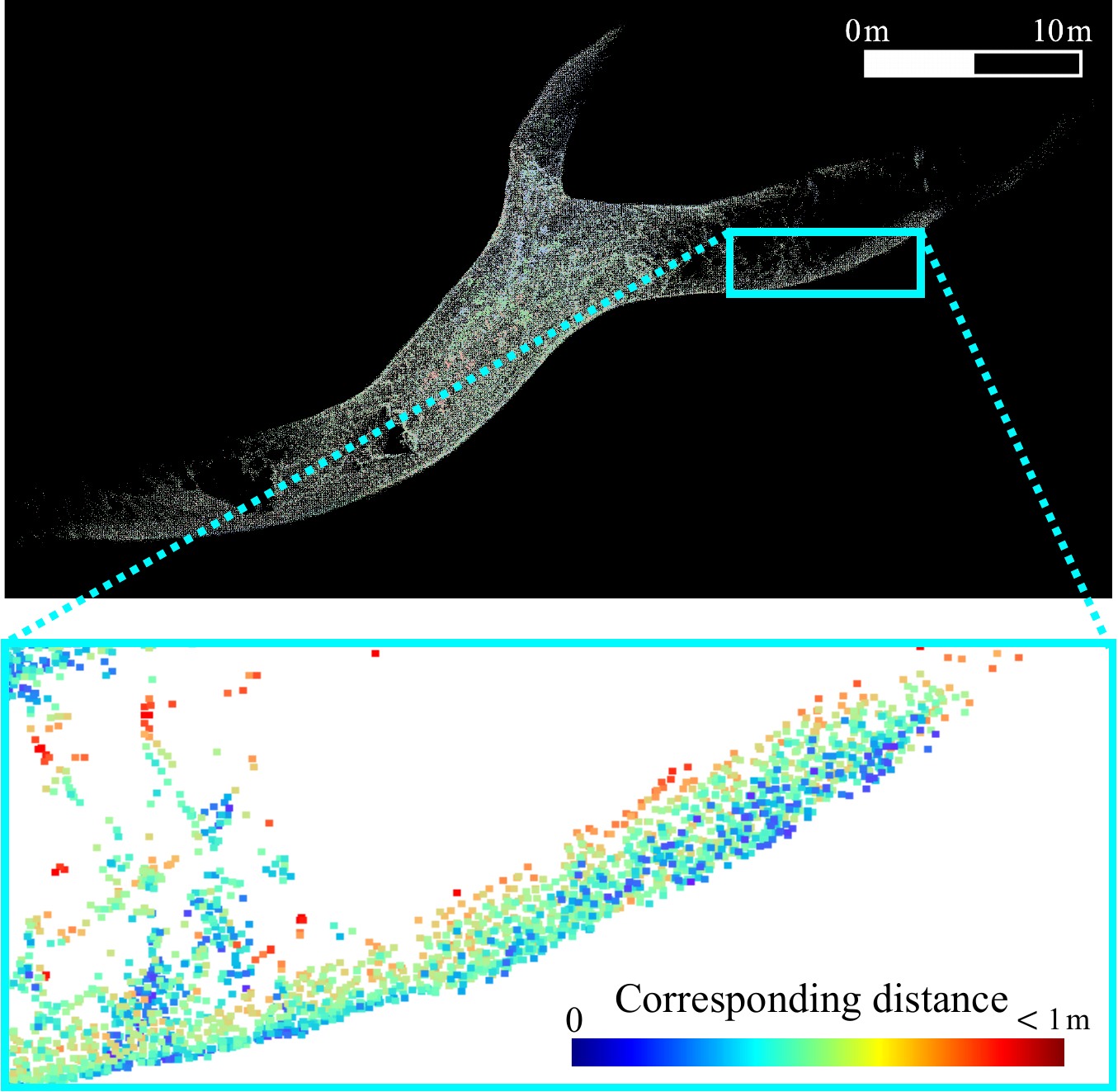}
        \caption{SKiD-SLAM w/o KISS-Matcher~\cite{lim2024kiss}}
    \end{subfigure}
    \begin{subfigure}{0.99\columnwidth}
        \includegraphics[trim={0 0 0 0},clip, width=\columnwidth]{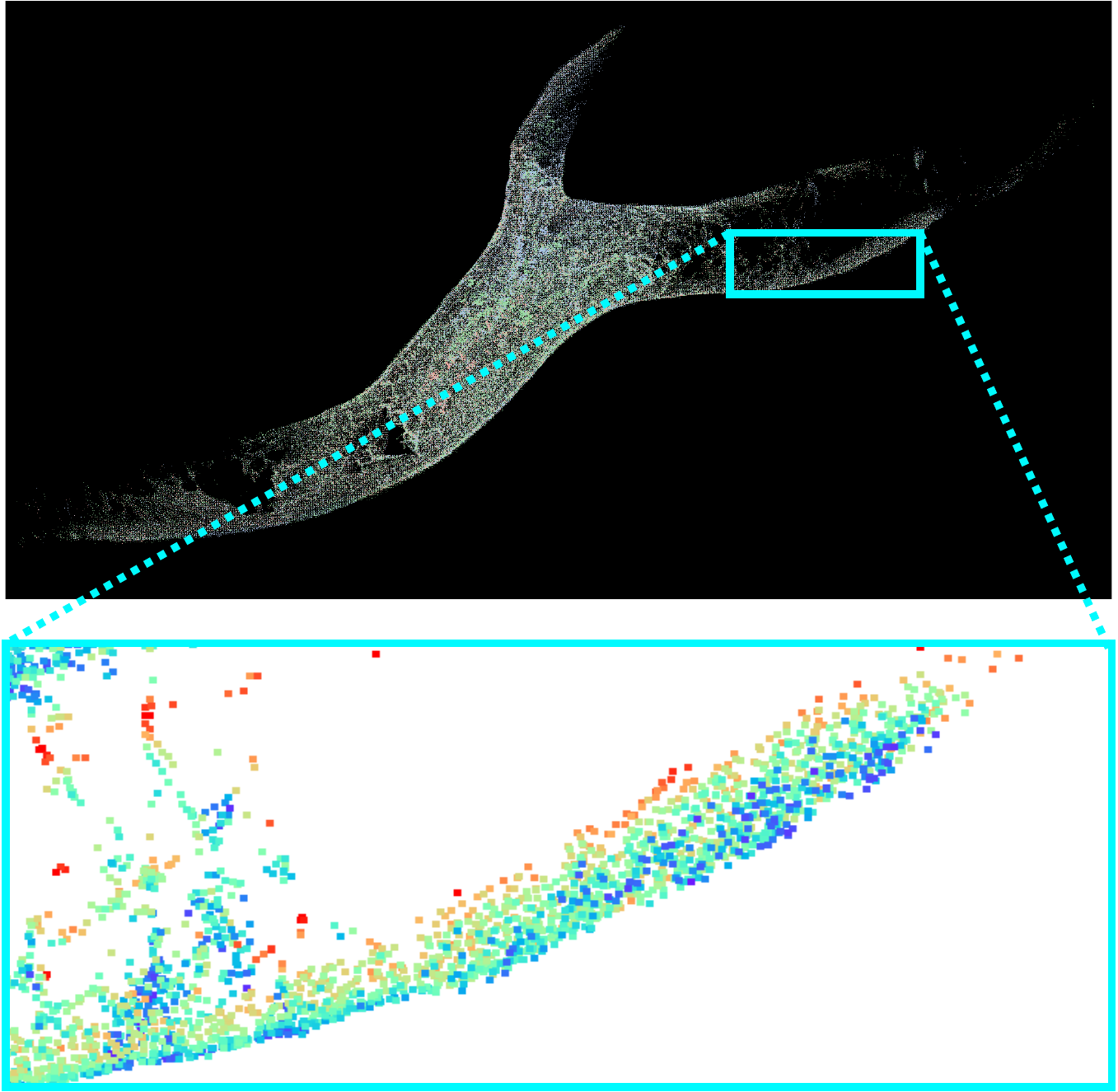}
        \caption{SKiD-SLAM w/o t-MSE}
    \end{subfigure}
    \begin{subfigure}{0.99\columnwidth}
        \includegraphics[trim={0 0 0 0},clip, width=\columnwidth]{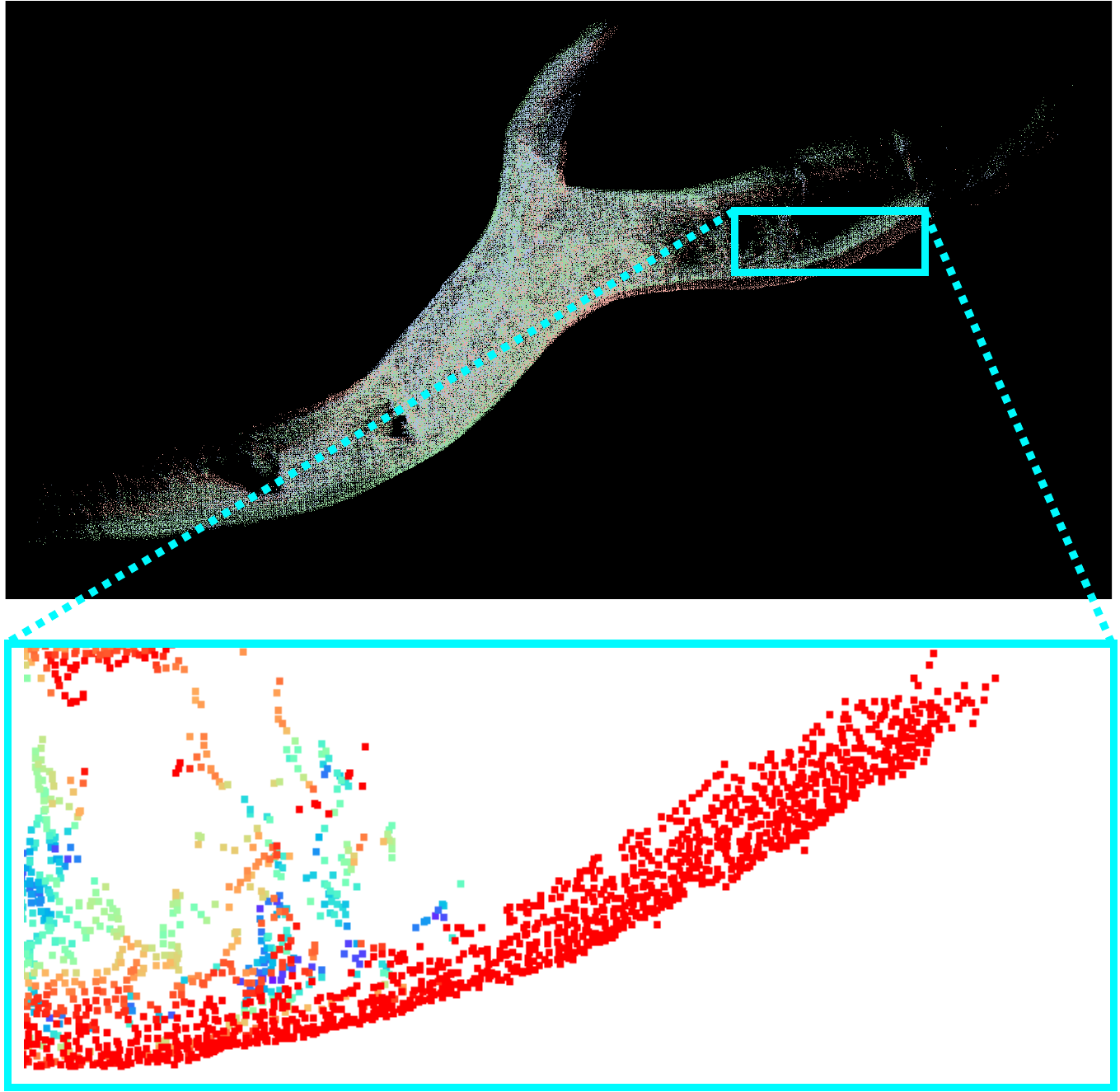}
        \caption{SKiD-SLAM w/o PCM~\cite{mangelson2018pairwise}}
    \end{subfigure}
    \caption{Component ablation study on the Majang lava-tube cave.
    Each panel disables one module: (a)~SOLiD replaced by Scan Context, (b)~without KISS-Matcher, (c)~without t-MSE, and (d)~without PCM.
    Removing SOLiD causes a clear failure and removing PCM degrades the merge, whereas the other modules have a smaller effect in this environment.}
    \label{fig:fig19_ablation_majang}
    \vspace{-5mm}
\end{figure*}

\begin{figure*}[!htbp]
    \centering
    \captionsetup{justification=justified}
    \captionsetup[subfigure]{justification=centering}
    \begin{subfigure}{0.99\columnwidth}
        \includegraphics[trim={0 0 0 0},clip, width=\columnwidth]{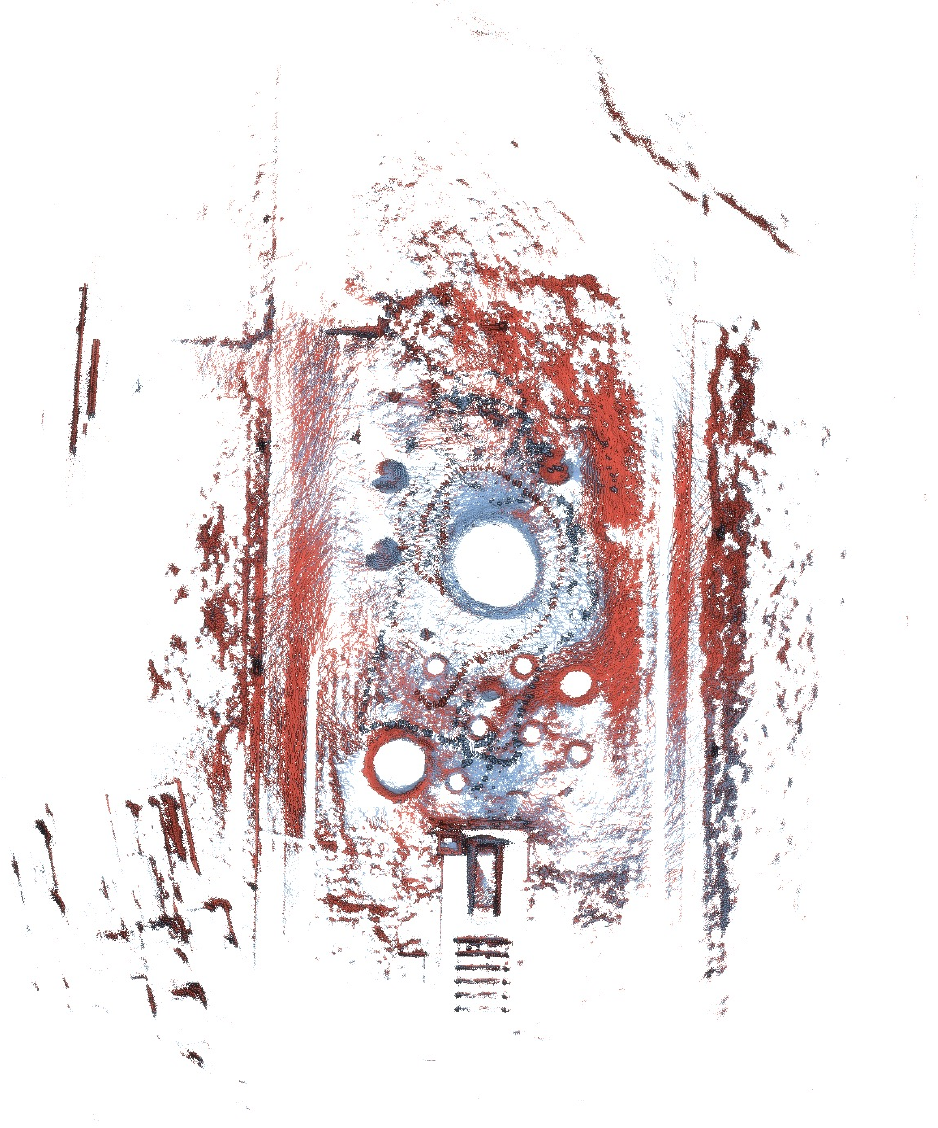}
        \caption{SKiD-SLAM w/o SOLiD~\cite{kim2024narrowing}}
    \end{subfigure}
    \begin{subfigure}{0.99\columnwidth}
        \includegraphics[trim={0 0 0 0},clip, width=\columnwidth]{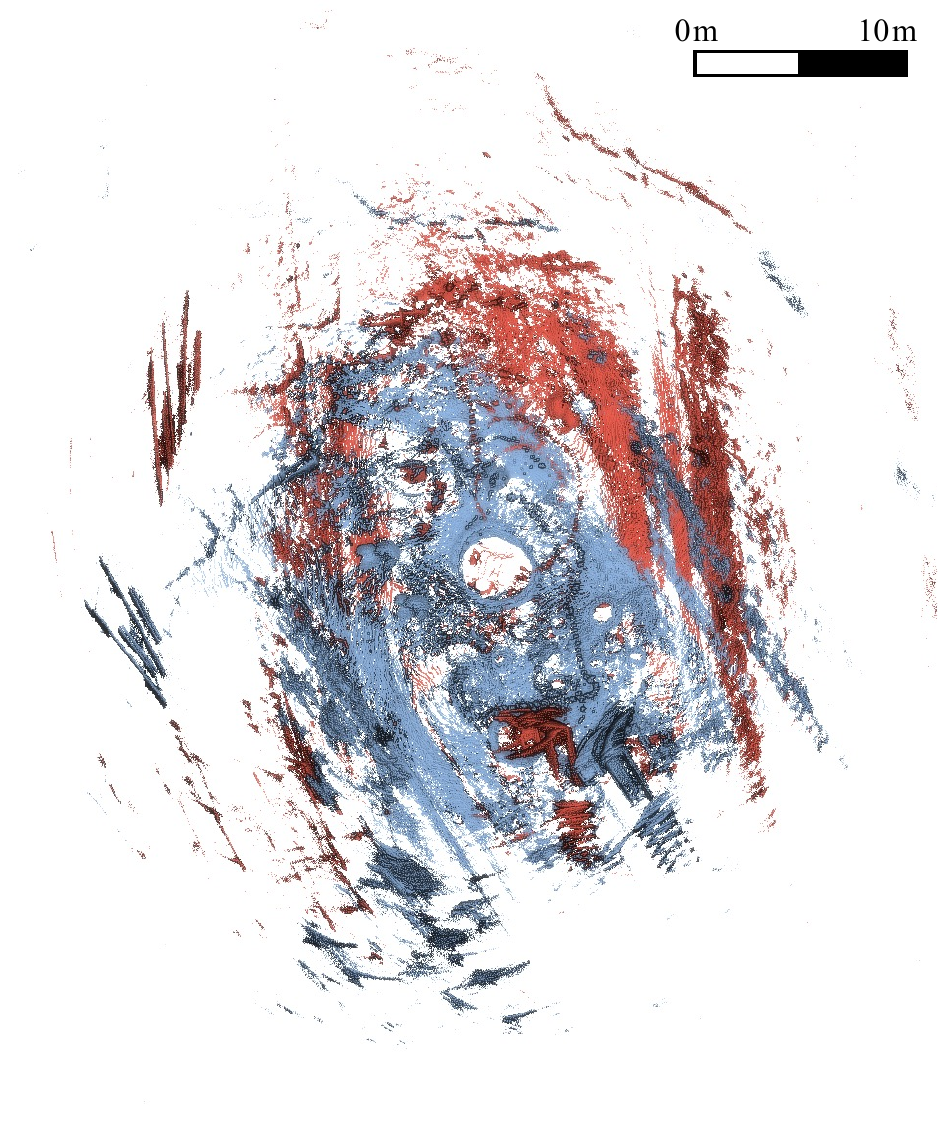}
        \caption{SKiD-SLAM w/o KISS-Matcher~\cite{lim2024kiss}}
    \end{subfigure}
    \begin{subfigure}{0.99\columnwidth}
        \includegraphics[trim={0 0 0 0},clip, width=\columnwidth]{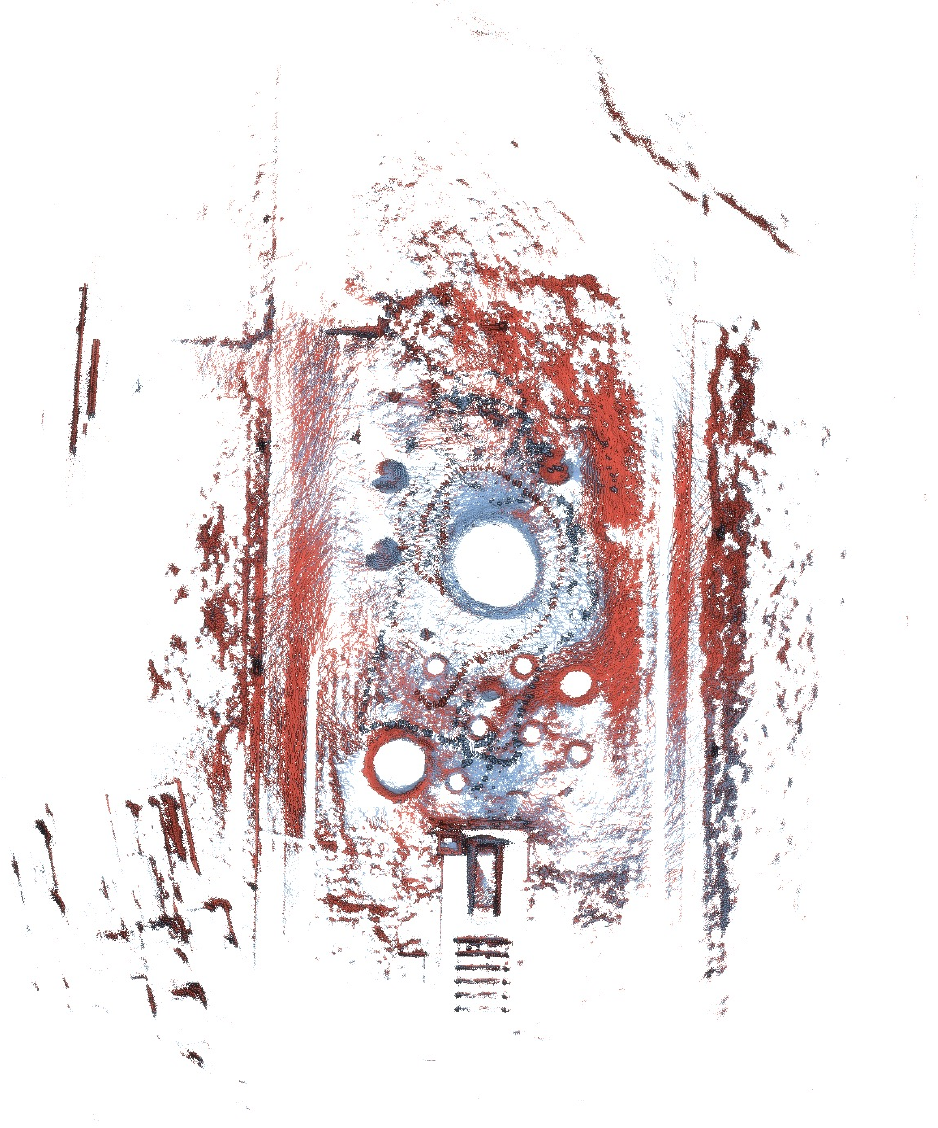}
        \caption{SKiD-SLAM w/o t-MSE}
    \end{subfigure}
    \begin{subfigure}{0.99\columnwidth}
        \includegraphics[trim={0 0 0 0},clip, width=\columnwidth]{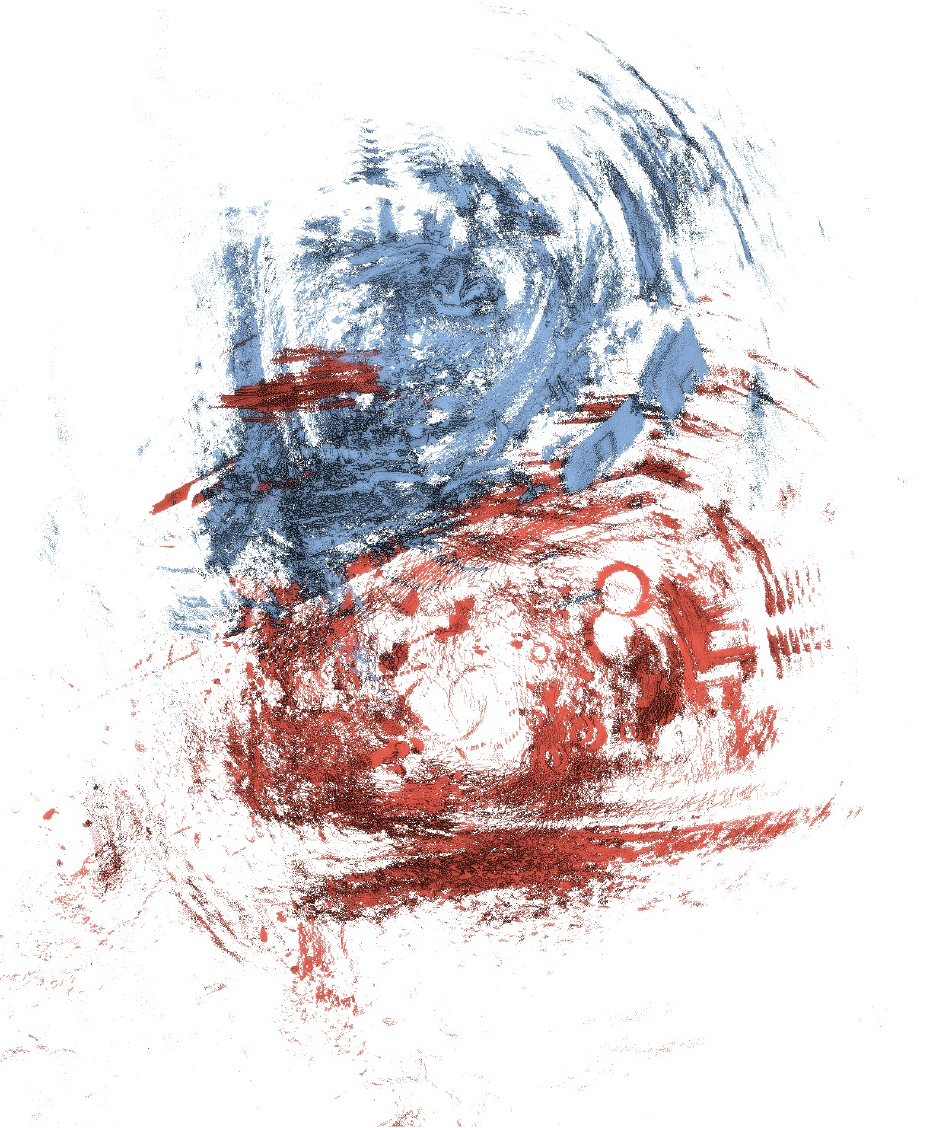}
        \caption{SKiD-SLAM w/o PCM~\cite{mangelson2018pairwise}}
    \end{subfigure}
    \caption{Component ablation study on the KICT planetary analog terrain.
    Each panel disables one module: (a)~SOLiD replaced by Scan Context, (b)~without KISS-Matcher, (c)~without t-MSE, and (d)~without PCM.
    Removing either KISS-Matcher (b) or PCM (d) causes the merge to fail, showing that the decisive component differs from the cave environment.}
    \label{fig:fig20_ablation_yeoncheon}
\end{figure*}

The truncated MSE shows no visible effect on either planetary site, as shown in Fig.~\ref{fig:fig19_ablation_majang}(c) and Fig.~\ref{fig:fig20_ablation_yeoncheon}(c).
Both deployments use homogeneous robot teams that share a near-identical sensor viewpoint, so the descriptor association rarely produces the large geometric mismatch that the truncated MSE is designed to reject.
The stage nonetheless provides a clear benefit when the inter-robot viewpoint diverges, which we examine in detail in the supplementary material.\footnote{\href{https://sparolab.github.io/research/skid_slam/}{\texttt{https://sparolab.github.io/research/skid\_slam/}}}

This regime arises most naturally in heterogeneous robot teams, where ground and aerial agents observe a shared place from markedly different perspectives, and we revisit this direction in \Cref{sec:limitation}.

\begin{figure*}[!htbp]
    \vspace{-3mm}
    \centering
    \captionsetup{justification=justified}
    \captionsetup[subfigure]{justification=centering}
    \begin{subfigure}{0.66\columnwidth}
        \includegraphics[trim={0 0 0 0},clip, width=\columnwidth]{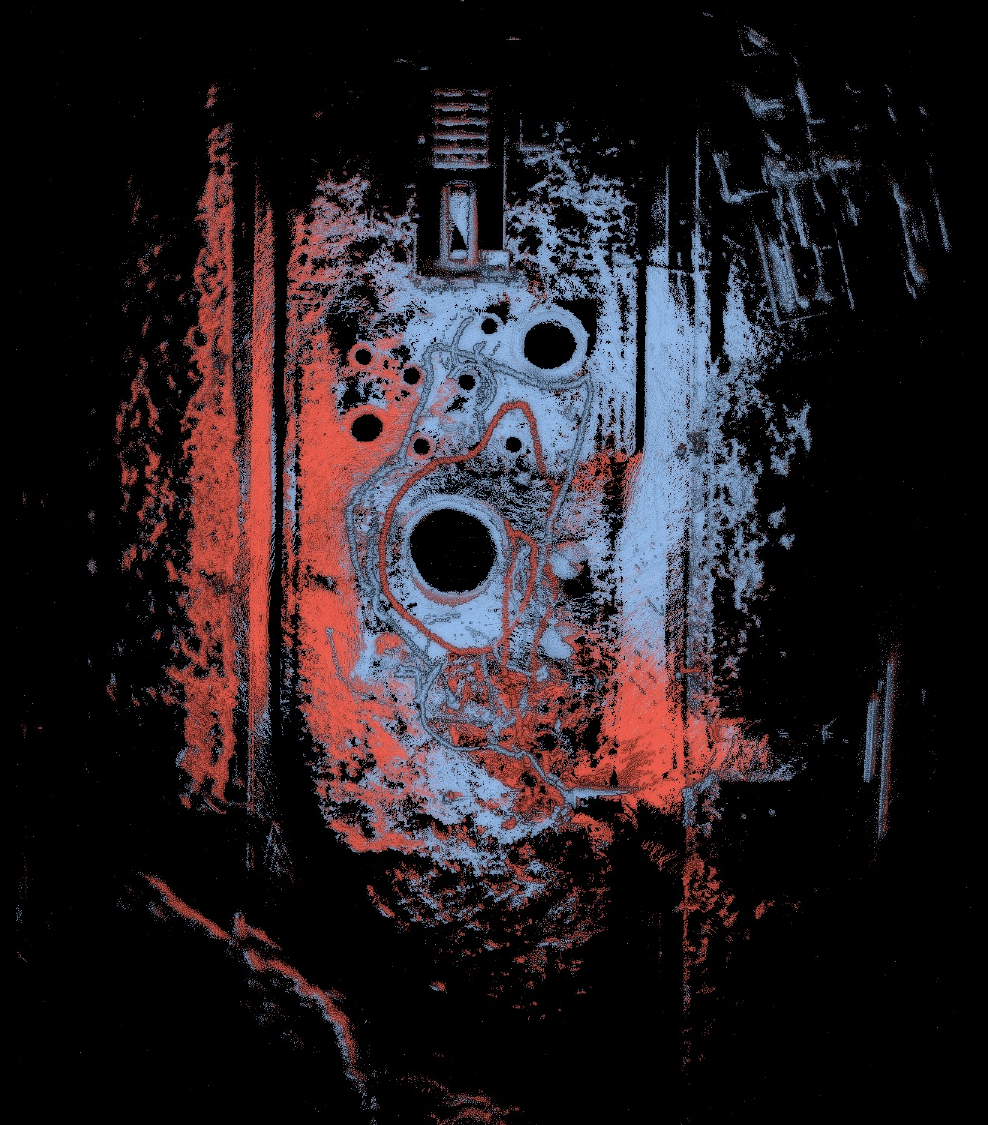}
        \caption{$36^\circ$ FOV (10$\%$)}
    \end{subfigure}
    \begin{subfigure}{0.66\columnwidth}
        \includegraphics[trim={0 0 0 0},clip, width=\columnwidth]{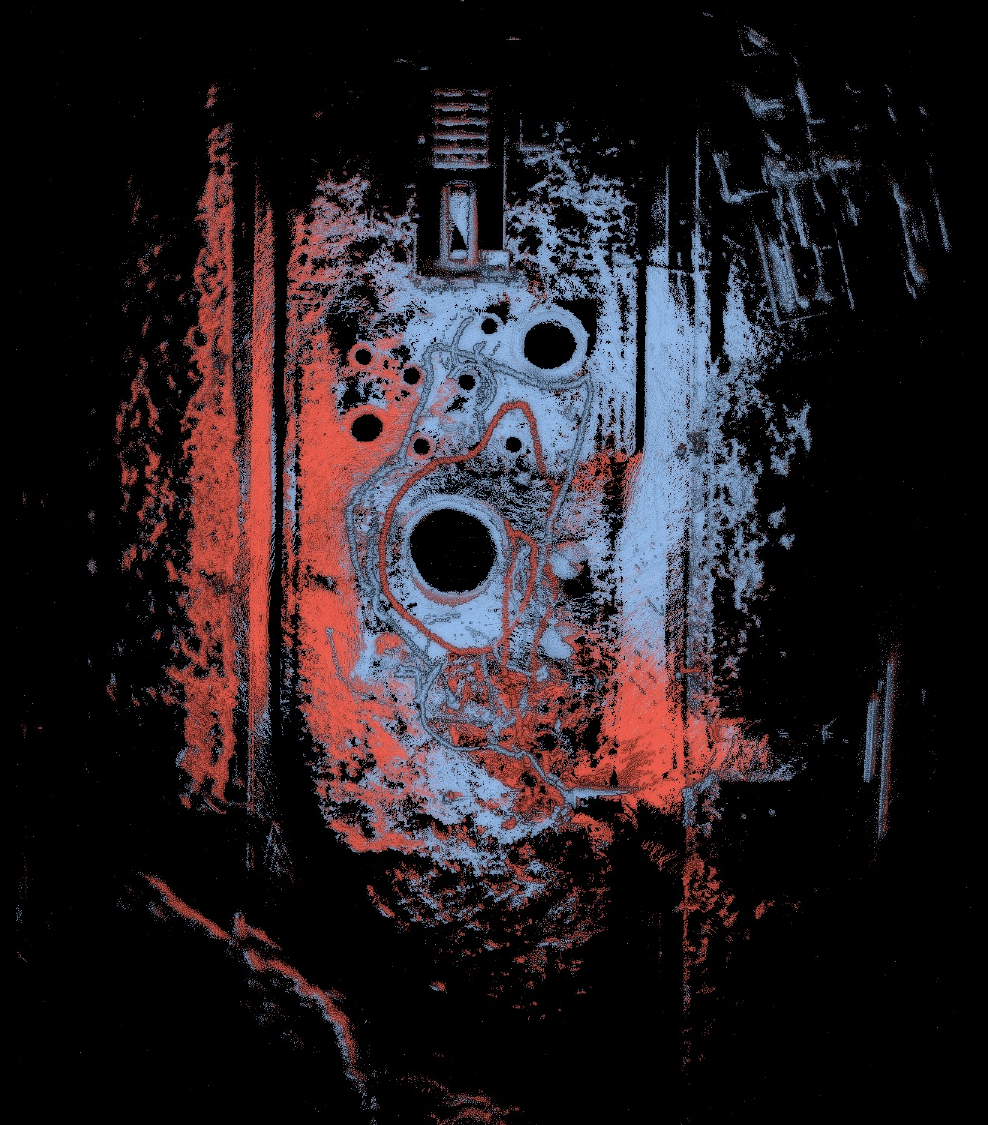}
        \caption{$72^\circ$ FOV (20$\%$)}
    \end{subfigure}
    \begin{subfigure}{0.66\columnwidth}
        \includegraphics[trim={0 0 0 0},clip, width=\columnwidth]{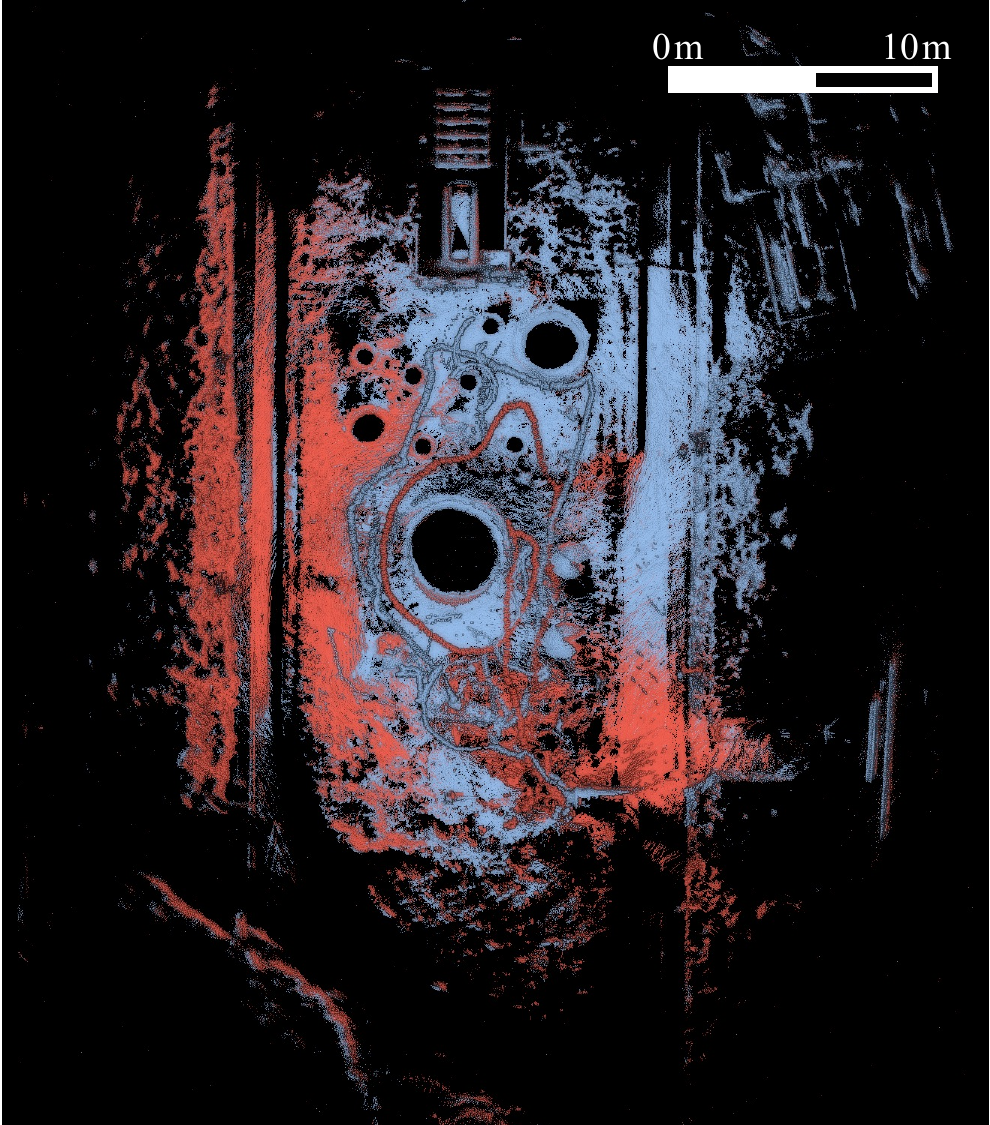}
        \caption{$108^\circ$ FOV (30$\%$)}
    \end{subfigure}
    \caption{Our SKiD-SLAM performance on the Aerial sequence with point cloud clipping to simulate ultra-low overlap. The point clouds are cropped to forward-facing FOVs of (a)~$36^\circ$ (10$\%$ of a full $360^\circ$ scan), (b)~$72^\circ$ (20$\%$), and (c)~$108^\circ$ (30$\%$).}
    \label{fig:fig21_overlap}
    \vspace{-0.5cm}
\end{figure*}
\begin{figure*}[!htbp]
    \centering
    \captionsetup{justification=justified}
    \captionsetup[subfigure]{justification=centering}
    \begin{subfigure}{0.99\columnwidth}
        \includegraphics[trim={0 0 0 0},clip, width=\columnwidth]{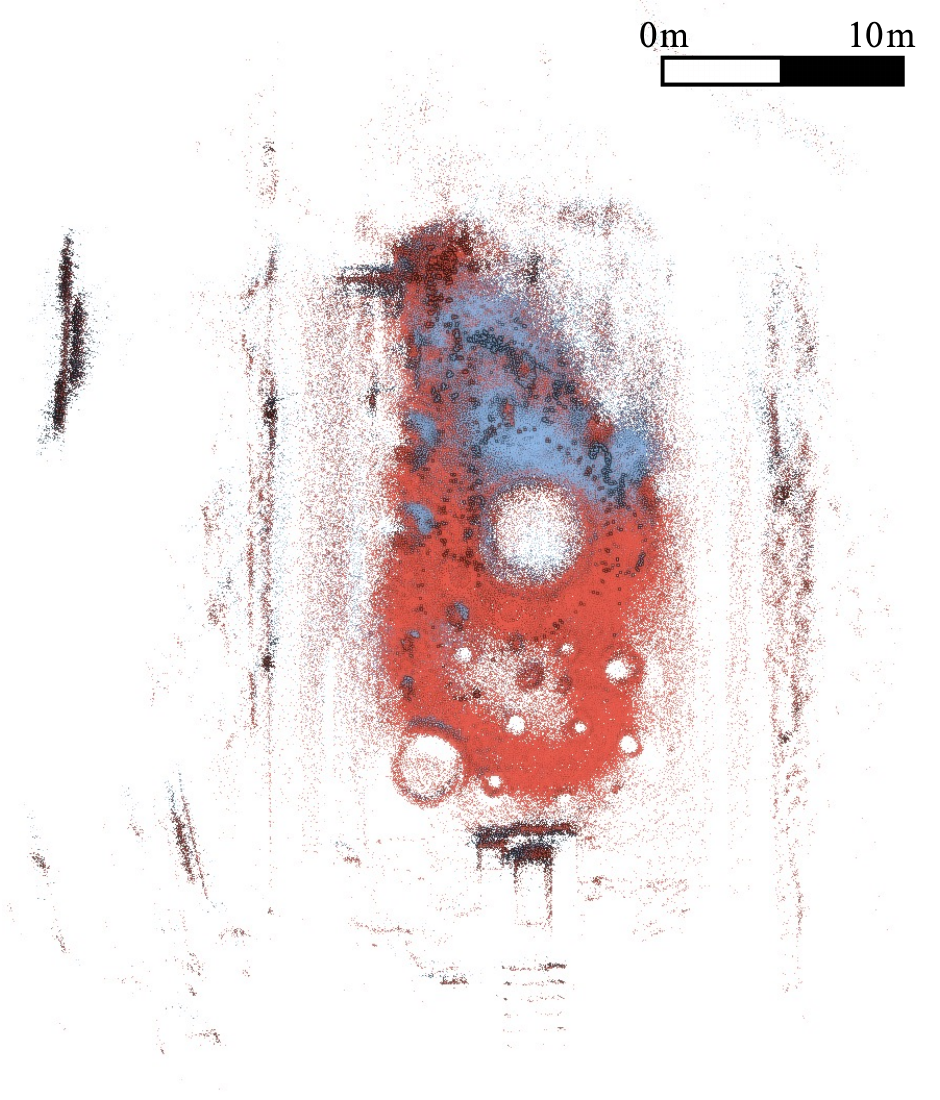}
        \vspace{-8mm}
        \caption{Delay = 2\,ms}
    \end{subfigure}
    \begin{subfigure}{0.99\columnwidth}
        \includegraphics[trim={0 0 0 0},clip, width=\columnwidth]{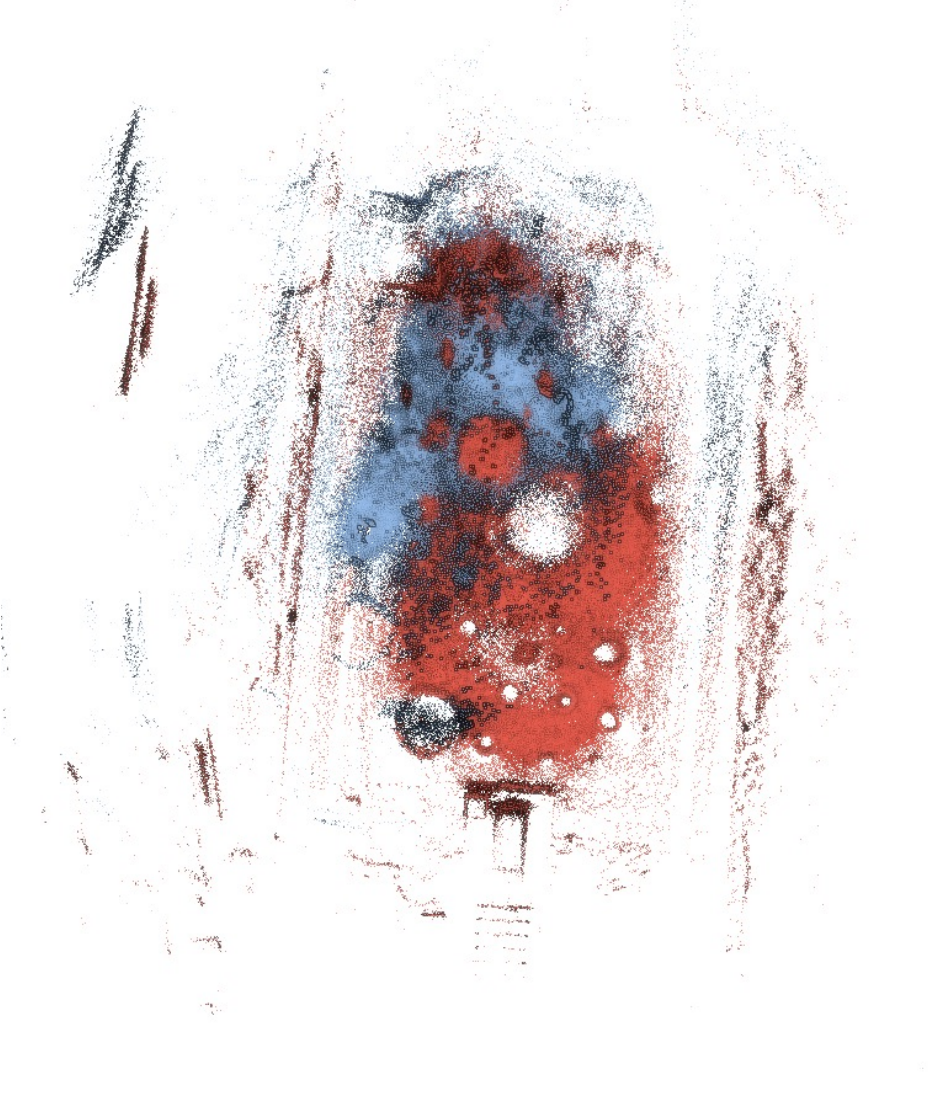}
        \vspace{-8mm}
        \caption{Delay = 200\,ms}
    \end{subfigure}
    \begin{subfigure}{0.99\columnwidth}
        \includegraphics[trim={0 0 0 0},clip, width=\columnwidth]{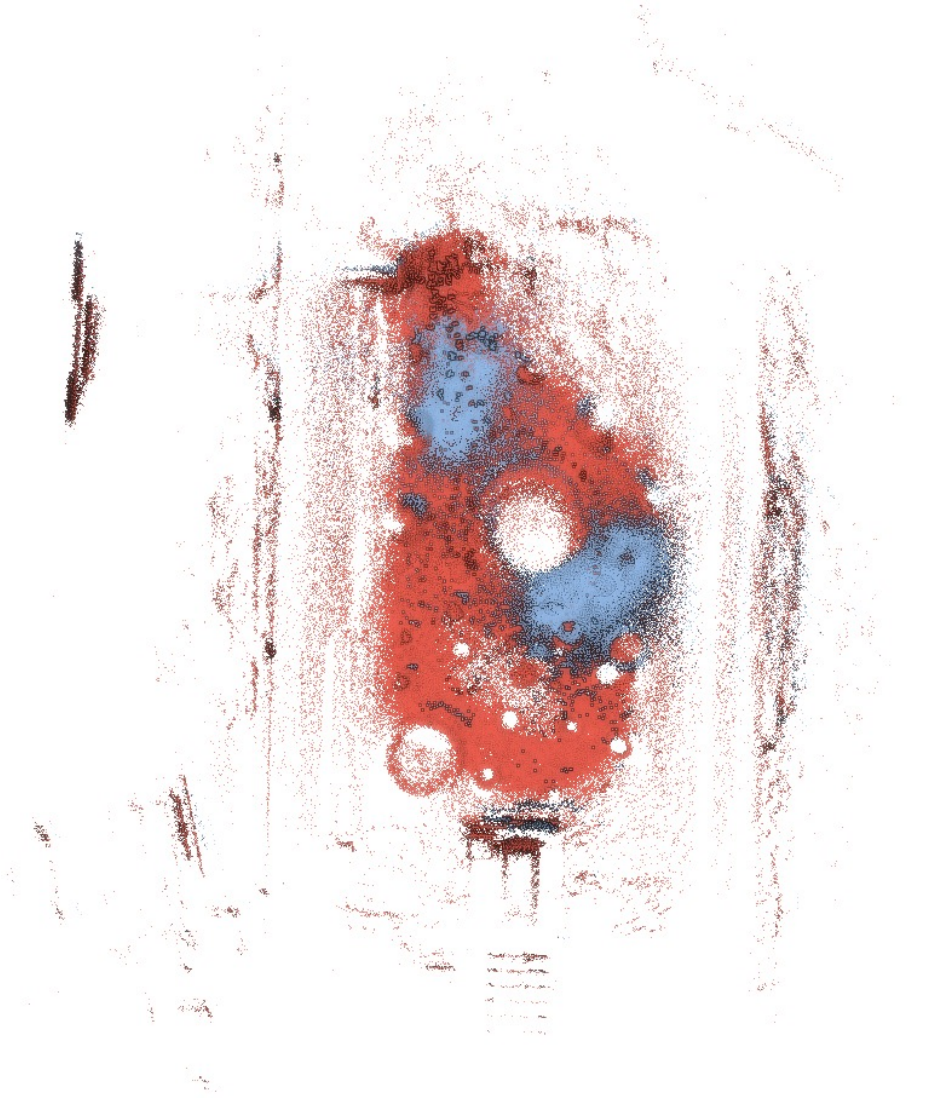}
        \vspace{-8mm}
        \caption{Packet loss = 50\%}
    \end{subfigure}
    \begin{subfigure}{0.99\columnwidth}
        \includegraphics[trim={0 0 0 0},clip, width=\columnwidth]{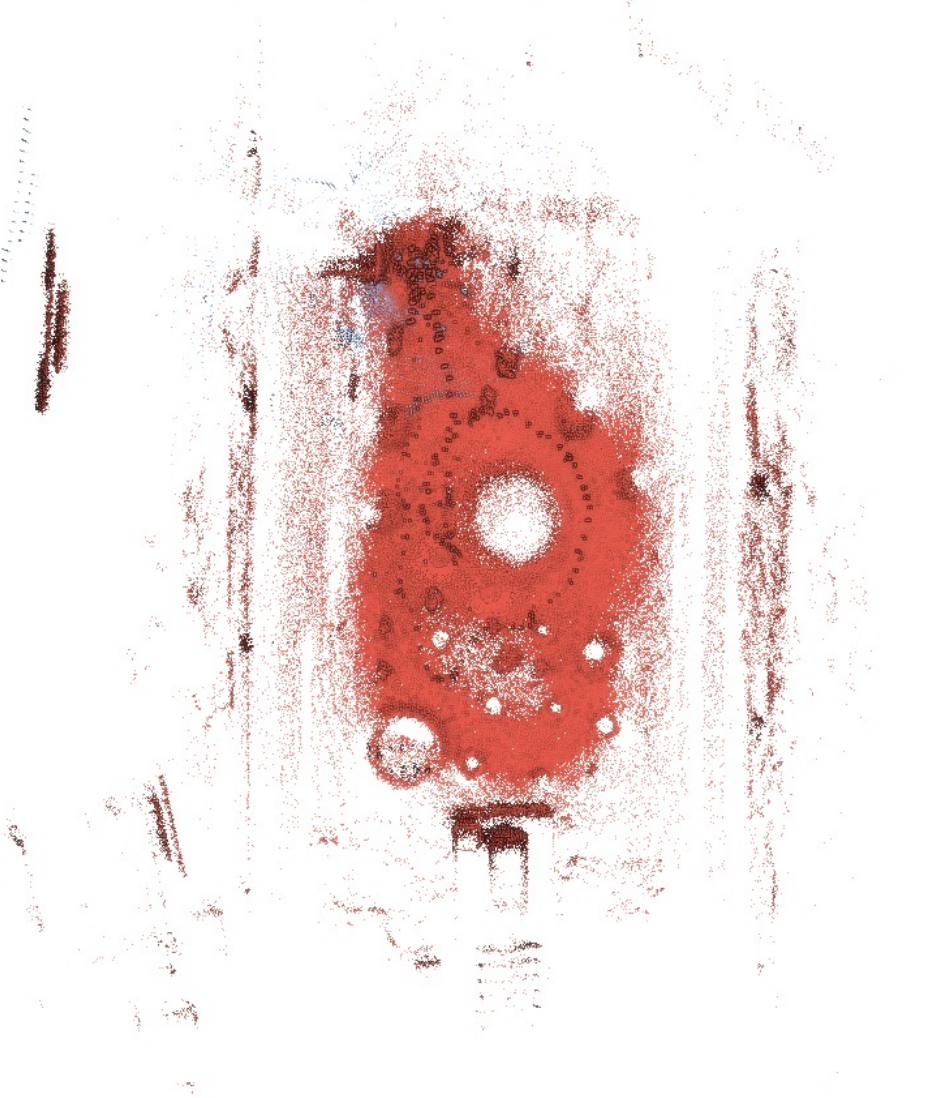}
        \vspace{-8mm}
        \caption{Packet loss = 75\%}
    \end{subfigure}
    \caption{Robustness evaluation of SKiD-SLAM under network impairments at the KICT planetary analog terrain using NetEm.
    (a)~With 2\,ms delay, maps merge successfully.
    (b)~With 200\,ms delay, map merging fails due to delayed loop closure detection.
    (c)~With 50\% packet loss, redundant keyframe exchange enables successful merging.
    (d)~With 75\% packet loss, insufficient data prevents merging.}
    \label{fig:fig22_network_test}
\end{figure*}
\begin{figure*}[!htbp]
    \vspace{-0.5cm}
    \centering
    \captionsetup{justification=justified}
    \captionsetup[subfigure]{justification=centering}
    \begin{subfigure}{\columnwidth}
        \includegraphics[trim={0 0 0 0},clip, width=\columnwidth]{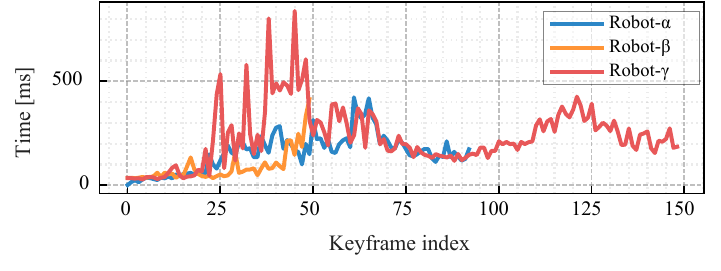}
        \caption{}
    \end{subfigure}
    \begin{subfigure}{0.99\columnwidth}
        \includegraphics[trim={0 0 0 0},clip, width=\columnwidth]{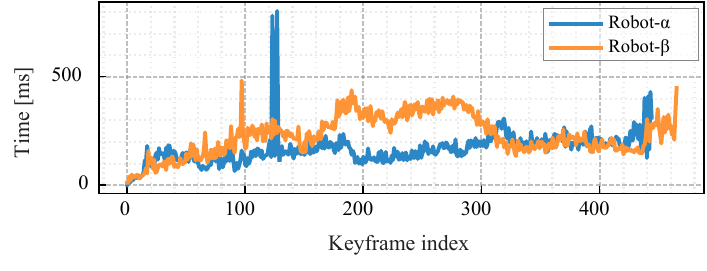}
        \caption{}
    \end{subfigure}
    \caption{Per-keyframe processing time of SKiD-SLAM for each robot on the embedded onboard computer at the two field sites: (a)~the Majang lava-tube cave (three robots: $\alpha$, $\beta$, and $\gamma$) and (b)~the KICT planetary analog terrain (two robots: $\alpha$ and $\beta$).}
    \label{fig:fig23_runtime}
    \vspace{-0.5cm}
\end{figure*}

\subsection{Effect of the PCM}
\label{sec:ablation:pcm}
One notable aspect is that PCM contributes on both planetary sites, unlike the preceding modules whose effect is site-dependent.
Removing PCM degrades the merge in the Majang lava-tube cave and in the KICT crater terrain alike, as shown in Fig.~\ref{fig:fig19_ablation_majang}(d) and Fig.~\ref{fig:fig20_ablation_yeoncheon}(d).
The effect is most pronounced on the crater terrain, where the larger map scale admits more candidate inter-robot loop closures and raises the chance that a false positive slips through the geometric verification.
PCM rejects these inconsistent loop closures before they enter the pose graph, which keeps the merged map globally consistent.

We conclude that PCM is a necessary safeguard for distributed map merging across both environments, rather than a refinement that matters only at a particular scale.

\subsection{Effect of Inter-Robot Overlap}
\label{sec:ablation:overlap}
Next, to test reliability under limited inter-robot overlap, we crop each robot's point cloud on the \texttt{Aerial} sequence from the GEODE dataset~\cite{chen2026heterogeneous} to a forward-facing FOV of 36$^\circ$, 72$^\circ$, or 108$^\circ$, corresponding to 10\%, 20\%, and 30\% of a full $360^\circ$ scan, respectively.
As shown in Fig.~\ref{fig:fig21_overlap}, our SKiD-SLAM still merges the multi-robot maps into a consistent global map at every tested FOV, including the narrowest 36$^\circ$ FOV.

This robustness to limited overlap makes the system well suited to recently emerging solid-state LiDARs, whose narrow field of view yields naturally low inter-robot overlap.
Such sensors are attractive for close-range planetary survey tasks, where dense reconstruction of rock and surface targets is required, and we expect SKiD-SLAM to extend to these platforms with minimal modification.

\subsection{Communication Robustness}
\label{sec:ablation:communication}
Beyond the limited bandwidth considered so far, planetary communication links remain unpredictable, exhibiting variable latency and packet loss as well.
To assess SKiD-SLAM's robustness to these additional impairments, we extend our evaluation by using NetEm~\cite{hemminger2005network} to emulate a range of degraded network conditions at the KICT planetary analog terrain.

\textbf{Network Delay:} We first test with 2\,ms and 200\,ms delays (Figs.~\ref{fig:fig22_network_test}(a) and \ref{fig:fig22_network_test}(b), respectively).
With a 2\,ms delay, typical of local robot-to-robot communication, the system successfully merges maps.
However, with a 200\,ms delay, map merging fails; the observed behavior is consistent with delayed descriptor exchange preventing timely inter-robot loop closure detection.
This indicates our system is designed for local multi-robot coordination rather than Earth-to-Mars operation.

\textbf{Packet Loss:} Next, we evaluate with 50\% and 75\% packet loss rates (Figs.~\ref{fig:fig22_network_test}(c) and \ref{fig:fig22_network_test}(d), respectively).
At 50\% packet loss, the system still completes map merging successfully, demonstrating resilience through redundant keyframe exchange.
At 75\% packet loss, insufficient descriptors and scans reach other robots, preventing successful merging.

In the tested conditions, SKiD-SLAM merges maps with a 2\,ms delay and 50\% packet loss, but not with a 200\,ms delay or 75\% packet loss.
This robustness stems from the lightweight SOLiD descriptor, which enables frequent updates under constrained bandwidth, and from redundant keyframe transmission, which provides resilience to packet loss.

\subsection{Computational Burden}
\label{sec:ablation:computational}
Finally, we measure the per-keyframe processing time of the complete SKiD-SLAM pipeline on the embedded platform.
As shown in Fig.~\ref{fig:fig23_runtime}, the measurements on the Intel NUC 11 Pro capture the full-pipeline computational demand at both field sites.
While the average per-keyframe processing time stays around 100--200\,ms, peak times occasionally exceed 500\,ms.

This confirms that the complete pipeline ran onboard the Intel NUC without external compute; \Cref{sec:limitation} discusses the remaining gap to flight-qualified processors.

\section{Lessons Learned}
\label{sec:lessons}
Beyond the quantitative and qualitative results above, our field deployments surfaced several operational lessons that are not visible in benchmark evaluation.
We report them here, as they shape how distributed multi-robot SLAM should be designed and operated for planetary missions.

\subsection{Communication Is More Than Bandwidth}
Our system was designed to respect a fixed inter-robot bandwidth budget, and \figref{fig:fig16_bandwidth} confirms that it operates within it.
In the field, however, we found that staying within the bandwidth budget is necessary but not sufficient.
The link itself was neither continuous nor symmetric.
In the Majang cave, the curved passages and self-similar walls repeatedly placed robots out of line of sight of one another, so the connection dropped intermittently regardless of the nominal data rate.
In addition, the shared wireless medium was subject to contention from teleoperation links and other equipment operating nearby, introducing interference that no per-robot rate budget captures.
These observations echo the design of prior missions, where ARCHES deployed a dedicated relay rover to maintain connectivity~\cite{schuster2020arches} and the CADRE rovers cannot transmit continuously because of drive-motor interference~\cite{saboia2024cadre}.
The lesson is that a deployable system must tolerate not only a low data rate but also intermittent, non-line-of-sight links affected by competition for the shared wireless channel; budget compliance and link robustness are distinct requirements, and both must be met.

\subsection{The Gap Between Analog and Extraterrestrial Conditions}
Our deployments reproduce the geometry of planetary caves and cratered terrain, but a terrestrial analog cannot reproduce the physical environment of an actual mission, and several of these absent factors bear directly on the sensors and mechanisms our system depends on.
The first is gravity.
Both sites operate under full Earth gravity, whereas reduced gravity changes wheel-terrain interaction, the settling behavior of disturbed material, and the dynamics of every platform, none of which our field data captures.
The second is the surface medium itself.
Lunar and Martian regolith carries fine, abrasive, and electrostatically charged dust that can coat optics, scatter or absorb LiDAR returns, and infiltrate moving parts, so a sensor that performs well in a damp Earth cave is not guaranteed to survive the same task in a harsh, dust-laden environment.
The third is the thin or absent atmosphere.
Mars's tenuous atmosphere offers little convective cooling, while lunar vacuum eliminates it; both environments impose severe thermal cycling beyond typical commercial ratings.
Validating a perception system for these conditions therefore requires testing that an analog site cannot provide, such as thermal-vacuum chamber trials and dust-exposure or ingress-protection assessments of the sensor payload, which we identify as necessary steps between the field results reported here and a flight-representative deployment.

\subsection{Toward Multimodal Perception Beyond Geometry}
Although our current pipeline relies solely on LiDAR geometry to remain lightweight on embedded hardware, the field deployments also highlighted the potential value of complementary sensing modalities. 
Because the Majang lava tube receives no natural light, perception in such environments may benefit from illumination-independent modalities, such as thermal imaging for material-dependent thermal contrast and ground-penetrating radar for subsurface structure.
We recorded complementary thermal imagery alongside our LiDAR data during the campaign using the thermal camera mounted on our platforms, and release it with this report.
This dataset allows the community to investigate whether such cues improve place recognition and registration in these environments.
Combining illumination-independent modalities with the geometric pipeline presented here is, in our view, a natural next step beyond the LiDAR-only perception used in this report.

\subsection{Toward Foundation-Model-Driven Autonomy}
A real mission also demands a level of autonomy that ground supervision cannot supply, since multi-minute communication delays and extended blackouts rule out remote intervention.
Recent work demonstrates robot foundation models navigating planetary-analog terrain by fusing learned policies with physics-based safety, reducing failure rates while retaining semantic navigation on a Mars-analog testbed~\citep{thakker2025risk}.
Vision foundation models likewise begin to overcome the scarcity of labeled planetary data, for instance by segmenting rocks across synthetic and real Martian domains to localize a rover within an aerial map collected by an aerial teammate~\citep{holden2025vision}.
Such ground-aerial arrangements point toward heterogeneous robot teams in which agents with complementary viewpoints share a common map, a setting where the viewpoint-robust verification we examine in \Cref{sec:ablation:tmse} becomes essential.
Coupling the geometric distributed SLAM presented here with these learned, multimodal priors is, in our view, the more compelling long-term direction for planetary autonomy.

\section{Limitations and Future Works}
\label{sec:limitation}
Our field deployments revealed several limitations that inform future directions for distributed multi-robot SLAM in planetary missions.

\subsection{Heterogeneous Team Constraints}
Our field tests deployed homogeneous teams at each site: wheeled rovers in the Majang lava-tube cave and quadruped robots at the KICT terrain.
Consequently, cross-platform inter-robot loop closure remains unverified under field conditions.
While SOLiD demonstrates robustness to moderate viewpoint variations, it was not designed for the substantial mounting differences between wheeled and legged platforms, where LiDAR sensors may be positioned at different heights and orientations.
Such heterogeneous configurations, common in proposed planetary missions like ARCHES~\cite{schuster2020arches}, require place recognition methods that remain invariant across platform-induced perspective changes.
Future work should develop descriptors specifically designed for cross-platform matching and conduct field campaigns that explicitly test mixed-platform teams.

\subsection{Theoretical Guarantees}
SKiD-SLAM as a whole lacks formal robustness guarantees under arbitrary measurement outliers and network conditions.
PCM is effective in practice, but its outcome depends on the validity of the pairwise consistency tests and on the available loop-closure graph.
Similarly, the truncated-MSE threshold used in our geometric-verification stage is selected empirically rather than derived from principled analysis.
These heuristic choices work reliably in our field tests but may fail under conditions not represented in our deployments.
Establishing theoretical foundations would strengthen confidence for deployment in irreversible planetary missions where failure recovery is impossible.
Future research should derive robustness conditions for distributed outlier rejection and develop principled methods for adaptive threshold selection based on environmental characteristics.

\subsection{Computational Constraints}
Our field deployments used Intel NUC computers, compact onboard platforms that ran the complete SKiD-SLAM pipeline with average processing times of 100--200\,ms but occasional peaks exceeding 500\,ms during inter-robot loop closures.
While these represent a reasonable projection for future COTS processors in space, they exceed the computational resources of current radiation-hardened flight hardware.
The memory footprint, particularly for storing scan descriptors and keyframe data, grows with trajectory length and may challenge resource-constrained systems on extended missions.
This gap between our field hardware and current flight-qualified processors highlights the need for continued optimization.
Future work should focus on reducing computational demands through selective keyframe retention, adaptive descriptor compression, and hardware-aware algorithm design to enable deployment on more constrained platforms.

\begin{figure*}[t]
    \centering
    \def\figwidth{\textwidth}%
        {%
        \includegraphics[clip, trim= 15 0 0 0, width=\figwidth]{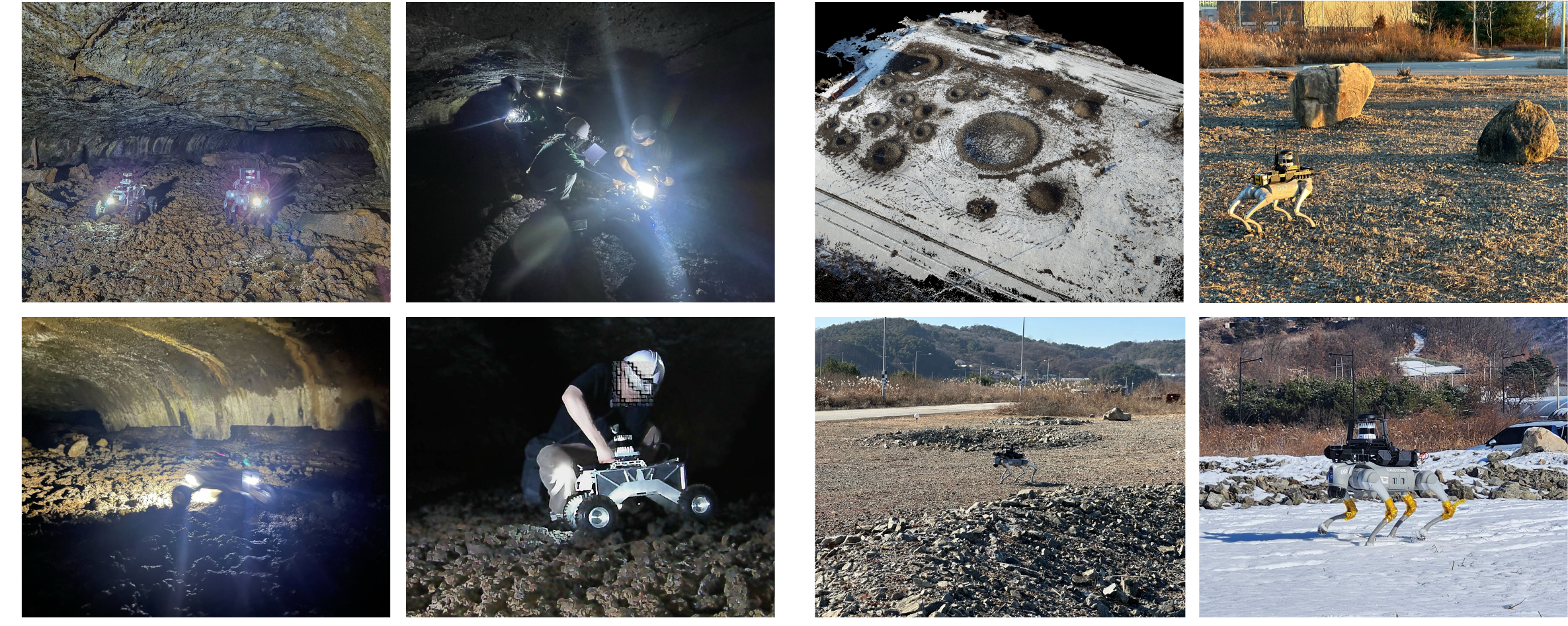}
        }
        \vspace{-0.4cm}
        \captionsetup{justification=justified}
        \caption{Representative scenes from the field deployments at the Majang lava-tube cave and the KICT planetary analog terrain.}
        \label{fig:fig24_field}
        \vspace{-0.3cm}
\end{figure*}

\subsection{Future Directions}
Beyond addressing these limitations, several directions merit investigation.
First, extending SKiD-SLAM to handle dynamic team reconfiguration, where robots join or leave the network during operation, would increase mission flexibility.
Second, incorporating semantic information could improve loop closure reliability in geometrically ambiguous environments.
Third, developing adaptive communication strategies that adjust descriptor exchange rates based on available bandwidth and mission phase could better utilize constrained links.
Fourth, future work will include benchmark evaluations with $N \geq 3$ robots and a systematic graded perceptual-aliasing study using controlled aliasing levels and survey-grade ground truth.
Finally, validation in more extreme analog environments, such as Antarctica's dry valleys or volcanic terrain, would strengthen confidence in the system's planetary readiness.
These advances would bring distributed multi-robot SLAM closer to the reliability required for planetary deployment.

\section{Conclusion}
\label{sec:conclusion}
This field report presented SKiD-SLAM, a distributed multi-robot LiDAR SLAM system designed for the severe communication and computation constraints of planetary missions.
We demonstrated its deployment across two representative mission concepts at planetary analog sites in South Korea, as illustrated in \figref{fig:fig24_field}: \caveconcept in the Majang lava-tube cave and \surfaceconcept at the KICT planetary analog terrain.
Our approach treated distributed SLAM as a systems integration challenge governed by planetary constraints, selecting and tailoring three modules, place recognition, registration, and distributed optimization, to operate within a 250\,kbps inter-robot bandwidth budget and embedded-class computational resources.

Our evaluation progressed from controlled benchmarks to field deployment.
The SOLiD descriptor achieved comparable accuracy to state-of-the-art methods while reducing descriptor-memory requirements by factors ranging from 6.4 to 720 across the compared methods, and KISS-Matcher provided robust registration without requiring initial alignment.
Field deployments validated system performance in challenging conditions: our SKiD-SLAM successfully merged robot maps in both the GPS-denied, lightless cave environment and the geometrically sparse crater terrain.
Critically, our SKiD-SLAM was the only evaluated method to achieve map merging within the planetary communication budget, while the baseline methods failed to complete map merging under the same bandwidth limit.

The field deployments revealed insights beyond controlled evaluation.
Scan drop ratios rose sharply as the imposed bandwidth decreased.
The geometric self-similarity of natural environments challenged place recognition in ways that urban benchmarks do not capture.
These findings underscore the importance of field validation for systems intended for planetary deployment.

This work demonstrates the feasibility of distributed multi-robot SLAM at terrestrial planetary analog sites under an imposed 250\,kbps communication budget.
Our released code and field datasets provide the community with resources to further develop and validate distributed SLAM systems for extreme environments.
Future planetary missions will require continued progress in cross-platform coordination, theoretical guarantees, and embedded optimization to enable robust multi-robot exploration of environments inaccessible to human exploration.

\section*{Acknowledgment}
We thank the Unmanned Exploration Laboratory~(UEL) for supporting our access to the field sites, \ie the Majang lava-tube cave and the KICT planetary analog terrain in Yeoncheon.
We also thank Prof. Sanghyun Kim at Kyung Hee University and Prof. Young-Sik Shin at Kyungpook National University for fruitful discussions on our field deployments.

\footnotesize
\begingroup
\setlength{\itemsep}{-0.3em}
\bibliographystyle{Packages/IEEEtranN}
\bibliography{Packages/string-short, Packages/references}

@misc{NASA_CADRE,
  author       = {{NASA}},
  title        = {{CADRE}: Cooperative Autonomous Distributed Robotic Exploration},
  howpublished = {[Online]. Available: \url{https://www.jpl.nasa.gov/missions/cadre/}},
  year         = {2025},
  note         = {Accessed: May 2, 2026},
}

@inproceedings{delaCroix2024CADRE,
  author    = {de la Croix, Jean-Pierre and Rossi, Federico and Brockers, Roland and Aguilar, Dustin and Albee, Keenan and Boroson, Elizabeth and Cauligi, Abhishek and Delaune, Jeff and Hewitt, Robert and Kogan, Dima and Lim, Grace and Morrell, Benjamin and Nakka, Yashwanth and Nguyen, Viet and Proen{\c{c}}a, Pedro and Rabideau, Gregg and Russino, Joseph and Saboia da Silva, Maira and Zohar, Guy and Comandur, Subha},
  title     = {Multi-Agent Autonomy for Space Exploration on the {CADRE} Lunar Technology Demonstration},
  booktitle = {2024 IEEE Aerospace Conference},
  year      = {2024},
  pages     = {1--14},
  doi       = {10.1109/AERO58975.2024.10521425},
}

@article{rabideau2025planning,
  title={Planning, scheduling, and execution on the {Moon}: the {CADRE} technology demonstration mission},
  author={Rabideau, Gregg and Russino, Joseph and Branch, Andrew and Dhamani, Nihal and Vaquero, Tiago Stegun and Chien, Steve and de la Croix, Jean-Pierre and Rossi, Federico},
  journal={arXiv preprint arXiv:2502.14803},
  year={2025}
}

@inproceedings{saboia2024cadre,
  title={{CADRE MoonDB}: Distributed database for multi-robot information-sharing and map-merging for lunar exploration},
  author={Saboia, Maira and Rossi, Federico and Nguyen, Viet and Lim, Grace and Aguilar, Dustin and de la Croix, Jean-Pierre},
  booktitle={Int. Workshop on Autonomous Agents and Multi-Agent Systems for Space Applications (MASSpace). Auckland, NZ},
  year={2024}
}

@article{stathoulopoulos2023frame,
  title={{FRAME}: Fast and robust autonomous {3D} point cloud map-merging for egocentric multi-robot exploration},
  author={Stathoulopoulos, Nikolaos and Koval, Anton and Agha-mohammadi, Ali-akbar and Nikolakopoulos, George},
  journal={arXiv preprint arXiv:2301.09213},
  year={2023}
}

@article{stathoulopoulos2024frame,
  title={{FRAME}: A Modular Framework for Autonomous Map-merging: Advancements in the Field},
  author={Stathoulopoulos, Nikolaos and Lindqvist, Bj{\"o}rn and Koval, Anton and Agha-Mohammadi, Ali-Akbar and Nikolakopoulos, George},
  journal=IEEE_J_RO,
  year={2024},
  publisher={IEEE}
}

@article{lajoie2025multi,
  title={Multi-Robot Decentralized Collaborative {SLAM} in Planetary Analogue Environments: Dataset, Challenges, and Lessons Learned},
  author={Lajoie, Pierre-Yves and Soma, Karthik and Bong, Haechan Mark and Lemieux-Bourque, Alice and Zhang, Rongge and Varadharajan, Vivek Shankar and Beltrame, Giovanni},
  journal={IEEE Transactions on Field Robotics},
  year={2025},
  publisher={IEEE}
}

@article{varadharajan2025multi,
  title={A multi-robot exploration planner for space applications},
  author={Varadharajan, Vivek Shankar and Beltrame, Giovanni},
  journal={IEEE Robotics and Automation Letters},
  year={2025},
  publisher={IEEE}
}

@article{jeong2024diter,
  title={{DiTer}: Diverse terrain and multimodal dataset for field robot navigation in outdoor environments},
  author={Jeong, Seokhwan and Kim, Hogyun and Cho, Younggun},
  journal={IEEE Sensors Letters},
  volume={8},
  number={3},
  pages={1--4},
  year={2024},
  publisher={IEEE}
}

@inproceedings{kim2025diter++,
  title={{DiTer++}: Diverse terrain and multi-modal dataset for multi-robot {SLAM} in multi-session environments},
  author={Kim, Juwon and Kim, Hogyun and Jeong, Seokhwan and Shin, Youngsik and Cho, Younggun},
  booktitle={2025 IEEE International Conference on Robotics and Automation (ICRA)},
  pages={12187--12193},
  year={2025},
  organization={IEEE}
}

@article{giubilato2026s3li,
  title={The {S3LI Vulcano} Dataset: A Dataset for Multi-Modal {SLAM} in Unstructured Planetary Environments},
  author={Giubilato, Riccardo and M{\"u}ller, Marcus Gerhard and Sewtz, Marco and Gonzalez, Laura Alejandra Encinar and Folkesson, John and Triebel, Rudolph},
  journal={arXiv preprint arXiv:2601.19557},
  year={2026}
}

@article{chang2022lamp,
  title={{LAMP 2.0}: A robust multi-robot {SLAM} system for operation in challenging large-scale underground environments},
  author={Chang, Yun and Ebadi, Kamak and Denniston, Christopher E and Ginting, Muhammad Fadhil and Rosinol, Antoni and Reinke, Andrzej and Palieri, Matteo and Shi, Jingnan and Chatterjee, Arghya and Morrell, Benjamin and others},
  journal={IEEE Robotics and Automation Letters},
  volume={7},
  number={4},
  pages={9175--9182},
  year={2022},
  publisher={IEEE}
}

@inproceedings{kim2022lt,
  title={{LT-mapper}: A modular framework for {LiDAR}-based lifelong mapping},
  author={Kim, Giseop and Kim, Ayoung},
  booktitle=C-ICRA,
  pages={7995--8002},
  year={2022},
  organization={IEEE}
}

@inproceedings{gil2025ephemerality,
  title={Ephemerality meets {LiDAR}-based lifelong mapping},
  author={Gil, Hyeonjae and Lee, Dongjae and Kim, Giseop and Kim, Ayoung},
  booktitle={2025 IEEE International Conference on Robotics and Automation (ICRA)},
  pages={3312--3319},
  year={2025},
  organization={IEEE}
}

@article{damigos2024communication,
  title={Communication-aware control of large data transmissions via centralized cognition and {5G} networks for multi-robot map merging},
  author={Damigos, Gerasimos and Stathoulopoulos, Nikolaos and Koval, Anton and Lindgren, Tore and Nikolakopoulos, George},
  journal={Journal of Intelligent \& Robotic Systems},
  volume={110},
  number={1},
  pages={22},
  year={2024},
  publisher={Springer}
}

@article{kang2025uni,
  title={{Uni-Mapper}: Unified Mapping Framework for Multi-Modal {LiDARs} in Complex and Dynamic Environments},
  author={Kang, Gilhwan and Kim, Hogyun and Choi, Byunghee and Jeong, Seokhwan and Shin, Young-Sik and Cho, Younggun},
  journal={Transactions on Intelligent Vehicles},
  volume={11},
  number={1},
  pages={104--121},
  year={2026}
}

@article{wei2024large,
  title={Large-scale multi-session point-cloud map merging},
  author={Wei, Hairuo and Li, Rundong and Cai, Yixi and Yuan, Chongjian and Ren, Yunfan and Zou, Zuhao and Wu, Huajie and Zheng, Chunran and Zhou, Shunbo and Xue, Kaiwen and others},
  journal={IEEE Robotics and Automation Letters},
  volume={10},
  number={1},
  pages={88--95},
  year={2024},
  publisher={IEEE}
}

@article{lim2025multi,
  title={{Multi-Mapcher}: Loop Closure Detection-Free Heterogeneous {LiDAR} Multi-Session {SLAM} Leveraging Outlier-Robust Registration for Autonomous Vehicles},
  author={Lim, Hyungtae and Kim, Daebeom and Myung, Hyun},
  journal={arXiv preprint arXiv:2511.00635},
  year={2025}
}

@article{huang2022disco,
  title={{DiSCo-SLAM}: Distributed {Scan Context}-enabled multi-robot {LiDAR} {SLAM} with two-stage global-local graph optimization},
  author={Huang, Yewei and Shan, Tixiao and Chen, Fanfei and Englot, Brendan},
  journal=IEEE_J_RAL,
  volume={7},
  number={2},
  pages={1150--1157},
  year={2022},
  publisher={IEEE}
}

@article{xie2021rdc,
  title={{RDC-SLAM}: A real-time distributed cooperative {SLAM} system based on {3D} {LiDAR}},
  author={Xie, Yuting and Zhang, Yachen and Chen, Long and Cheng, Hui and Tu, Wei and Cao, Dongpu and Li, Qingquan},
  journal=IEEE_J_ITS,
  volume={23},
  number={9},
  pages={14721--14730},
  year={2021},
  publisher={IEEE}
}

@article{zhong2023dcl,
  title={{DCL-SLAM}: A distributed collaborative {LiDAR} {SLAM} framework for a robotic swarm},
  author={Zhong, Shipeng and Qi, Yuhua and Chen, Zhiqiang and Wu, Jin and Chen, Hongbo and Liu, Ming},
  journal=IEEE_J_SENSOR,
  volume={24},
  number={4},
  pages={4786--4797},
  year={2023},
  publisher={IEEE}
}

@article{lajoie2023swarm,
  title={{Swarm-SLAM}: Sparse decentralized collaborative simultaneous localization and mapping framework for multi-robot systems},
  author={Lajoie, Pierre-Yves and Beltrame, Giovanni},
  journal=IEEE_J_RAL,
  volume={9},
  number={1},
  pages={475--482},
  year={2023},
  publisher={IEEE}
}

@article{lyu2025drcm,
  title={{DRCM-CSLAM}: Distributed Robust and Communication-efficient Multi-Robot Cooperative {LiDAR}-inertial {SLAM}},
  author={Lyu, Pin and Li, Jiong and Lai, Jizhou and Fang, Wei and Wang, Qian},
  journal={IEEE Transactions on Instrumentation and Measurement},
  year={2025},
  publisher={IEEE}
}

@article{wang2026communication,
  title={Communication-Robust Asynchronous Distributed {LiDAR} Collaborative Smoothing and Mapping},
  author={Wang, Jiancheng and Cai, Chenyuan and Tan, JinQian and Chen, Haoyao},
  journal={IEEE Robotics and Automation Letters},
  year={2026},
  publisher={IEEE}
}

@article{dominguez2025cooperative,
  title={Cooperative robotic exploration of a planetary skylight surface and lava cave},
  author={Dom{\'\i}nguez, Ra{\'u}l and P{\'e}rez-del-Pulgar, Carlos and Paz-Delgado, Gonzalo J and Polisano, Fabio and Babel, Jonathan and Germa, Thierry and Dragomir, Iulia and Ciarletti, Val{\'e}rie and Berthet, Anne-Claire and Danter, Leon Cedric and others},
  journal={Science Robotics},
  volume={10},
  number={105},
  pages={eadj9699},
  year={2025},
  publisher={American Association for the Advancement of Science}
}

@article{arm2023scientific,
  title={Scientific exploration of challenging planetary analog environments with a team of legged robots},
  author={Arm, Philip and Waibel, Gabriel and Preisig, Jan and Tuna, Turcan and Zhou, Ruyi and Bickel, Valentin and Ligeza, Gabriela and Miki, Takahiro and Kehl, Florian and Kolvenbach, Hendrik and others},
  journal={Science robotics},
  volume={8},
  number={80},
  pages={eade9548},
  year={2023},
  publisher={American Association for the Advancement of Science}
}

@article{krawciw2025sharing,
  title={Sharing the Load: Distributed Model-Predictive Control for Precise Multi-Rover Cargo Transport},
  author={Krawciw, Alexander and Lilge, Sven and Antonyshyn, Luka and Barfoot, Timothy D},
  journal={arXiv preprint arXiv:2510.18766},
  year={2025}
}

@inproceedings{nash2026software,
  title={A Software-Based Approach to Radiation Mitigation for Planetary Missions},
  author={Nash, Jeremy and Verma, Vandi and Graser, Evan and Wang, Haoda and Kubiak, Gerik and Maimone, Mark and Myint, Steven and Biesiadecki, Jeffrey and Romano, Philip},
  booktitle={2026 IEEE Aerospace Conference},
  pages={1--18},
  year={2026},
  organization={IEEE}
}

@article{tranzatto2022cerberus,
  title={{CERBERUS} in the {DARPA Subterranean Challenge}},
  author={Tranzatto, Marco and Miki, Takahiro and Dharmadhikari, Mihir and Bernreiter, Lukas and Kulkarni, Mihir and Mascarich, Frank and Andersson, Olov and Khattak, Shehryar and Hutter, Marco and Siegwart, Roland and others},
  journal={Science Robotics},
  volume={7},
  number={66},
  pages={eabp9742},
  year={2022},
  publisher={American Association for the Advancement of Science}
}

@article{agha2022nebula,
  title={{NeBula}: {TEAM CoSTAR}'s robotic autonomy solution that won {Phase II} of {DARPA Subterranean Challenge}},
  author={Agha, Ali and Otsu, Kyohei and Morrell, Benjamin and Fan, David D and Thakker, Rohan and Santamaria-Navarro, Angel and Kim, Sung-Kyun and Bouman, Amanda and Lei, Xianmei and Edlund, Jeffrey and others},
  journal={Field robotics},
  volume={2},
  pages={1432--1506},
  year={2022},
  publisher={FRPS}
}

@article{choudhary2017distributed,
  title={Distributed mapping with privacy and communication constraints: Lightweight algorithms and object-based models},
  author={Choudhary, Siddharth and Carlone, Luca and Nieto, Carlos and Rogers, John and Christensen, Henrik I and Dellaert, Frank},
  journal={The International Journal of Robotics Research},
  volume={36},
  number={12},
  pages={1286--1311},
  year={2017},
  publisher={SAGE Publications Sage UK: London, England}
}

@article{schuster2020arches,
  title={The {ARCHES} space-analogue demonstration mission: Towards heterogeneous teams of autonomous robots for collaborative scientific sampling in planetary exploration},
  author={Schuster, Martin J and M{\"u}ller, Marcus G and Brunner, Sebastian G and Lehner, Hannah and Lehner, Peter and Sakagami, Ryo and D{\"o}mel, Andreas and Meyer, Lukas and Vodermayer, Bernhard and Giubilato, Riccardo and others},
  journal={IEEE Robotics and Automation Letters},
  volume={5},
  number={4},
  pages={5315--5322},
  year={2020},
  publisher={IEEE}
}

@article{lee2025soft,
  title={Soft deployable airless wheel for lunar lava tube intact exploration},
  author={Lee, Seong-Bin and Cho, Namsuk and Lee, Geonho and Lee, Seungju and Kim, Junseo and Shim, Gyujin and Jang, Jong Tai and Kim, Se Kwon and Seo, TaeWon and Sim, Chae Kyung and others},
  journal={Science Robotics},
  volume={10},
  number={109},
  pages={eadx2549},
  year={2025},
  publisher={American Association for the Advancement of Science}
}

@misc{RerunSDK,
  title = {{Rerun}: A Visualization {SDK} for Multimodal Data},
  author = {{Rerun Development Team}},
  year = {2024},
  howpublished = {\url{https://www.rerun.io}},
  note = {Available from \url{https://www.rerun.io} and \url{https://github.com/rerun-io/rerun}}
}

@article{golombek1997overview,
  title={Overview of the {Mars Pathfinder} mission and assessment of landing site predictions},
  author={Golombek, MP and Cook, RA and Economou, Thanasis and Folkner, WM and Haldemann, AFC and Kallemeyn, PH and Knudsen, Jens Martin and Manning, RM and Moore, HJ and Parker, TJ and others},
  journal={Science},
  volume={278},
  number={5344},
  pages={1743--1748},
  year={1997},
  publisher={American Association for the Advancement of Science}
}

@article{crisp2003mars,
  title={{Mars Exploration Rover} mission},
  author={Crisp, Joy A and Adler, Mark and Matijevic, Jacob R and Squyres, Steven W and Arvidson, Raymond E and Kass, David M},
  journal={Journal of Geophysical Research: Planets},
  volume={108},
  number={E12},
  year={2003},
  publisher={Wiley Online Library}
}

@article{grotzinger2012mars,
  title={{Mars Science Laboratory} mission and science investigation},
  author={Grotzinger, John P and Crisp, Joy and Vasavada, Ashwin R and Anderson, Robert C and Baker, Charles J and Barry, Robert and Blake, David F and Conrad, Pamela and Edgett, Kenneth S and Ferdowski, Bobak and others},
  journal={Space science reviews},
  volume={170},
  number={1},
  pages={5--56},
  year={2012},
  publisher={Springer}
}

@article{farley2020mars,
  title={{Mars 2020} mission overview},
  author={Farley, Kenneth A and Williford, Kenneth H and Stack, Kathryn M and Bhartia, Rohit and Chen, Al and De La Torre, Manuel and Hand, Kevin and Goreva, Yulia and Herd, Christopher DK and Hueso, Ricardo and others},
  journal={Space Science Reviews},
  volume={216},
  number={8},
  pages={142},
  year={2020},
  publisher={Springer}
}

@article{li2021overview,
  title={Overview of the {Chang’e-4} mission: Opening the frontier of scientific exploration of the lunar far side},
  author={Li, Chunlai and Zuo, Wei and Wen, Weibin and Zeng, Xingguo and Gao, Xingye and Liu, Yuxuan and Fu, Qiang and Zhang, Zhoubin and Su, Yan and Ren, Xin and others},
  journal={Space Science Reviews},
  volume={217},
  number={2},
  pages={35},
  year={2021},
  publisher={Springer}
}

@misc{isro2023chandrayaan3,
  author       = {{ISRO}},
  title        = {{Chandrayaan-3}},
  year         = {2023},
  howpublished = {[Online]. Available: \url{https://www.isro.gov.in/Chandrayaan3_New.html}},
  note         = {Accessed: 2026-06-17},
}

@inproceedings{balaram2018mars,
  title={{Mars Helicopter} technology demonstrator},
  author={Balaram, Bob and Canham, Timothy and Duncan, Courtney and Grip, H{\aa}vard F and Johnson, Wayne and Maki, Justin and Quon, Amelia and Stern, Ryan and Zhu, David},
  booktitle={2018 AIAA atmospheric flight mechanics conference},
  pages={0023},
  year={2018}
}

@inproceedings{dettmann2022corob,
  title={{COROB-X}: A Cooperative robot team for the exploration of lunar skylights},
  author={Dettmann, Alexander and Voegele, Thomas and Oc{\'o}n, Jorge and Dragomir, Iulia and Govindaraj, Shashank and De Benedetti, Matteo and Ciarletti, Val{\'e}rie and Hassen-Khodja, Rafik and Germa, Thierry and Viards, Raphael and others},
  booktitle={ASTRA 2022 16th Symposium on Advanced Space Technologies in Robotics and Automation},
  year={2022}
}

@misc{jaxa2024slim,
  author       = {{JAXA}},
  title        = {Result and Achievements of the Lunar Excursion Vehicle ({LEV-1})
                  on board Smart Lander for Investigating {Moon} ({SLIM})},
  year         = {2024},
  howpublished = {[Online]. Available: \url{https://global.jaxa.jp/press/2024/01/20240125-2_e.html}},
  note         = {Accessed: 2026-06-17},
}

@inproceedings{inazawa2024autonomous,
  title={Autonomous Control of Lunar Excursion Vehicle 2 ({LEV-2})},
  author={Inazawa, Mariko and Hirano, Daichi and Sutoh, Masataku and Sawada, Hirotaka and Nagata, Masaharu and Sakoda, Gen},
  booktitle={2024 International Conference on Space Robotics (iSpaRo)},
  pages={224--230},
  year={2024},
  organization={IEEE}
}

@article{koktas2024communications,
  title={Communications for the planet {Mars}: Past, present, and future},
  author={Koktas, Enes and Ba{\c{s}}ar, Ertu{\u{g}}rul},
  journal={IEEE Aerospace and Electronic Systems Magazine},
  volume={39},
  number={9},
  pages={216--258},
  year={2024},
  publisher={IEEE}
}

@misc{ieee802154,
  author       = {{IEEE}},
  title        = {{IEEE} Standard for Low-Rate Wireless Networks},
  note         = {{IEEE} Std 802.15.4-2020 (Revision of {IEEE} Std 802.15.4-2015)},
  howpublished = {[Online]. Available: \url{https://standards.ieee.org/ieee/802.15.4/7029/}},
  year         = {2020},
}

@inproceedings{berger2001rad750,
  title={The {RAD750}---A Radiation Hardened {PowerPC} Processor for High Performance Spaceborne Applications},
  author={Berger, Richard W. and Bayles, Devin and Brown, Ronald and Doyle, Scott and Kazemzadeh, Abbas and Knowles, Ken and Moser, David and Rodgers, John and Saari, Brian and Stanley, Dan and Grant, B.},
  booktitle={2001 {IEEE} Aerospace Conference Proceedings (Cat. No. 01TH8542)},
  volume={5},
  pages={2263--2272},
  year={2001},
  organization={IEEE},
  doi={10.1109/AERO.2001.931184}
}

@misc{ieee802154g,
  author       = {{IEEE}},
  title        = {{IEEE} Standard for Local and Metropolitan Area Networks
                  -- Part 15.4: Low-Rate Wireless Personal Area Networks
                  ({LR-WPANs}) Amendment 3: {PHY} Specifications for
                  Low-Data-Rate, Wireless, Smart Metering Utility Networks},
  note         = {{IEEE} Std 802.15.4g-2012},
  howpublished = {[Online]. Available: \url{https://standards.ieee.org/ieee/802.15.4g/5053/}},
  year         = {2012},
}

@inproceedings{hemminger2005network,
  title={Network Emulation with {NetEm}},
  author={Hemminger, Stephen},
  booktitle={Proceedings of Australia's 6th National Linux Conference ({linux.conf.au})},
  year={2005}
}

@article{kaess2012isam2,
  title={{iSAM2}: Incremental smoothing and mapping using the {Bayes} tree},
  author={Kaess, Michael and Johannsson, Hordur and Roberts, Richard and Ila, Viorela and Leonard, John J and Dellaert, Frank},
  journal={The International Journal of Robotics Research},
  volume={31},
  number={2},
  pages={216--235},
  year={2012},
  publisher={Sage Publications Sage UK: London, England}
}

@article{burgard2005coordinated,
  title={Coordinated multi-robot exploration},
  author={Burgard, Wolfram and Moors, Mark and Stachniss, Cyrill and Schneider, Frank E},
  journal={IEEE Transactions on robotics},
  volume={21},
  number={3},
  pages={376--386},
  year={2005},
  publisher={IEEE}
}

@inproceedings{kim2018scan,
  title={{Scan Context}: Egocentric spatial descriptor for place recognition within {3D} point cloud map},
  author={Kim, Giseop and Kim, Ayoung},
  booktitle=C-IROS,
  pages={4802--4809},
  year={2018},
  organization={IEEE}
}

@inproceedings{cop2018delight,
  title={{DELIGHT}: An efficient descriptor for global localisation using {LiDAR} intensities},
  author={Cop, Konrad P and Borges, Paulo VK and Dub{\'e}, Renaud},
  booktitle=C-ICRA,
  pages={3653--3660},
  year={2018},
  organization={IEEE}
}

@inproceedings{wang2020lidar,
  title={{LiDAR Iris} for loop-closure detection},
  author={Wang, Ying and Sun, Zezhou and Xu, Cheng-Zhong and Sarma, Sanjay E and Yang, Jian and Kong, Hui},
  booktitle=C-IROS,
  pages={5769--5775},
  year={2020},
  organization={IEEE}
}

@incollection{john2009iris,
  title={How Iris Recognition Works},
  author={Daugman, John},
  booktitle={The Essential Guide to Image Processing},
  edition={2},
  chapter={25},
  pages={715--739},
  year={2009},
  publisher={Academic Press}
}

@inproceedings{berton2022rethinking,
  title={Rethinking visual geo-localization for large-scale applications},
  author={Berton, Gabriele and Masone, Carlo and Caputo, Barbara},
  booktitle=C-CVPR,
  pages={4878--4888},
  year={2022}
}

@inproceedings{takeuchi20063,
  title={A {3-D} scan matching using improved {3-D} normal distributions transform for mobile robotic mapping},
  author={Takeuchi, Eijiro and Tsubouchi, Takashi},
  booktitle=C-IROS,
  pages={3068--3073},
  year={2006},
  organization={IEEE}
}

@inproceedings{vidanapathirana2022logg3d,
  title={{LoGG3D-Net}: Locally guided global descriptor learning for {3D} place recognition},
  author={Vidanapathirana, Kavisha and Ramezani, Milad and Moghadam, Peyman and Sridharan, Sridha and Fookes, Clinton},
  booktitle=C-ICRA,
  pages={2215--2221},
  year={2022},
  organization={IEEE}
}

@article{ma2022overlaptransformer,
  title={{OverlapTransformer}: An efficient and yaw-angle-invariant transformer network for {LiDAR}-based place recognition},
  author={Ma, Junyi and Zhang, Jun and Xu, Jintao and Ai, Rui and Gu, Weihao and Chen, Xieyuanli},
  journal=IEEE_J_RAL,
  volume={7},
  number={3},
  pages={6958--6965},
  year={2022},
  publisher={IEEE}
}

@article{kim2024narrowing,
  title={Narrowing your {FOV} with {SOLiD}: Spatially organized and lightweight global descriptor for {FOV}-constrained {LiDAR} place recognition},
  author={Kim, Hogyun and Choi, Jiwon and Sim, Taehu and Kim, Giseop and Cho, Younggun},
  journal=IEEE_J_RAL,
  year={2024},
  publisher={IEEE}
}

@article{yang2020teaser,
  title={{TEASER}: Fast and certifiable point cloud registration},
  author={Yang, Heng and Shi, Jingnan and Carlone, Luca},
  journal=IEEE_J_RO,
  volume={37},
  number={2},
  pages={314--333},
  year={2020},
  publisher={IEEE}
}

@article{cattaneo2022lcdnet,
  title={{LCDNet}: Deep loop closure detection and point cloud registration for {LiDAR} {SLAM}},
  author={Cattaneo, Daniele and Vaghi, Matteo and Valada, Abhinav},
  journal={IEEE Transactions on Robotics},
  volume={38},
  number={4},
  pages={2074--2093},
  year={2022},
  publisher={IEEE}
}

@inproceedings{lim2024kiss,
  title={{KISS-Matcher}: Fast and Robust Point Cloud Registration Revisited},
  author={Lim, Hyungtae and Kim, Daebeom and Shin, Gunhee and Shi, Jingnan and Vizzo, Ignacio and Myung, Hyun and Park, Jaesik and Carlone, Luca},
  booktitle={2025 {IEEE} International Conference on Robotics and Automation ({ICRA})},
  pages={11104--11111},
  year={2025},
  organization={IEEE},
  doi={10.1109/ICRA55743.2025.11127458}
}

@article{fischler1981random,
  title={Random sample consensus: a paradigm for model fitting with applications to image analysis and automated cartography},
  author={Fischler, Martin A. and Bolles, Robert C.},
  journal={Communications of the {ACM}},
  volume={24},
  number={6},
  pages={381--395},
  year={1981},
  doi={10.1145/358669.358692}
}

@article{yang2020graduated,
  title={Graduated non-convexity for robust spatial perception: From non-minimal solvers to global outlier rejection},
  author={Yang, Heng and Antonante, Pasquale and Tzoumas, Vasileios and Carlone, Luca},
  journal=IEEE_J_RAL,
  volume={5},
  number={2},
  pages={1127--1134},
  year={2020},
  publisher={IEEE}
}

@article{tian2022kimera,
  title={{Kimera-Multi}: Robust, distributed, dense metric-semantic {SLAM} for multi-robot systems},
  author={Tian, Yulun and Chang, Yun and Arias, Fernando Herrera and Nieto-Granda, Carlos and How, Jonathan P and Carlone, Luca},
  journal=IEEE_J_RO,
  volume={38},
  number={4},
  year={2022},
  publisher={IEEE}
}

@inproceedings{mangelson2018pairwise,
  title={Pairwise consistent measurement set maximization for robust multi-robot map merging},
  author={Mangelson, Joshua G and Dominic, Derrick and Eustice, Ryan M and Vasudevan, Ram},
  booktitle=C-ICRA,
  pages={2916--2923},
  year={2018},
  organization={IEEE}
}

@article{small_gicp,
author = {Kenji Koide},
title = {{small\_gicp: Efficient and parallel algorithms for point cloud registration}},
journal = {Journal of Open Source Software},
month = aug,
number = {100},
pages = {6948},
volume = {9},
year = {2024},
doi = {10.21105/joss.06948}
}

@article{lajoie2022towards,
  title={Towards collaborative simultaneous localization and mapping: a survey of the current research landscape},
  author={Lajoie, Pierre-Yves and Ramtoula, Benjamin and Wu, Fang and Beltrame, Giovanni},
  journal={Field Robotics},
  volume={2},
  pages={971--1000},
  year={2022},
  publisher={FRPS}
}

@article{novakova2022correcting,
  title={Correcting laser scanning intensity recorded in a cave environment for high-resolution lithological mapping: A case study of the {Gouffre Georges}, {France}},
  author={Nov{\'a}kov{\'a}, Michaela and Gallay, Michal and {\v{S}}upinsk{\'y}, Jozef and Ferr{\'e}, Eric and Asti, Riccardo and de Saint Blanquat, Michel and Bajolet, Flora and Sorriaux, Patrick},
  journal={Remote Sensing of Environment},
  volume={280},
  pages={113210},
  year={2022}
}

@inproceedings{rusu2009fast,
  title={Fast point feature histograms ({FPFH}) for {3D} registration},
  author={Rusu, Radu Bogdan and Beetz, Michael},
  booktitle=C-ICRA,
  pages={3212--3217},
  year={2009},
  organization={IEEE}
}

@incollection{smith1990estimating,
  title={Estimating uncertain spatial relationships in robotics},
  author={Smith, Randall and Self, Matthew and Cheeseman, Peter},
  booktitle={Autonomous robot vehicles},
  pages={167--193},
  year={1990},
  publisher={Springer}
}

@article{dellaert2012factor,
  title={Factor graphs and {GTSAM}: A hands-on introduction},
  author={Dellaert, Frank},
  journal={Georgia Institute of Technology, Tech. Rep},
  volume={2},
  number={4},
  year={2012}
}

@article{umeyama1991least,
  title={Least-squares estimation of transformation parameters between two point patterns},
  author={Umeyama, Shinji},
  journal=IEEE_J_PAMI,
  volume={13},
  number={04},
  pages={376--380},
  year={1991},
  publisher={IEEE Computer Society}
}

@misc{grupp2017evo,
  title={{evo}: {Python} package for the evaluation of odometry and {SLAM}.},
  author={Grupp, Michael},
  howpublished={\url{https://github.com/MichaelGrupp/evo}},
  year={2017}
}

@inproceedings{rusinkiewicz2001efficient,
  title={Efficient variants of the {ICP} algorithm},
  author={Rusinkiewicz, Szymon and Levoy, Marc},
  booktitle={Proceedings third international conference on 3-D digital imaging and modeling},
  pages={145--152},
  year={2001},
  organization={IEEE}
}

@article{low2004linear,
  title={Linear least-squares optimization for point-to-plane {ICP} surface registration},
  author={Low, Kok-Lim},
  journal={Chapel Hill, Univ. of North Carolina},
  volume={4},
  number={10},
  pages={1--3},
  year={2004}
}

@inproceedings{segal2009generalized,
  title={{Generalized-ICP}.},
  author={Segal, Aleksandr and Haehnel, Dirk and Thrun, Sebastian},
  booktitle=C-RSS,
  volume={2},
  number={4},
  pages={435},
  year={2009},
  organization={Seattle, WA}
}

@inproceedings{koide2021voxelized,
  title={Voxelized {GICP} for fast and accurate {3D} point cloud registration},
  author={Koide, Kenji and Yokozuka, Masashi and Oishi, Shuji and Banno, Atsuhiko},
  booktitle=C-ICRA,
  pages={11054--11059},
  year={2021},
  organization={IEEE}
}

@article{chen2026heterogeneous,
  title={Heterogeneous {LiDAR} dataset for benchmarking robust localization in diverse degenerate scenarios},
  author={Chen, Zhiqiang and Qi, Yuhua and Feng, Dapeng and Zhuang, Xuebin and Chen, Hongbo and Hu, Xiangcheng and Wu, Jin and Peng, Kelin and Lu, Peng},
  journal={The International Journal of Robotics Research},
  volume={45},
  number={1},
  pages={6--22},
  year={2026},
  publisher={SAGE Publications Sage UK: London, England}
}

@article{lim2024quatro++,
  title={{Quatro++}: Robust global registration exploiting ground segmentation for loop closing in {LiDAR SLAM}},
  author={Lim, Hyungtae and Kim, Beomsoo and Kim, Daebeom and Mason Lee, Eungchang and Myung, Hyun},
  journal={The International Journal of Robotics Research},
  volume={43},
  number={5},
  pages={685--715},
  year={2024}
}

@inproceedings{zhang2014loam,
  title={{LOAM}: {LiDAR} Odometry and Mapping in Real-time},
  author={Zhang, Ji and Singh, Sanjiv},
  booktitle=C-RSS,
  volume={2},
  number={9},
  pages={1--9},
  year={2014}
}

@inproceedings{shan2020lio,
  title={{LIO-SAM}: Tightly-coupled {LiDAR} Inertial Odometry via Smoothing and Mapping},
  author={Shan, Tixiao and Englot, Brendan and Meyers, Drew and Wang, Wei and Ratti, Carlo and Rus, Daniela},
  booktitle=C-IROS,
  pages={5135--5142},
  year={2020},
  organization={IEEE}
}

@article{xu2021fast,
  title={{FAST-LIO}: A Fast, Robust {LiDAR}-inertial Odometry Package by Tightly-Coupled Iterated {Kalman} Filter},
  author={Xu, Wei and Zhang, Fu},
  journal=IEEE_J_RAL,
  volume={6},
  number={2},
  pages={3317--3324},
  year={2021},
  publisher={IEEE}
}

@article{bai2022faster,
  title={{Faster-LIO}: Lightweight Tightly Coupled {LiDAR}-Inertial Odometry Using Parallel Sparse Incremental Voxels},
  author={Bai, Chunge and Xiao, Tao and Chen, Yajie and Wang, Haoqian and Zhang, Fang and Gao, Xiang},
  journal=IEEE_J_RAL,
  volume={7},
  number={2},
  pages={4861--4868},
  year={2022},
  publisher={IEEE}
}

@inproceedings{chen2023direct,
  title={Direct {LiDAR}-Inertial Odometry: Lightweight {LIO} with Continuous-Time Motion Correction},
  author={Chen, Kenny and Nemiroff, Ryan and Lopez, Brett T.},
  booktitle=C-ICRA,
  year={2023},
  organization={IEEE}
}

@inproceedings{pfreundschuh2024coin,
  title={{COIN-LIO}: Complementary Intensity-Augmented {LiDAR} Inertial Odometry},
  author={Pfreundschuh, Patrick and Oleynikova, Helen and Cadena, Cesar and Siegwart, Roland and Andersson, Olov},
  booktitle=C-ICRA,
  year={2024},
  organization={IEEE}
}

@article{xu2022fast,
  title={{FAST-LIO2}: Fast direct {LiDAR}-inertial odometry},
  author={Xu, Wei and Cai, Yixi and He, Dongjiao and Lin, Jiarong and Zhang, Fu},
  journal={IEEE Transactions on Robotics},
  volume={38},
  number={4},
  pages={2053--2073},
  year={2022},
  publisher={IEEE}
}

@article{zhang2026deep,
  title={Deep learning-based point cloud registration: A comprehensive survey and taxonomy},
  author={Zhang, Yu-Xin and Gui, Jie and Yu, Baosheng and Cong, Xiaofeng and Gong, Xin and Tao, Wenbing and Tao, Dacheng},
  journal={International Journal of Computer Vision},
  volume={134},
  number={4},
  pages={168},
  year={2026},
  publisher={Springer}
}

@article{thakker2025risk,
  title={Risk-Guided Diffusion: Toward Deploying Robot Foundation Models In Space, Where Failure Is Not An Option},
  author={Thakker, Rohan and Patnaik, Adarsh and Kurtz, Vince and Frey, Jonas and Becktor, Jonathan and Moon, Sangwoo and Royce, Rob and Kaufmann, Marcel and Georgakis, Georgios and Roth, Pascal and others},
  journal={arXiv preprint arXiv:2506.17601},
  year={2025}
}

@inproceedings{holden2025vision,
  title={Vision foundation models for domain generalisable cross-view localisation in planetary ground-aerial robotic teams},
  author={Holden, Lachlan and Dayoub, Feras and Candela, Alberto and Harvey, David and Chin, Tat-Jun},
  booktitle={2025 International Conference on Space Robotics (iSpaRo)},
  pages={154--160},
  year={2025},
  organization={IEEE}
}

@article{kim2026commerge,
  title={{Commerge}: Communication-Efficient, Robust, and Fast {LiDAR} Map Merging Framework for Multi-Robot Coordination in Resource-Constrained Scenarios},
  author={Kim, Hogyun and Choi, Jiwon and Kim, Juwon and Yang, Geonmo and Jeong, Seokhwan and Lim, Hyungtae and Cho, Younggun},
  journal={arXiv preprint arXiv:2606.25386},
  year={2026}
}

@STRING{C-CVPR = {Proc. {IEEE} Conf. on Comput. Vision and Pattern Recog.}}

@STRING{C-ICRA = {Proc. {IEEE} Intl. Conf. on Robot. and Automat.}}

@STRING{C-IROS = {Proc. {IEEE}/{RSJ} Intl. Conf. on Intell. Robots and Sys.}}

@STRING{C-RSS = {Proc. Robot.: Science \& Sys. Conf.}}

@STRING{IEEE_J_ITS = {{IEEE} Trans. Intell. Transport. Sys.}}

@STRING{IEEE_J_PAMI = {{IEEE} Trans. Pattern Analysis and Machine Intell.}}

@STRING{IEEE_J_RAL = {{IEEE} Robot. and Automat. Lett.}}

@STRING{IEEE_J_RO = {{IEEE} Trans. Robot.}}

@STRING{IEEE_J_SENSOR = {{IEEE} Sensors J.}}
\endgroup

\vfill\pagebreak

\begin{IEEEbiography}[{\includegraphics[width=1in,height=1.25in,clip,keepaspectratio]{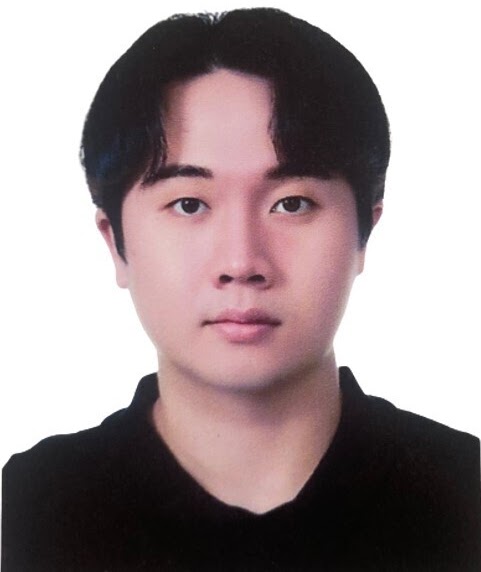}}]
{Hogyun Kim}~(Student Member, IEEE)~received the B.S. degree in naval architecture and ocean engineering from Inha University, Incheon, Republic of Korea, in 2023, and is currently pursuing the Ph.D. degree with the Spatial AI and Robotics~(SPARO) Laboratory, Department of Electrical and Computer Engineering.
His research interests include simultaneous localization and mapping~(SLAM), multi-robot systems, collaborative intelligence, and field robotics.

Mr. Kim serves as a student representative for the IEEE International Symposium on Safety, Security, and Rescue Robotics~(SSRR) 2026.
His team was selected as a finalist for the Best Paper Award on Robot Perception at IEEE ICRA 2026.
His team won second place in the Hilti x Trimble SLAM Challenge 2026.
\end{IEEEbiography}

\begin{IEEEbiography}[{\includegraphics[width=1in,height=1.25in,clip,keepaspectratio]{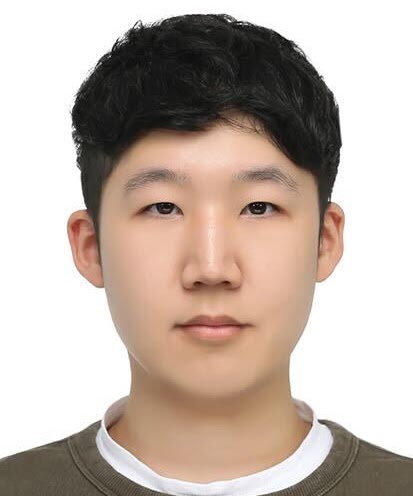}}]
{Juwon Kim}~(Student Member, IEEE)~is currently pursuing the M.S. degree with the Spatial AI and Robotics~(SPARO) Laboratory, Department of Electrical and Computer Engineering, Inha University, Incheon, Republic of Korea.
His research interests include advanced state estimation for quadruped locomotion, simultaneous localization and mapping~(SLAM), and the application of state space models for high-rate time-series processing in legged robots.
\end{IEEEbiography}

\begin{IEEEbiography}[{\includegraphics[width=1in,height=1.25in,clip,keepaspectratio]{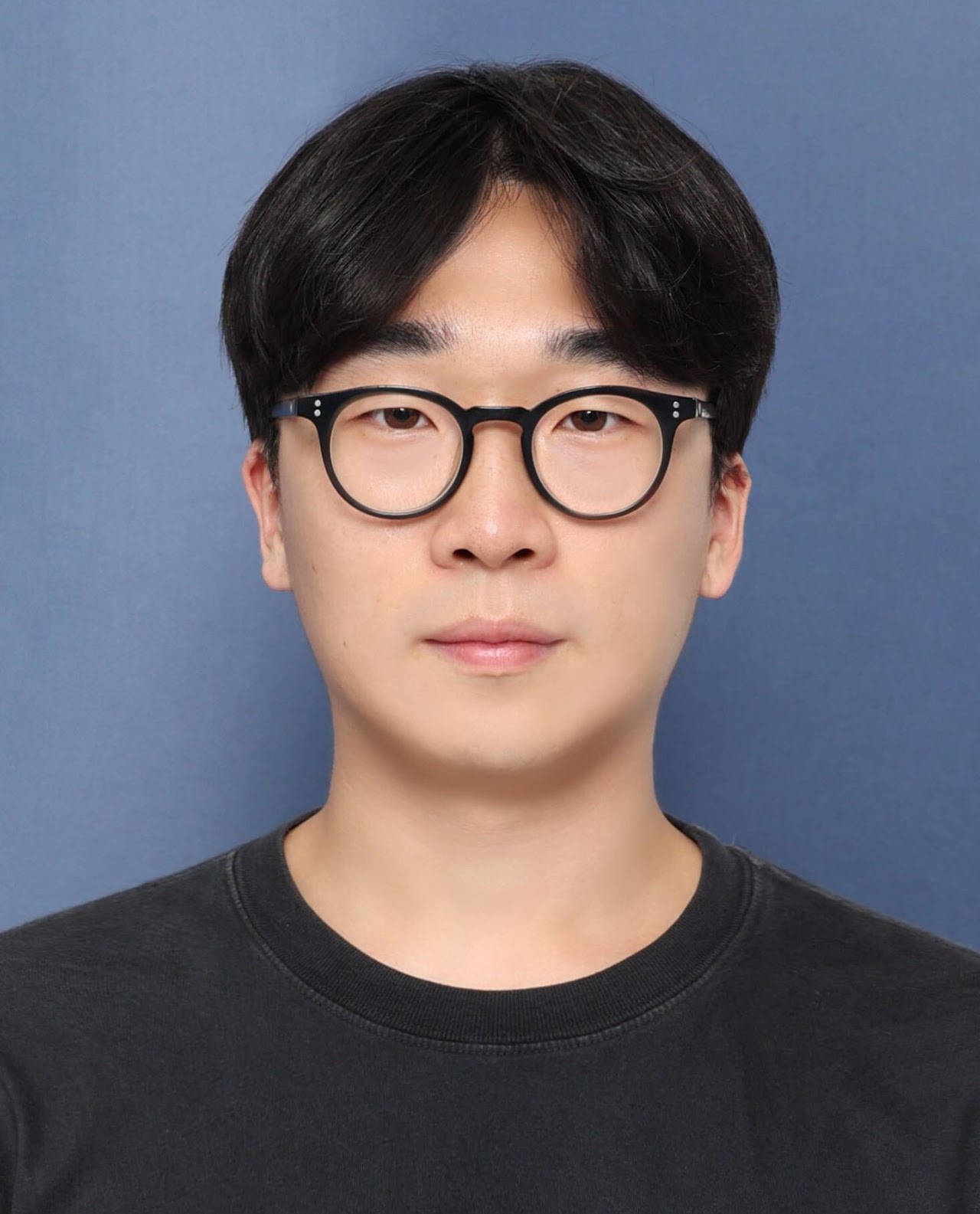}}]
{Dongjin Cho}~(Student Member, IEEE)~received the B.S. degree in electronic engineering from Tech University of Korea~(TUKorea), Siheung, Republic of Korea, in 2024, and the M.S. degree in electrical and computer engineering from Inha University, Incheon, Republic of Korea, in 2026, where he is currently pursuing the Ph.D. degree with the Spatial AI and Robotics~(SPARO) Laboratory.
His research interests include perception for long-term autonomy and traversability estimation.
\end{IEEEbiography}

\begin{IEEEbiography}[{\includegraphics[width=1in,height=1.25in,clip,keepaspectratio]{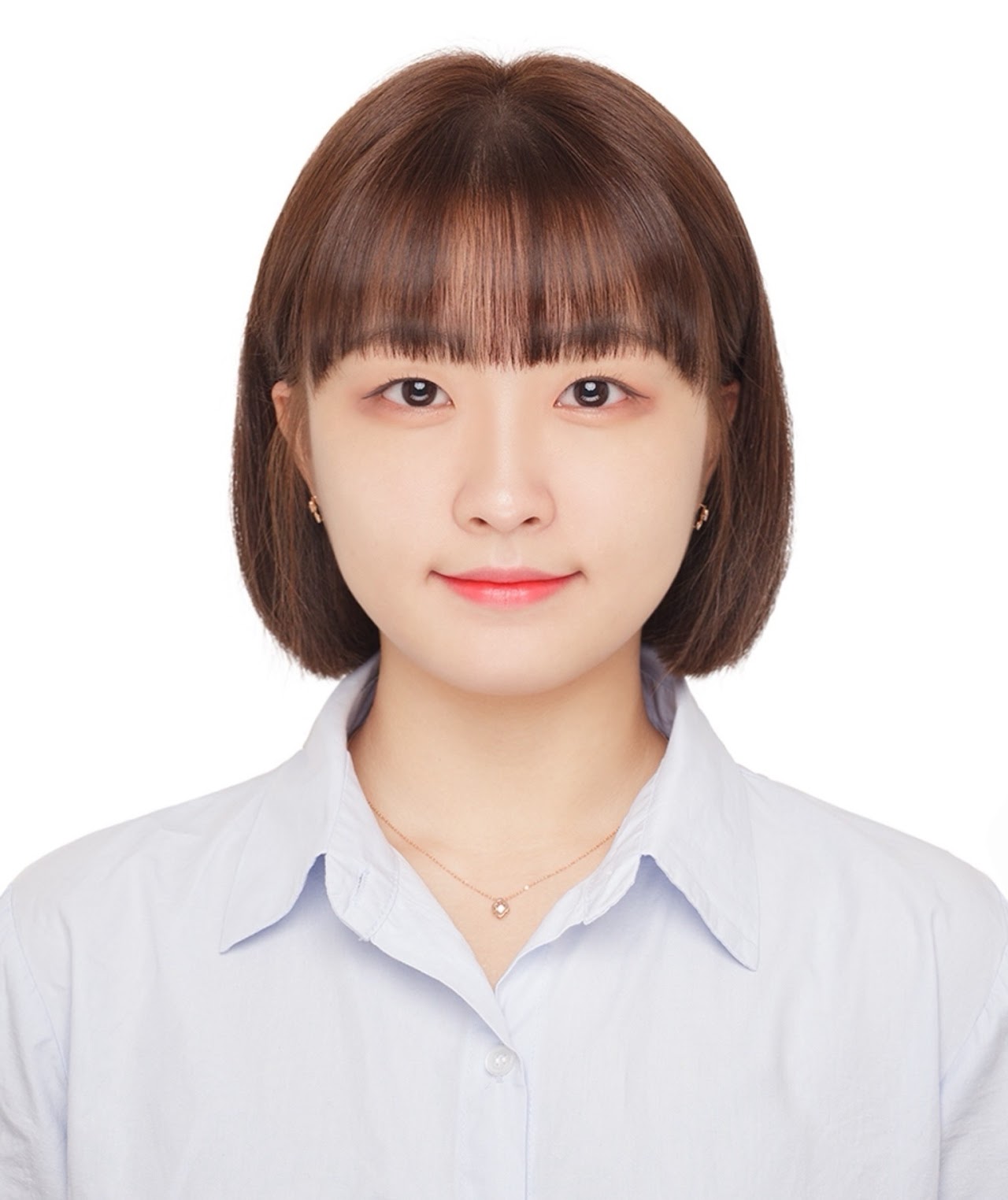}}]
{Jiwon Choi}~(Student Member, IEEE)~received the B.S. degree in electrical engineering from Inha University, Incheon, Republic of Korea, in 2024, and is currently pursuing the Ph.D. degree with the Spatial AI and Robotics~(SPARO) Laboratory, Department of Electrical and Computer Engineering.
Her research interests include self-supervised inertial learning and inertial odometry, resilient state estimation under sensor degradation and limited supervision, proprioceptive state estimation for legged robots from inertial, kinematic, and contact sensing, and multi-modal simultaneous localization and mapping~(SLAM).

Ms. Choi received the Outstanding WiRA Student Paper Award and was selected as a finalist for the Best Paper Award on Robot Perception at IEEE ICRA 2026.
Her team won second place in the Hilti x Trimble SLAM Challenge 2026.
\end{IEEEbiography}

\newpage

\begin{IEEEbiography}[{\includegraphics[width=1in,height=1.25in,clip,keepaspectratio]{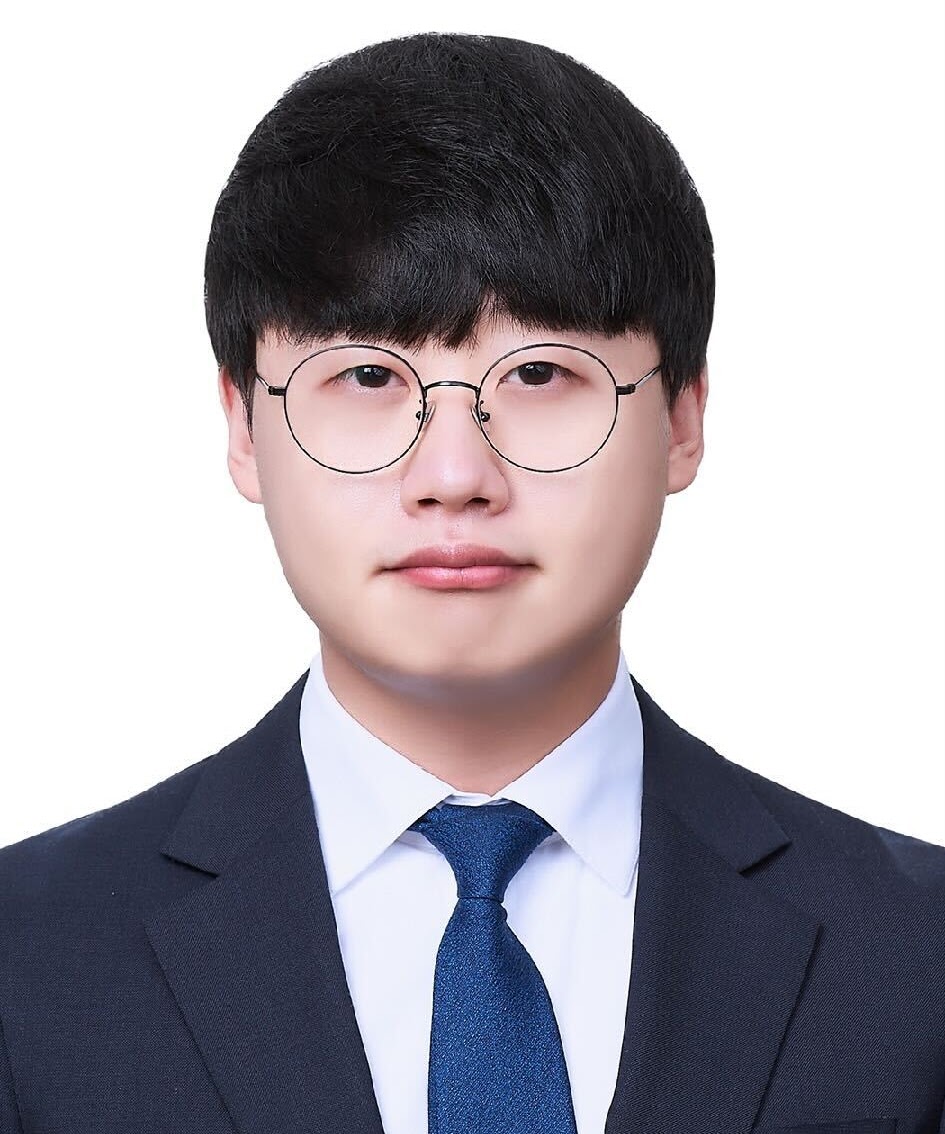}}]
{Geonmo Yang}~(Student Member, IEEE)~received the B.S. degree in electronic engineering from Tech University of Korea~(TUKorea), Siheung, Republic of Korea, in 2022, and is currently pursuing the Ph.D. degree with the Spatial AI and Robotics~(SPARO) Laboratory, Department of Electrical and Computer Engineering, Inha University, Incheon, Republic of Korea.
His research interests include perception for underwater and marine robotics, multi-modal simultaneous localization and mapping~(SLAM), and robotics for long-term autonomy.

His team was selected as a finalist for the Best Paper Award on Robot Perception at IEEE ICRA 2026.
\end{IEEEbiography}
\vspace{\biogapA}

\begin{IEEEbiography}[{\includegraphics[width=1in,height=1.25in,clip,keepaspectratio]{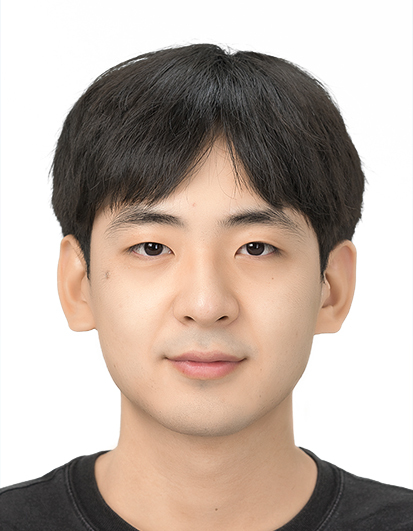}}]
{Seokhwan Jeong}~(Student Member, IEEE)~received the B.S. degree in electrical engineering and the M.S. degree in electrical and computer engineering from Inha University, Incheon, Republic of Korea, in 2023 and 2025, respectively, where he is currently pursuing the Ph.D. degree with the Spatial AI and Robotics~(SPARO) Laboratory.
His research interests include sensor calibration.
\end{IEEEbiography}
\vspace{\biogapB}

\begin{IEEEbiography}[{\includegraphics[width=1in,height=1.25in,clip,keepaspectratio]{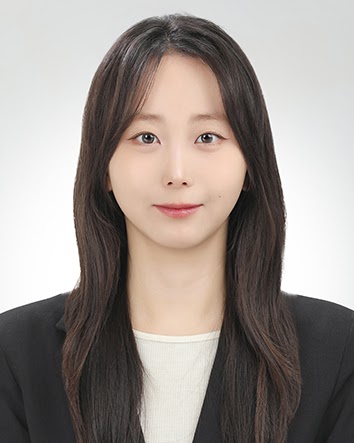}}]
{Jiyun Lee}~(Student Member, IEEE)~received the B.S. degree in computer engineering from Inha University, Incheon, Republic of Korea, in 2025, and is currently pursuing the M.S. degree with the Spatial AI and Robotics~(SPARO) Laboratory, Department of Electrical and Computer Engineering.
Her research interests include reinforcement learning and autonomous navigation.
\end{IEEEbiography}
\vspace{\biogapC}

\begin{IEEEbiography}[{\includegraphics[width=1in,height=1.25in,clip,keepaspectratio]{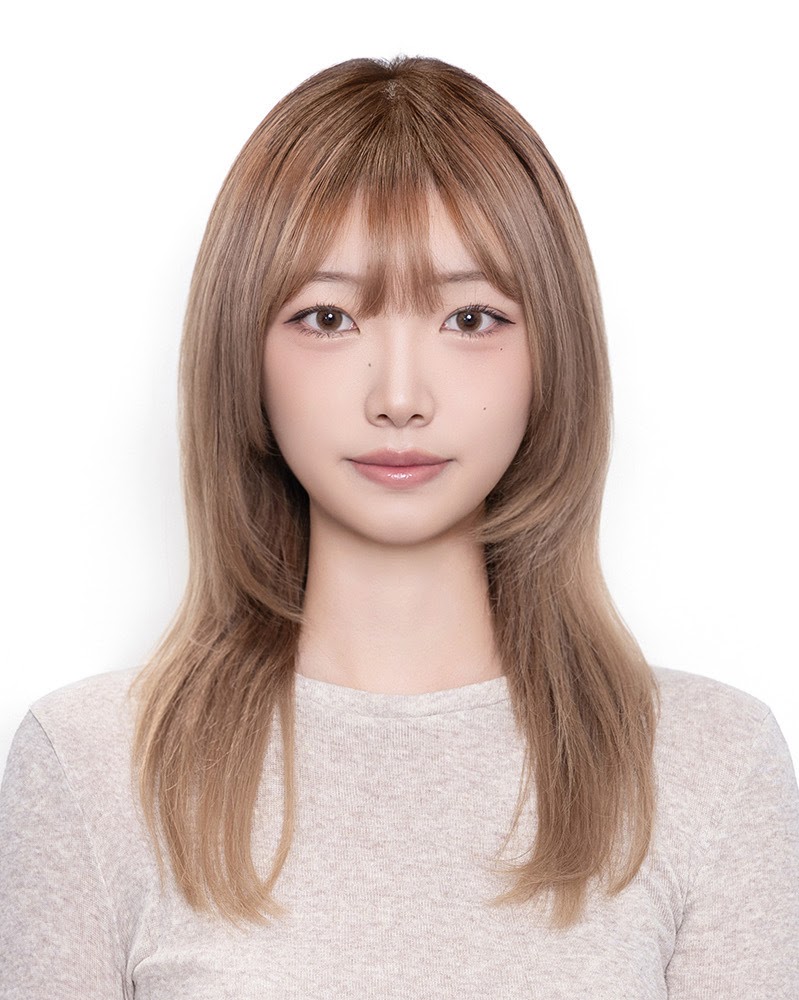}}]
{Miryeong Park}~(Student Member, IEEE)~received the B.S. degree in aerospace engineering from Inha University, Incheon, Republic of Korea, in 2025, and is currently pursuing the M.S. degree with the Spatial AI and Robotics~(SPARO) Laboratory, Department of Electrical and Computer Engineering.
Her research interests include autonomous exploration, path planning, and field robotics.
\end{IEEEbiography}

\newpage

\begin{IEEEbiography}[{\includegraphics[width=1in,height=1.25in,clip,keepaspectratio]{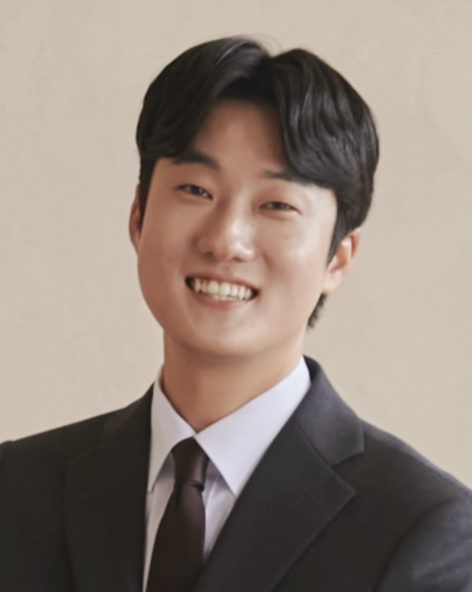}}]
{Hyungtae Lim}~(Member, IEEE)~received the B.S. degree in mechanical
engineering and the M.S. and Ph.D. degrees in electrical engineering from
the Korea Advanced Institute of Science and Technology~(KAIST), Daejeon,
Republic of Korea, in 2018, 2020, and 2023, respectively.

He was a Postdoctoral Associate with the Laboratory for Information \&
Decision Systems~(LIDS) and the SPARK Lab at the Massachusetts Institute of
Technology~(MIT), Cambridge, MA, USA. Before joining MIT, he was a
Postdoctoral Fellow with the Urban Robotics Laboratory at KAIST. During his
doctoral studies, he was a Visiting Scholar with StachnissLab at the
University of Bonn, Bonn, Germany, and a Research Intern with NAVER LABS,
Seongnam, Republic of Korea. His research interests include simultaneous
localization and mapping~(SLAM), robust 3-D registration and perception,
long-term map management, spatial AI, and deep learning for mobile robots
and autonomous vehicles.

Dr. Lim is an Associate Editor for the IEEE Robotics and Automation Letters.
He was selected as an RSS Pioneer in 2024 and received the 2022 IEEE Robotics
and Automation Letters Best Paper Award and the IEEE ICRA 2025 Outstanding
Reviewer Award. His team won first place in the Hilti SLAM Challenge at IEEE
ICRA 2023, the Nothing Stands Still Challenge at IEEE ICRA 2024, the Hilti x
Trimble SLAM Challenge 2026, and the GOOSE 2D Fine-Grained Semantic
Segmentation Challenge at IEEE ICRA 2026.
\end{IEEEbiography}

\newpage

\begin{IEEEbiography}[{\includegraphics[width=1in,height=1.25in,clip,keepaspectratio]{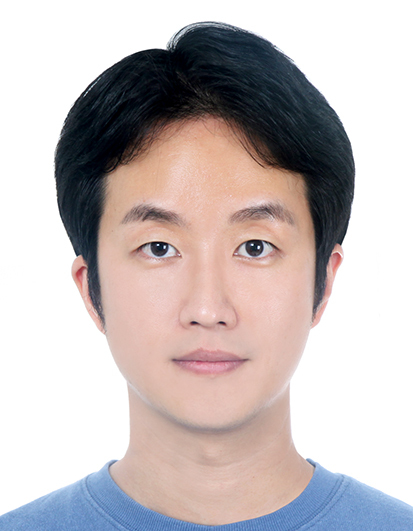}}]
{Younggun Cho}~(Member, IEEE)~received the B.S. degree in electrical engineering from Inha University, Incheon, Republic of Korea, in 2013, and the M.S. degree in electrical engineering and the Ph.D. degree in civil and environmental engineering, with a dual degree in the Robotics Program, from the Korea Advanced Institute of Science and Technology~(KAIST), Daejeon, Republic of Korea, in 2015 and 2020, respectively.

From 2020 to 2021, he was an Assistant Professor with the Department of Robot Engineering, Yeungnam University, Gyeongsan, Republic of Korea. 
In 2021, he joined the School of Electrical and Electronic Engineering, Inha University, Incheon, Republic of Korea, where he is currently an Associate Professor and leads the Spatial AI and Robotics~(SPARO) Laboratory. 
Since 2025, he has also served as a Technical Advisor to Riibotics. His research interests include multi-robot SLAM and collaborative spatial intelligence, robust perception for field robots, learning-assisted mapping and localization, and long-term robot autonomy in challenging environments.

Dr. Cho is an Associate Editor for the IEEE Robotics and Automation Letters and a Co-Chair of the IEEE Robotics and Automation Society Technical Committee on Marine Robotics. He currently serves on the organizing committees of the IEEE International Symposium on Safety, Security, and Rescue Robotics~(SSRR) 2026 and the IEEE International Conference on Robotics and Automation~(ICRA) 2027. 
He has also served on the organizing committees of IEEE UR, ICCAS, and KRoC over multiple years. 
He received the New Generation Star Project Award at IEEE/RSJ IROS 2025 and the Best Paper Award from ICROS in 2024.
In 2026, a paper from his group was selected as an IEEE ICRA Award Finalist, and his team placed second in the SLAM category of the Hilti x Trimble SLAM Challenge at IEEE ICRA.
\end{IEEEbiography}

\end{document}